\documentclass{article}
\usepackage{arxiv}

\usepackage[utf8]{inputenc} 
\usepackage[T1]{fontenc}    
\usepackage{url}   
\usepackage{booktabs} 
\usepackage{amsfonts} 
\usepackage{nicefrac} 
\usepackage{microtype}
\usepackage{lipsum}
\usepackage{graphicx}
\graphicspath{ {./figures/} }

\usepackage{siunitx}
\usepackage{multirow}
\usepackage[table]{xcolor}
\usepackage{listings} 
\usepackage{xcolor} 
\usepackage{amsmath}
\usepackage{amssymb}
\usepackage{amsthm}
\usepackage{mathtools}
\usepackage{tikz}
\usetikzlibrary{matrix,chains,positioning,decorations.pathreplacing,arrows}
\usepackage{algorithm}
\usepackage[normalem]{ulem}
\usepackage[noend]{algpseudocode}
\usepackage{filecontents}
\usepackage{xr-hyper}
\usepackage{hyperref} 
\usepackage{cleveref}

\makeatletter
\newcommand*{\addFileDependency}[1]{
  \typeout{(#1)}
  \@addtofilelist{#1}
  \IfFileExists{#1}{}{\ypeout{No file #1.}}
}
\makeatother

\makeatletter
\newcommand*{\addAuxFileDependency}[1]{
  \makeatletter\@input{x#1.tex}\makeatother
}
\makeatother

\newcommand*{\myexternaldocument}[1]{%
    \externaldocument[#1:]{#1}%
    \addAuxFileDependency{#1}%
    \addFileDependency{#1.tex}%
    \addFileDependency{#1.aux}%
}

\small

\newcommand{\parsm}{\boldsymbol{\omega_{\mathrm{s}}}}
\newcommand{\parmkm}{\mathbf{p}}
\newcommand{\parsmt}{\boldsymbol{\omega_{\mathrm{t}}}}
\newcommand{\parlumped}{\boldsymbol{\omega}}
\newcommand{\opfunbc}{\phi}
\newcommand{\jd}{{j}_{\mathrm{d}}}
\newcommand{\jt}{{j}_{\mathrm{t}}}
\newcommand{\jm}{{j}_{\mathrm{m}}}

\newcommand{\kUDE}{KINN}
\newcommand{\bcD}{\mathbf{\mathrm{C}_{ic}^D}}
\newcommand{\opnc}{\mathbf{\mathrm{C}_{N}}}
\newcommand {\bc}{IC}
\newcommand {\SM}{Supplementary Material}

\newcommand{\obsonly}{\textit{Q}}
\newcommand{\semiquant}{\textit{SQ}}
\newcommand{\semiquantn}{\textit{SQ}+n}
\newcommand{\erresp}{\boldsymbol{\varepsilon}}
\newcommand{\xvec}{\mathbf{x}}
\newcommand{\dxvec}{\mathbf{\dot{x}}}

\newcommand{\fmap}{\Pi}

\numberwithin{equation}{section}
\numberwithin{equation}{subsection}

\algrenewcommand{\algorithmicrequire}{\textbf{Input:}}
\algrenewcommand{\algorithmicensure}{\textbf{Output:}}

\crefname{equation}{Eq.}{Eqs.}
\Crefname{equation}{Equation}{Equations}
\crefname{table}{Table}{Tables}
\Crefname{table}{Table}{Tables}
\crefname{figure}{Fig.}{Figs.}
\Crefname{figure}{Figure}{Figures}

\title{Kinetics-Informed Neural Networks}

\author{
  Gabriel S. Gusmão\textsuperscript{a}, Adhika P. Retnanto\textsuperscript{a,1}, Shashwati~C.~da~Cunha\textsuperscript{a,1}, Andrew J. ~Medford\textsuperscript{a}\\\\
  \textsuperscript{a}School of Chemical \& Biomolecular Engineering\\
  Georgia Institute of Technology\\
  Atlanta, GA 30332\\
  \texttt{\{gusmaogabriels, aretnanto6, shashwatidc, ajm\}@gatech.edu}\\\\
  \textit{\textsuperscript{1}These authors contributed equally to this work.}
}

\myexternaldocument{supp}

\begin{document}

\maketitle

\begin{abstract}
Chemical kinetics and reaction engineering consists of the phenomenological framework for the disentanglement of reaction mechanisms, optimization of reaction performance and the rational design of chemical processes. Here, we utilize feed-forward artificial neural networks as basis functions to solve ordinary differential equations (ODEs) constrained by differential algebraic equations (DAEs) that describe microkinetic models (MKMs). We present an algebraic framework for the mathematical description and classification of reaction networks, types of elementary reaction, and chemical species. Under this framework, we demonstrate that the simultaneous training of neural nets and kinetic model parameters in a regularized multi-objective optimization setting leads to the solution of the inverse problem through the estimation of kinetic parameters from synthetic experimental data. We analyze a set of scenarios to establish the extent to which kinetic parameters can be retrieved from transient kinetic data, and assess the robustness of the methodology with respect to statistical noise. This approach to inverse kinetic ODEs can assist in the elucidation of reaction mechanisms based on transient data.
\end{abstract}


\section{Introduction}
\par

In recent decades, considerable attention has been devoted to chemical process intensification through the optimal design of unit operations based on rigorous modeling of thermodynamics, heat, mass and momentum transport, and chemical reactions~\cite{Kim2017ModularReview}. Selectivity and yield of a chemical transformation, which are key factors in economic evaluation and design of associated unit operations, are determined by the intrinsic kinetics, transport resistances and thermodynamics constraints. The detailed description of chemical kinetics, especially in heterogeneous catalysis, is hence of utmost importance in the design and operation of chemical processes. Kinetic models provide the phenomenological framework for the elucidation of reaction mechanism, optimization of reaction performance, and process design~\cite{Medford2018}. Structurally, the detailed description of a chemical reaction network consists of the set of elementary steps involving bond breaking and/or formation from reactants to products~\cite{Boudart1984KineticsReactions,Dumesic1993}. Chemical kinetics can be described in terms of the transition-state barriers between thermodynamically stable intermediate states. Despite the high complexity of chemical kinetics networks, only few elementary steps typically control catalytic activity~\cite{Jones2008,Ulissi2017a, Stegelmann2009}. 

The mathematical description of chemical process models has departed from kinetic models that assume a single rate-limiting step and become more comprehensive~\cite{McBride2019OverviewEngineering}. The analysis and elucidation of reaction mechanism have benefited from the increasing accuracy
of computational chemistry methods such as density functional theory (DFT). These advances in computational capabilities have led to an increased prevalence of ``bottom-up'' kinetic models. In the bottom-up approach, DFT-derived free energies from proposed reaction mechanisms give rise to free energy diagrams. The ensuing microkinetic models (MKMs) have provided insights into rate-controlling steps~\cite{Stegelmann2009} and trends in catalytic activity in terms of reaction conditions and catalyst compositions~\cite{Nrskov2011,Medford2015b}. However, bottom-up modelling of complex reaction networks requires in-depth understanding of the detailed atomic structure of active sites as well as the molecular structure of the thermodynamically stable intermediate species and dominant elementary reactions. Moreover, assumptions are often required to connect the molecular-scale models with measurable catalytic quantities such as turnover frequencies, yields, and selectivities. The mean-field approximation (MFA) is typically invoked to avoid the need for a specific spatial description of catalytic active domains~\cite{Dumesic2008}. Transport limitations on the surface are regarded as negligible or averaged out uniformly across the available catalyst surface. This requires adsorbate-adsorbate interactions to be either neglected or included as coverage-dependent properties~\cite{Getman2009Oxygen-CoverageSurface,Lausche2013a,Mhadeshwar2004}. For more heterogeneous or non-ideal systems, rigorous kinetic Monte Carlo methods can be used to build coarse-grained representations of catalyst active sites, allowing for an explicit description of diffusive processes and surface interactions~\cite{Chatterjee2007}. However, these models are more difficult to parameterize, and in the absence of diffusion limitations and adsorbate-adsorbate interaction, studies by Hoffmann and Bliggard~\cite{Hoffmann2018} and by Andersen et. al.~\cite{Andersen2017} have shown that results derived from mean-field models approximate those of complex kinetic Monte Carlo schemes. Moreover, Reuter et. al. demonstrated that MFA models can be re-parameterized to reflect the results of complex kinetic Monte Carlo with great generalization properties~\cite{Sabbe2012}. These results suggest that the mean-field framework is promising for even complex catalytically active domains, though it may be necessary to depart from a direct mapping back to the molecular-scale system to obtain reliable results with the MFA.

Despite advances in first-principles and empirical mechanism elucidation, obtaining a bottom-up description of dynamic catalyst states over time-on-stream under operating conditions remains inherently challenging and computationally expensive.
It is also dependent on the specific system under study, since changes in catalyst state under different operating conditions may translate into shifts in the predominant chemical pathway, altering the selectivity and yield~\cite{Reuter2016}. For instance, the carbide-promoted Fischer-Tropsch reaction is a complex mechanism where the presence of carbide-, oxygen- and metallic-rich states exhibit strong dependence on reaction conditions~\cite{Jin2017ElementaryConditions,Chen2019}. Adsorbate-adsorbate interactions may also play a significant role in determining the preferential chemical pathway and the related rate-controlling steps~\cite{Mhadeshwar2004,Grabow2010UnderstandingMetals}. Furthermore, catalytic systems often encompass multiple active-site domains with different prevalent chemical pathways in which diffusive processes and adsorbate-adsorbate interaction may become relevant, as exemplified by Li et. al. in their comparison between mean-field approximation and kinetic Monte Carlo approaches for CO oxidation MKMs~\cite{Li2021EvaluatingInteractions}.  These nuances pose significant challenges in devising ab initio bottom-up descriptions of kinetic systems, particularly in the case of complex multi-component catalysts or large product/reactant molecules.

An alternative approach to kinetic model development is the ``top-down'' approach, where kinetic model parameters are fitted to the results of kinetic experiments. The top-down approach avoids the need for an explicit description of the catalyst active site, and can accommodate ``lumped'' reactions that include multiple elementary steps. However, the resulting models provide less detailed insight, and may not generalize well to other reaction conditions. In the top-down approach to chemical kinetics, the acquisition of steady-state data has traditionally been the foremost empirical procedure for the analysis and validation of reaction mechanism hypotheses~\cite{Dumesic1993}. However, acquiring intrinsic kinetic steady-state data is inherently laborious ~\cite{Perez-Ramirez2000Six-flowStudies}. The design of experiments must ensure that mass- and heat-transfer limitations are negligible to detach transport processes from kinetics, and experiments must be run for long periods of times to reach steady state. More importantly, steady-state data is only controlled by the rate-limiting step at a given condition, overshadowing the detailed description of other elementary steps that may be relevant under other operational scenarios~\cite{Wei2004IsotopicCatalysts,DeDeken1982STEAMDESIGN.,Xu1989MethaneKinetics,Numaguchi1988INTRINSICREFORMING,Rangarajan2012Language-orientedRING}. 

In contrast, dynamic or transient techniques to acquire kinetic data have been applied for nearly half a century~\cite{Kobayashi1972ApplicationDioxide,Kobayashi1972ApplicationState,Kobayashi1972ApplicationTemperatures,Biloen1983TransientMethods,Bennett1976TheCatalysis}, and have recently received renewed interest due to advances in numerical methods and computational processing capabilities. Transient kinetics methods such as temporal analysis of products~\cite{Gleaves1988TemporalResolution,Gleaves2010TemporalCatalysts,Morgan2017FortyProducts} (TAP) associated with temperature-programmed reaction spectroscopy~\cite{Madix2006TheSpectroscopy,Reece2019DissectingInformation,Reece2021MovingReactors} (TPRS), step-response experiments, steady-state isotopic transient kinetic analysis~\cite{Happel1978,Ledesma2014RecentTechnique,Shannon1995CharacterizationReaction} (SSITKA), and tapered element oscillating microbalance (TEOM)~\cite{Berger2008DynamicKinetics} are so information-dense that a single series of states, gathered at their respective time points from one experiment, conveys kinetic information that would require numerous steady-state experiments. However, this information is convoluted with various other effects, making it difficult to extract meaningful intrinsic kinetic parameters. The temporal evolution of chemical system states can be mathematically represented as sets of coupled ordinary or partial differential equations encompassing the various state-changing steps, such as the elementary reactions and diffusive processes involved in a chemical reaction. In the past decades intensive effort has been put into constructing ab-initio microkinetic models, but recently there has been an increased focus on the related issue of catalyst optimization, parameter estimation or model selection based on experimental data~\cite{Aghalayam2000ConstructionMechanisms,Aghalayam2000,Rubert-Nason2014,Caruthers2003CatalystExperimentation,Sjoblom2007,Mhadeshwar2003ThermodynamicMechanisms,Yonge2021TAPsolver:Experiments}. Such techniques simultaneously probe multiple elementary steps, potentially enabling more robust top-down microkinetic or ``mesokinetic'' models to be developed from less experimental data.
Notably, both top-down and bottom-up kinetic models are often informed by additional experimental techniques such as operando~\cite{Topse2003DevelopmentsCatalysts} and in-situ  photoelectron (XPS, LEED, TEM) and infrared (FT-IR) spectroscopy, temperature programmed reduction (TRP) and desorption (TPD), as well as numerous other material science characterization techniques including NMR, XAFS and XRD~\cite{Bligaard2016,Toyao2020,Medford2018}. The coupling of these approaches with kinetic models is generally qualitative or semi-quantitative, and there are limited examples where bottom-up and top-down models are quantitatively combined or compared~\cite{Rubert-Nason2014,Bhandari2020CombiningCatalysis,Bhandari2020ReactionModeling,Rangarajan2017Sequential-Optimization-BasedSystems,Herron2016OptimizationDesign}. As the volume and variety of catalysis data increases, it is necessary to explore new frameworks that facilitate the quantitative fusion of information from various sources. One promising approach is the application of optimization techniques to minimize a loss function that contains terms penalizing the deviation between the model and various types of catalytic data\cite{Rubert-Nason2014,Bhandari2020CombiningCatalysis}. 

Optimization of kinetic models requires the model to be evaluated many times, and relies on computing derivatives of the model with respect to the kinetic parameters. To balance model accuracy with the computational cost of successive evaluation of high-fidelity models, surrogate models (SMs), also referred to as reduced-order, meta- or black-box models, have become an alternative functional mapping between model inputs and outputs~\cite{McBride2019OverviewEngineering}. SMs are models of models for which, given data, parameters are estimated from the minimization of some error metric or merit function. The construction of SMs necessarily relies on experimental data or data generated by mathematically complex phenomenological or first-principles models. SMs also require a proper choice of sampling strategies for the input space, since the computational cost of randomly sampling high dimensional spaces increases exponentially due to the ``curse of dimensionality''. Uniform, random, Latin-hypercube, central composite design and quasi-random low-discrepancy are the most commonly adopted strategies for input space exploration~\cite{Forrester2009}, although more advanced strategies have recently been devised~\cite{Garud2017}. Surrogate models can be based on physical models, flexible data-driven models, or a combination thereof. In flexible model architectures, such as neural nets (NNs), model fitting generally depends on the ability to calculate gradients across the multiple layers through the chain-rule in the back-propagation algorithm~\cite{HECHT-NIELSEN1992TheoryNetwork}. However, to avoid the time-consuming process of estimating gradients or building the back-propagation structure, procedural algorithms for automatic or algorithmic differentiation (AD)~\cite{GriewankEvaluatingEdition} are used to evaluate gradients across NN layers.

Several authors have proposed methods that rely on NNs for the solution of scientific problems involving ODEs and PDEs either as approximators for their solutions or as surrogate models. Seminal work regarding the utilization of NNs as approximators for the solution of differential equations was introduced by Lagaris et. al.~\cite{Lagaris1998ArtificialEquations}, which set the groundwork for the recent development of the so called physics-informed neural networks, PINNs, by Karniadakis et al.~\cite{Raissi2019Physics-informedEquations,Raissi2017PhysicsEquations}. In the context of surrogate model construction, other endeavors have demonstrated that particular NN structures, such as residual networks (ResNets), resemble Euler-type discretization through successive time-steps~\cite{Lu2017BeyondEquations,Haber2017StableNetworks,Ruthotto2018DeepEquations}. Subsequently, Neural ODE (NODE) surrogate models have emerged from the analogy between the discrete time-domain in ResNet ODEs at their depth limits, which consists of an application of NNs to ODE that contiguously describes the physical model in its domain independent of discretization~\cite{Chen2018NeuralEquations}, i.e. in the hypothetical case of infinite recursive application of ResNets, where the discretization converges to a continuum. Following works have corroborated that ResNet-type deep NNs converge to Neural ODEs in the deep limit under stochastic gradient descent minimization~\cite{Avelin2020NeuralWeights}. The fundamental difference between PINNs and NODEs lies in their respective mappings. A PINN acts as a surrogate ``approximator'' for an ODE or PDE solution, where NNs constitute the functional mapping between the independent variables and the states and their derivatives, and the physical model provides the phenomenological connection or restriction between states and their respective gradients/derivatives that NNs must satisfy. In NODEs, NNs instead constitute as mapping from states to their gradients/derivatives, in which case it acts as a true surrogate model. While NODEs can replace unknown parts of a differential equation, PINNs are a basis set for the solution of known differential equations. 

In this work, we focus on the utilization of modified PINNs as a Surrogate Approximator (SA) for the solution of microkinetic models (MKMs) under the MFA, which are fully represented by the chemical system's stoichiometric matrix~\cite{Stoltze2000,Gusmao2015ASystems}. The MFA acts as a general surrogate for detailed kinetics, which could include not only the backbone elementary reactions but also diffusive steps. MKMs consist of a set of elementary or irreducible reactions. The elementary reactions involve the interaction of at most two individual species and are represented by power-law kinetics, in which the probability of occurrence is proportional to their concentrations, and whose probability of success is given by the Arrhenius law. We use this general mathematical representation to approach the solution of known MKMs and address the issue of parameter fitting for kinetically relevant steps in complete MKMs. 

By relying on PINNs, we resort to the universal approximation theorem for artificial neural networks (NNs)~\cite{Cybenko1989,Hornik1990,Hornik1991,Leshno1993} to propose strategies towards the design of SAs that can be utilized for solving kinetics ODEs, where differential algebraic equations (DAE) constraints must be satisfied. We introduce a framework for the classification of reaction network types according to the nature of involved elementary reactions (e.g. homogeneous systems, ad/desorption, reactions between surface intermediates). We further create reaction-type examples and apply our PINN-based approach, demonstrating the need for semi-quantitative information on intermediate composition to fully recover ground-truth information on the MKM parameters. The results provide new tools for analysis of transient kinetic data and insight into top-down analysis of complex reaction networks. The framework has the potential to enable improved design of transient kinetic experiments and fusion of multi-modal transient experimental data in the future.


\section{Methods}

\subsection{Mean-Field Chemical Kinetics Framework}\label{subsec:mf_mkm_fmwrk}
 Power-law kinetic models encompass a linear combination of chemical transformation events, whose sampling frequency scales linearly with the concentration of each participant, and whose probability of success is given by the Arrhenius law. The resulting set of intertwined molecular events is represented as a stack of ODEs, assigning the rate of change of each species to the power-law kinetic expression arising from stoichiometry. Let $\mathbf{c}$ represent an array of concentrations or  concentration-related state variables like partial pressures, concentrations and coverage fractions, for species that are unbound (like gases) or bound (adsorbed molecules or radicals), at time $t$. Then their rates of change, $\mathbf{\dot{c}}$, or the MKM, is written as \eqref{eqn:kin_ode}:
\begin{equation}
    \mathbf{\dot{c}}= \mathbf{M}\,\mathbf{r}(\mathbf{c},\theta)= \mathbf{M}\left(\mathbf{k}(\theta)\circ f(\mathbf{c})\right)\label{eqn:kin_ode}
\end{equation}
Where $\mathbf{M}\in\mathbb{Z}^{n\times m}$ is the corresponding stoichiometry matrix and $~{f(\cdot):\mathbb{R}^n_+\to\mathbb{R}^m_+}$ maps the concentrations $~{\mathbf{c}:=\{\mathbf{c}\,|\,\mathbf{c}\in\mathbb{R}^n_+\}}$ to the concentration-based terms of power-law kinetics, and $~{\mathbf{k}:=\{\mathbf{k}(\theta)~\in~\mathbb{R}^m_+,\,\theta\in\mathbb{R}_{+}\}}$ is the temperature- and binding-energy-dependent, Arrhenius-like rate constant term. In this work, $\theta$ corresponds to temperature (rather than coverage) to keep a consistent standard of bold uppercase variables corresponding to matrices, normal uppercase variables corresponding to elements of a matrix, bold lowercase variables corresponding to vectors, and normal lowercase variables corresponding to scalars.

\subsubsection{Complex Reaction Networks - \textit{gdacs} representation}
We denote the subset of $\mathbf{c}$ for gas-phase species as $\mathbf{c_g} := \{{c}_i\,|\, i\in\mathcal{C}_g\}$, for bounded gas-counterpart coverage fractions as $\mathbf{c_a}:=\{c_i\,|\,i\in\mathcal{C}_a\}$, and for bounded intermediates/radicals on the catalyst surface as $\mathbf{c_s}:=\{c_i\,|\,i\in\mathcal{C}_s\}$ (the distinction between $a$ and $s$ is whether or not the adsorbed species is also sufficiently stable in the gas phase to be detected). Thus $\cup_{i\in\{\mathbf{g},\mathbf{a},\mathbf{s}\}}\mathcal{C}_i=\{1,2,...,m\,|\,m\in\mathbb{N}\}$ and $\cap_{i\in\{\mathbf{g},\mathbf{a},\mathbf{s}\}}\mathcal{C}_i=\emptyset$ (i.e. each species can only be in a single subset, and all species must be assigned to a subset). The state-dependent reaction rate vector, denoted as $\mathbf{r}(\mathbf{c},\theta):=\{\mathbf{r}\in\mathbb{R}^m\,|\,\mathbf{r}={k(\theta)}\circ f(\mathbf{c})\}$, comprises rates of reaction (frequencies) associated with different kinds of elementary reactions: reactions in homogeneous phase, $\mathbf{r_g} := \{{r}_i\,|\, i\in\mathcal{G}_g\}$, those involving adsorption-desorption, $\mathbf{r_d} := \{{r}_i\,|\, i\in\mathcal{G}_d\}$, reactions between adsorbed molecules, $\mathbf{r_a} := \{{r}_i\,|\, i\in\mathcal{G}_a\}$, reactions involving adsorbed molecules and radicals/intermediates on the surface, $\mathbf{r_c} := \{{r}_i\,|\, i\in\mathcal{G}_c\}$, and reactions between intermediates on the surface, $\mathbf{r_s} := \{{r}_i\,|\, i\in\mathcal{G}_s\}$, such that $\cup_{i\in\{\mathbf{g},\mathbf{d},\mathbf{a},\mathbf{c},\mathbf{s}\}}\mathcal{G}_i=\{1,2,...,n\,|\,n\in\mathbb{N}\}$ and $\cap_{i\in\{\mathbf{g},\mathbf{d},\mathbf{a},\mathbf{c},\mathbf{s}\}}\mathcal{G}_i=\emptyset$. 

A full stoichiometry matrix in \eqref{eqn:stoich_full} embodies different types of reaction-type submatrices, where ~{$\{\mathbf{i}\}\subset\{\mathbf{g},\mathbf{d},\mathbf{a},\mathbf{c},\mathbf{s}\}$} are the corresponding reaction types $\mathbf{i}$ as columns of $\mathbf{M}$. One can, hence, classify the submatrices in the full stoichiometry matrix in terms of reaction- and species-types, as summarized in \cref{tab:stoich_explain}.

\begin{table}[hbt!]
    \centering
    \caption{Stoichiometry Matrix Depth Connectivity - Reaction and Species Types}
    \label{tab:stoich_explain}
\begin{tabular}{lccccc}
\toprule
\begin{tabular}{@{}l@{}}{Species Types} \\{}\\{Reaction Types}\end{tabular}&{}&\begin{tabular}{@{}c@{}}\textit{Gas} \\\textit{phase}\\\textbf{g}\end{tabular}&\begin{tabular}{@{}c@{}}\textit{Adsorbed} \\\textit{Molecules}\\\textbf{a}\end{tabular}&\begin{tabular}{@{}c@{}}\textit{Surface} \\\textit{Radicals}\\\textbf{s}\end{tabular}\\
\midrule
\textit{Homogeneous}&\textbf{g}&$\mathbf{M_{gg}}$&-&-\\\\[-2.025ex]
\textit{Ad/desroption}&\textbf{d}&$\mathbf{M_{gd}}$&$\mathbf{M_{ad}}$&-\\\\[-2.025ex]
\textit{Between Adsorbed Molecules}&\textbf{a}&-&$\mathbf{M_{aa}}$&-\\\\[-2.025ex]
\textit{Between Adsorbed Molecules and Radicals}&\textbf{c}&-&$\mathbf{M_{ac}}$&$\mathbf{M_{sc}}$\\
\textit{Between Radicals}&\textbf{s}&-&-&$\mathbf{M_{ss}}$\\
\bottomrule
\end{tabular}
\end{table}

The stoichiometry matrix is then a composition of reaction-type submatrices, as follows in \eqref{eqn:stoich_full}.
\begin{align}
    \begin{bmatrix} \mathbf{\dot{c}_g}\\\mathbf{\dot{c}_a}\\\mathbf{\dot{c}_s}\end{bmatrix}=\begin{bmatrix} \mathbf{M_{gg}}&\mathbf{{M_{gd}}}&\mathbf{0}&\mathbf{0}&\mathbf{0}\\\mathbf{0}&\mathbf{{M_{ad}}}&\mathbf{{M_{aa}}}&\mathbf{{M_{ac}}}&\mathbf{0}\\\mathbf{0}&\mathbf{0}&\mathbf{0}&\mathbf{M_{sc}}&\mathbf{M_{ss}}\end{bmatrix}\mathbf{r}(\mathbf{c},\theta)
    \label{eqn:stoich_full}
\end{align}

Complex reaction paths with multiple reacting surface intermediates may yield matrices of type \textbf{gdacs}, encompassing all possible reaction types. These networks may be seen under circumstances where alternate reaction pathways are possible depending on operating conditions, or when multiple possible pathways are being simultaneously considered in the solution of an inverse kinetics problem. Hence, multiple potential mechanisms may be regarded as feasible, and their parameters estimated together for the elucidation of the prevailing pathways. 

\begin{figure}[ht!]
    \centering
    \includegraphics[trim=3.73in 3.71in 3.73in 3.71in,keepaspectratio=true,scale=1.,width=6.in,clip=True]{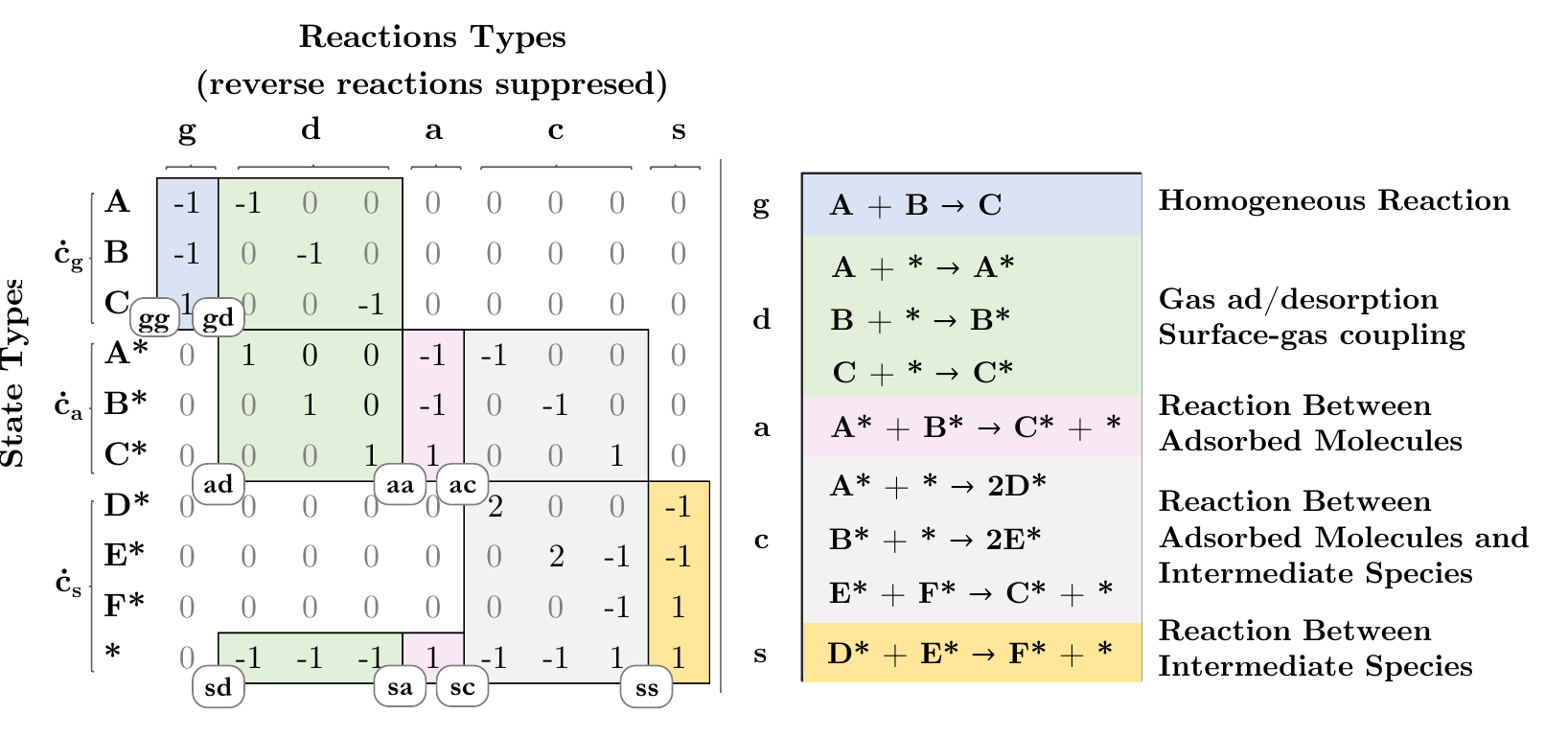}
    \caption{Complex Reaction Network Stoichiometry Matrix Example.}\label{fig:nx_types}
\end{figure}
For instance, combining the \textit{g}, \textit{da} and \textit{dcs} reaction types listed above yields the matrix in Fig. \ref{fig:nx_types} (inverse reactions are suppressed). The \textit{gdacs} reaction network thus represents the maximum variety and depth in complexity, involving all types of species, either directly measurable through analytical methods or mathematically inferred from underlying material balances.  This framework allows for the classification of types of elementary reactions based solely on the chemical reaction stoichiometry matrix. In \cref{sec:results}, we build anecdotal examples of reaction network subtypes to guide the discussion on the need for operando analyses for the data-driven estimation of kinetic parameters, and serve as scaffolds for the construction of chemical kinetics PINNs. 

\subsubsection{Reaction Network Examples}

\subsubsubsection{\textbf{Reaction Network Type \textit{g}:}}\label{sec:typeg} In the simplest homogeneous case conveyed in \ref{reac:hom1}, species $A$ and $B$ reversibly react to form $C$ in gas phase or in solution. To simplify the problem solution, the rate constants $\mathbf{k}$ are considered state- and spatially-independent, which implies an isothermal concentrated-parameter closed system. Various chemical reactions can be represented in terms of \ref{reac:hom1}, such as dissociation, dimerization, and lumped-kinetics surrogates derived from complex kinetics data. A simple lumped-kinetics that could be represented as in \ref{reac:hom1} would be ethylene dimerization to 1-butene, where $A$ and $B$ would represent ethylene, C$_2$H$_4$ and $C$ would denote 1-butene, C$_4$H$_8$.
\begin{align}
    A+B&\underset{k_{\text{-}1}^g}{\stackrel{k_1^g}{\rightleftharpoons}} C\label{reac:hom1}\tag{\textbf{g}.1}
\end{align}
\subsubsubsection{\textbf{Reaction Network Type \textit{da}:}}\label{sec:typeda} For the \textit{da} example, we transpose the reactive step from the homogeneous phase in \cref{reac:hom1} to the heterogeneous phase. by including the adsorption and desorption elementary steps, \textbf{d}.(1-3). and a surface reaction between adsorbed molecules, \cref{reac:a1}. The set of elementary reactions may be a simple representation of catalyzed dimerization reaction, in the same spirit of \cref{reac:hom1}, except that the reaction occurs in a single step on a catalyst surface.
\begin{align}
    A+*&\underset{k_{\text{-}1}^d}{\stackrel{k_1^d}{\rightleftharpoons}} A*\label{reac:adsA}\tag{\textbf{d}.1}\\
    B+*&\underset{k_{\text{-}2}^d}{\stackrel{k_2^d}{\rightleftharpoons}} B*\label{reac:adsB}\tag{\textbf{d}.2}\\
    C+*&\underset{k_{\text{-}3}^d}{\stackrel{k_3^d}{\rightleftharpoons}} C*\label{reac:adsC}\tag{\textbf{d}.3}\\[-2ex]
    \mathclap{\vspace{-10pt}\rule{5cm}{0.2pt}}\notag \\[-1ex]
    A*+B*&\underset{k_{\text{-}1}^a}{\stackrel{k_1^a}{\rightleftharpoons}} C*\label{reac:a1}\tag{\textbf{a}.1}
\end{align}
\subsubsubsection{\textbf{Reaction Network Type \textit{dc}:}} The \textit{a}-type reaction is replaced by two \textit{d}-type reactions, i.e. reaction involving one intermediate species ($D*$), which does not have a corresponding gas phase species. The \textit{d}-type adsorption and desorption steps, \textbf{d}.(1-3), are suppressed, i.e. implicitly included in the set of elementary reactions comprising the \textit{dc}-example below. An example of a reaction that could be represented overall in such terms would be ethylene epoxidation, where $A$ would represent ethylene, $B$ molecular oxygen, O$_2$, $D$ atomic oxygen, O, and $C$ would represent ethylene epoxide.
\begin{align}
    B*+*&\underset{k_{\text{-}1}^c}{\stackrel{k_1^c}{\rightleftharpoons}} 2D*\tag{\textbf{c}.1}\\
    A*\:+\:D*&\underset{k_{\text{-}2}^c}{\stackrel{k_2^c}{\rightleftharpoons}} C*\:+\:*\tag{\textbf{c}.2}
\end{align}
\subsubsubsection{\textbf{Reaction Network Type \textit{dcs}:}} Elementary reactions between surface intermediates exclusively are embodied by \textit{s}-type reactions. In the following example, $D*$, $E*$ and $F*$ are reaction intermediates with no stable desorbed counterpart. In terms of heterogeneous kinetics complexity depth, \textit{s}-type reactions entail intermediates that cannot be directly measured or inferred from ordinary gas/liquid phase effluent/heads-space analytical chemistry techniques, such as gas/liquid chromatography and mass spectroscopy. Butene cracking followed by hydrogenation to ethane could be represented by such a reaction mechanism, where $A$ would entail butene, $D$ ethylene, $B$ molecular hydrogen, H$_2$, $E$ atomic hydrogen, H, $F$ is the ethyl intermediate, and $C$ ethane.
\begin{align}
A*+*&\underset{k_{\text{-}3}^c}{\stackrel{k_3^c}{\rightleftharpoons}} 2D*\tag{\textbf{c}.3}\\
B*+*&\underset{k_{\text{-}4}^c}{\stackrel{k_4^c}{\rightleftharpoons}} 2E*\tag{\textbf{c}.4}\\
D*\:+\:E*&\underset{k_{\text{-}1}^s}{\stackrel{k_{1}^s}{\rightleftharpoons}} F*\:+\:*\tag{\textbf{s}.1}\\
F*\:+\:E*&\underset{k_{\text{-}5}^c}{\stackrel{k_5^c}{\rightleftharpoons}} C*\:+\:*\tag{\textbf{c}.5}
\end{align}

\subsection{Physics-Informed Neural Networks as Surrogate Approximators}

The universal approximation theorem confers to feed-forward (FF) non-polynomial NNs the ability to approximate any continuous function given an appropriate number of neurons~\cite{Cybenko1989,Hornik1990UniversalNetworks,Hornik1991,Leshno1993}. The FFNN structures have proven to serve as appropriate basis functions for ODEs, which may also be extended to PDEs~\cite{Meade1994TheNetworks,Meade1994SolutionNetworks,Lagaris1998ArtificialEquations}. FFNNs can hence be used to approximate ODE initial value problems associated with chemical kinetics. We further show their use in a collocation framework to solve properly scaled inverse problems, under suitable choice of activation function and NN architecture necessary for the solution of stiff kinetic ODEs.

Let $\parsm=\{\mathbf{W}^j\,|\,\mathbf{W}^j\in\mathbb{R}^{n_j\times m_j},\,j\in\{1,2,...,n_l+1\}\}$ denote a parameter tensor for an $n_l$-layer feed-forward neural network (FFNN). Let $\mathbf{x}\in\mathbb{R}^{n}$ be a state vector and $\mathbf{t}\in\mathbb{R}^{p}$ be collocation points, and $\phi_k(\cdot)\,:\,\mathbb{R}^{m}\to\mathbb{R}^m$ be the activation function associated with layer $j$.  The input-output structure for the FFNN can be represented by the recurrence relation in \eqref{eqn:ff_nn}.
\begin{equation}
\begin{aligned}
    \mathbf{y}_{j}&=\left\{\phi_j\left(\mathbf{W}^{j}\mathbf{y}_{j-1}\right)\,\big{|}\,j\in\{1,2,...,n_l\},\,y_0=t\right\}\\
    \mathbf{x}(t,\parsm)&=\mathbf{y}_{n_l+1}=\mathbf{W}^{n_l+1}\mathbf{y}_{n_l}\label{eqn:ff_nn}
\end{aligned}
\end{equation}
We refer to the general use of neural-networks methods as surrogate approximators to solve kinetic ODEs as {\kUDE}s, kinetics-informed neural networks, using PINNs as a scaffold with suitable modifications to enforce structural initial conditions (ICs) or other phenomenological constraints associated with the kinetics of heterogeneous catalytic systems. The general first-order non-homogeneous coupled ODE implicit representation of an MKM in terms of the \kUDE~is shown in \eqref{eqn:ode_nn}, which can be generalized to $p$\textsuperscript{th} order if $\mathbf{x}$ is of class $\mathcal{C}^p$ with respect to $t$.
\begin{equation}
    g[t,\mathbf{x},\mathbf{\dot{x}}]=g(t,\mathbf{x}(t,\parsm),\mathbf{\dot{x}}(t,\parsm))=g(t,\parsm)=0\label{eqn:ode_nn}
\end{equation}
Where $\mathbf{x}$ is the output of the SA encompassing one or more NNs and suitable output transformation layer, and $\mathbf{\dot{x}}$ is evaluated through AD of the former with respect to the single-input input layer. The degrees of freedom in the algebraic equation are all contained in the SA parameters, $\parsm$, which therefore determines the solution of \eqref{eqn:ode_nn} given proper ICs. The time derivative of the SA, $\mathbf{\dot{x}}$, is evaluated with respect to the single-input input layer through AD coupled with just-in-time (JIT) compilation through the JAX library~\cite{47008,jax2018github}.The objective function of the neural network is trained with the \emph{Adam} optimization algorithm. The MKM implicit ODE can be represented in terms of \eqref{eqn:ode_nn} as shown in \eqref{eqn:ode_nn_mkm}.
\begin{equation}
    \erresp_{\dxvec_i}=g(t_i,\parsm)=\dxvec(t_i,\parsm)- \mathbf{M}\left(\mathbf{k}(\theta)\circ \fmap(\xvec(t_i,\parsm))\right)\label{eqn:ode_nn_mkm}
\end{equation}
The implicit form, $g$, and its short-hand $\erresp_{\dxvec_i}$ are indiscriminately utilized in the following sections to denote residuals of the SA-based formulation to be minimized over collocation point in integration domains. In addition, fitting residuals associated with a particular data-point $i$ are denoted by $\erresp_{\xvec_i}$ as follows in \cref{eqn:obs_res}, where $\tilde{\xvec}_i$ are measured state values at $t_i$.
\begin{align}
    \erresp_{\xvec_i}(\parsm) &= \xvec(t_i,\parsm)-\tilde{\xvec}_i\label{eqn:obs_res}
\end{align}
In the discussions that follow in the context of the utilization of FFNN as basis for the solution of kinetic ODE, we generally denote by $\mathbf{x}$, the outputs of the FFNN that approximate the ground-truth solution of the related ODE. Although other state variables, such as temperature and pressure, could in principle also be represented in $\mathbf{x}$, for the isothermal-isobaric cases considered here
\begin{equation*}
    \mathbf{x}\approx\mathbf{c}.
\end{equation*}

\subsubsection{Structural Boundary Conditions}\label{subsec:bcN}

Structural Dirichlet {\bc}s are implemented by designing operators that impose certain behavior onto the NN in regions or points in the integration domain. In general, an operator $\mathbf{C_{bc}^{D}}[\cdot]$ must be devised to enforce Dirichlet-type initial conditions, i.e. $\mathbf{x}(\parsm,t_0)=\mathbf{x}_0~\forall~\parsm$. A natural choice to satisfy a Dirichlet {\bc} is the hyperbolic tangent function $\opfunbc(t)=\tanh(\kappa t)\in \mathcal{C}^1$, such that $\tanh(0)=0$ and $\partial_t\tanh(0)=1$. Here $\kappa\in\mathbb{R}_+$ is a characteristic time constant that must be determined for each chemical system or also modelled in terms of independent variables, as shown in \cref{sec:kappa}. Here we utilize $\mathbf{\hat{x}}$ and $\mathbf{x}[\mathbf{\hat{x}}]$ as the NN and SA outputs, respectively, with parameters $\parsm$ and independent variable $t$ . The {\bc} operator is defined as $\mathbf{C_{bc}^D}[\mathbf{\hat{x}}(\parsm,t),\mathbf{x}_0]=\mathbf{\hat{x}}(\parsm,t)\opfunbc(\kappa(t-t_0))+\mathbf{x}_0\left(1-\opfunbc(\kappa(t-t_0))\right)$.

From simple inspection, the {\bc} is universally enforced in \eqref{eqn:bc0} and gradients are a function of SA parameters in \eqref{eqn:dbc0}.
\begin{align}
\mathbf{\hat{x}}(\parsm,t)\opfunbc(\kappa(t-t_0))|_{t=t_0}+\mathbf{x}_0\left(1-\opfunbc(\kappa(t-t_0))\right)|_{t=t_0}&=\mathbf{\hat{x}}(\parsm,t_0)\opfunbc(0)+\mathbf{x}_0\left(1-\opfunbc(0)\right)=\mathbf{x}_0\label{eqn:bc0}\\
\partial_t~\mathbf{\hat{x}}(\parsm,t)\opfunbc(\kappa(t-t_0))|_{t=t_0}+\partial_t~\mathbf{x}_0\left(1-\opfunbc(\kappa(t-t_0))\right)|_{t=t_0}&=\notag\\{\mathbf{\dot{\hat{x}}}}(\parsm,t_0)&\opfunbc(0)+\mathbf{x}(\parsm,t_0){\partial_t{\dot{\opfunbc}}}(0)-\mathbf{x}_0{\partial_t{\dot{\opfunbc}}}(0)\label{eqn:dbc0}\\
&=\mathbf{\hat{x}}(\parsm,t_0)-\mathbf{x}_0\notag
\end{align}
Such that the SA vanishes at $t=t_0$, \eqref{eqn:bc0}, with the continuous time-derivative of $\mathbf{x}$ equal to the NN output, $\mathbf{\hat{x}}$, \eqref{eqn:dbc0}. Therefore, any Dirichlet {\bc} can be automatically satisfied in this form for {\kUDE}s. The same approach can be extended to Neumann {\bc}s with proper choice of $\opfunbc$ function. The application of the operator $\mathbf{C_{bc}^{D}}$ on MKMs follows naturally by making $\mathbf{x}\,{\approx}\,\mathbf{c}$ in \eqref{eqn:kin_ode} as depicted by \eqref{eqn:ode_nn_mkm}.
\subsubsection{Characteristic Time}\label{sec:kappa}
In the case of  \textit{tanh} as choice of $\phi$ for $\bcD$, a proper choice for characteristic time constant can be circumvented by replacing $\kappa$ by an ancillary shallow NN that can be learned along with the underlying SA NNs, with a monotonically increasing output layer. In this work we adopted an exponential form for $\kappa$ whenever solving the forward ODEs, as in \eqref{eqn:kappa_fun}.
\begin{equation} \label{eqn:kappa_fun}
    \kappa = e^{\mathbf{u}(\parsmt,t-t_0)}
\end{equation}
Where $\mathbf{u}$ is a shallow FFNN with parameters $\parsmt$. The {\bc} operator on $\phi$ and $\kappa$, $\bcD$, can thus be represented in terms of $\opfunbc$ and $\mathbf{u}$ as follows in \eqref{eqn:bcd_tanh}, where parameter dependencies were suppressed. For the sake of simplicity, we further assume $\kappa$ is embedded into the SA structure $\mathbf{x}$ so its weights, $\parsmt$, are included in $\parsm$. 
\begin{equation}\label{eqn:bcd_tanh}
    \mathbf{C_{bc}^D}[\mathbf{\hat{x}}(t),\mathbf{x}_0]=\mathbf{\hat{x}}(t)\tanh\left(e^{\mathbf{u}(t-t_0)}(t-t_0)\right)+\mathbf{x}_0\left(1-\tanh\left(e^{\mathbf{u}(t-t_0)}(t-t_0)\right)\right)
\end{equation}
\subsubsection{Normalization DAE Constraints}\label{subsec:normconstr}
MKMs typically treat the concentration of adsorbed and intermediate species in terms of fractions of the total number of available active sites. It is hence desirable for SAs to structurally enforce normalization, eliminating a degree of freedom. We propose the projection of radius-one hypersphere onto the natural basis to enforce inherent normalization. Let $\mathbf{x}(\parsm,t)\in\mathbb{R}^k$ be the output values of the constrained surrogate related to the neural network $\hat{x}(\parsm,t)\in\mathbb{R}^{k-1}$, such that
\begin{align}\label{eqn:normconstr}
\begin{split}
    {x}_i(\parsm,t)&=\left(1-\sin^{2}\left({\hat{x}}_i(\parsm,t)\right)\right)\prod_{j<i}\sin^{2}\left({\hat{x}}_j(\parsm,t)\right)\;\forall\;i<p;\;{i,j}\in\mathbb{N}\\
    {x}_p(\parsm,t)&=\prod_{j<p}\sin^{2}\left({\hat{x}}_j(\parsm,t)\right)
\end{split}
\end{align}
Such a trigonometric transformation enforces bounds to $\mathbf{x}$, such that $0\le\mathbf{x}\le1$ and $\sum\mathbf{x}=1\;\forall\;\mathbf{\hat{x}}\in\mathbb{R}^{p-1}$.
The SA output can hence be denoted as the result of the normalization-constraint operator $\mathrm{C_N}$ over the NN output $\mathbf{\hat{x}}$, i.e. $\mathbf{x}=\mathrm{C_N}[\mathbf{\hat{x}}]$. The normalization-constraint operator implicitly embeds DAE constraints to the ODE solutions, since it encodes $\sum \dot{\mathbf{x}}=0$ for species coverage. 
\subsection{\kUDE~Forward Problem}

The surrogate approach to solving the forward problem involves training the neural network to satisfy the {\bc}s, as well as the differential form specifying the reaction kinetics in \eqref{eqn:ode_nn}, over a defined time grid. We constrain the analysis to non-spatially dependent closed systems, formulating the problem for an ideal, uniform (perfectly stirred), pseudohomogeneous batch reactor model, where the underlying ODE describing the state-variables evolution over time is analogous to \eqref{eqn:kin_ode}. The approach would also hold for the steady-state solution over space of fixed or packed-bed reactors. The general solution for the forward problem, i.e. solving the ODE, consists of minimizing the implicit form in \eqref{eqn:ode_nn} given consistent {\bc}s for the states, where the real-valued function $g$ corresponds to \eqref{eqn:kin_ode} in standard form given fixed model parameters, $\mathbf{\parmkm}$, which is outlined by \eqref{eqn:ode_nn_mkm}. The initial value is represented by $\mathbf{x_0}$, and the residual is evaluated over the collocation points, $\mathbf{t}$, as shown in \eqref{eqn:ode_nn_opt}. This represents the minimization of the residual $\jm$, under some L$2$-norm, between the Jacobians of the states from the derivative-based rate laws and the surrogate approximator. AD provides the Jacobians, i.e. $\mathbf{\dot{x}}$ and $\partial_{\parsm}\,\jm(t,\parsm)$, for the objective function. The \emph{Adam} optimization algorithm is used to minimize the residual, $\jm$. 
\begin{equation}
\begin{aligned}
\min_{\parsm} \quad \jm(\parsm)&=||g(\mathbf{t},\parsm)||_2^2\\&=\frac{1}{n}\sum_i^n\erresp_{\dxvec_i}^T\erresp_{\dxvec_i} \\
&\textrm{s.t.} \quad  \xvec(t_0,\parsm)=\mathbf{x_0} \quad 
   \label{eqn:ode_nn_opt}
\end{aligned}
\end{equation}
Importantly, in the case of MKMs for which surface coverage fractions are in play, the normalization of elements of $g$ can be structurally enforced as shown in \ref{subsec:normconstr}, by having $g(\mathbf{x}(\mathbf{\hat{x}}))$ as in \eqref{eqn:normconstr}, which imposes a DAE constraint. Furthermore, when dealing with mixed normalized (coverage) and non-normalized (bulk-phase) chemical species, the problem can be split into two separate underlying NNs which are concatenated in the SA output layer: a non-normalized one representing the mapping between time and gas-phase concentrations and a normalized one mapping between time and surface coverages. The implicit representation of MKM \kUDE~is given in \eqref{eqn:ode_nn_mkm}. When structural {\bc}s are applied and normalization constraints are included, the surrogate model consists of the following combinations of underlying NNs and operators, \eqref{eqn:ode_op_c}, using the same nomenclature as defined in \ref{subsec:mf_mkm_fmwrk}.
\begin{align}
    \mathbf{x}&=\bcD\begin{bmatrix}\mathbf{\hat{x}_{g}}\\\opnc[\mathbf{\hat{x}_{as}}]\end{bmatrix}\label{eqn:ode_op_c}
\end{align}
Where $\mathbf{\hat{x}_g}$ corresponds to the output layer of the NN associated with gas (unbound) species and $\mathbf{\hat{x}_{as}}$ encompasses the output layer of subjacent NN that conveys the $p-1$ angular components that, once mapped through the operator $\mathbf{\mathrm{C}_N}$, outputs $p$ normalized coverage fractions. Finally, the initial-value {\bc} operator $\bcD$ is applied onto the combined (concatenated) NNs.
~\begin{algorithm}
\caption{Forward \kUDE}\label{alg:\kUDE_fwd_training}
\begin{algorithmic}[1]
        \State \algorithmicrequire{ Boundary conditions $(t_0,\xvec_0)$, $\parmkm$, maximum iterations $n_{max}$, tolerance $tol$}
        \State{Initialize NN, SA parameters $\parsm$, residual $\jm(\parsm)$}
        \While{iter < $M$ \textbf{or} $\jm(\parsm)$ < $tol$}   
        \State Predict states $\xvec$ from SA over $\mathbf{t}$ 
        \State Predict states derivatives $\dxvec$ from SA over $\mathbf{t}$ 
        \State Compute objective function $\jm(\parsm)$ over $\mathbf{t}$
        \State Compute gradients of $\jm(\parsm)$ with AD; $\partial_{\parsm}\jm(\parsm)$
        \State Update SA parameters $\parsm$ using \emph{Adam}
        \EndWhile
        \State \algorithmicensure{ Concentration profile $\xvec(t,\parsm)$}
\end{algorithmic}
\end{algorithm}

\subsection{\kUDE~Inverse Problem}
Under the same model assumptions of the forward MKM problem, we extend the PINN-based~\kUDE~approach to MKM inverse ODE problems by letting $\tilde{\mathbf{x}}\in\mathbb{R}^{n\times d}_+$ represent observed (measured) concentrations of $n$ chemical species over the respective $d$ time-points $\mathbf{t}\in\mathbb{R}^{d}_{+}$, and including it in the objective function. As in the forward problem, state variables are represented by surrogates from separate NNs and applied operators, as in \eqref{eqn:ode_op_c}, i.e. $\mathbf{x}(t,\parsm)$. The inverse problem consists of the forward {\kUDE} model with the inclusion of interpolation of observed datapoints, $\mathbf{\tilde{x}}$ as a regularization term in the objective function. 
In the regularized setup, the error to be minimized consists of the combination of the residual between the SA and the observed data (interpolation), $\jd$, and that between SA and the kinetic model describing the behavior (physics regularization), $\jm$, in a least-squares sense, similarly to the approach by Raissi \textit{et al.}~\cite{Raissi2019Physics-informedEquations}. As in the forward problem, the combined error is minimized using the \emph{Adam} optimization algorithm. 
\begin{align*}
    \jd(\parsm) &= \frac{1}{n}\sum_i^n\erresp_{\xvec_i}^T\erresp_{\xvec_i}\\
    \jm(\parsm,\parmkm) &=  \frac{1}{n}\sum_i^n\erresp_{\dxvec_i}^T\erresp_{\dxvec_i}
\end{align*}
Finally, let $\parlumped=\{\parsm,\,\parmkm\}$ combine the residual from the forward-\kUDE~as regularization hyperparameter with weight $\alpha$ into the inverse ODE.
\begin{equation}
\begin{aligned}
\min_{\parlumped} \quad &\jt(\parlumped)= \jm(\parsm,\parmkm) + \alpha \jd(\parsm) \\
\textrm{s.t.}\quad&\parmkm\in\mathbb{R}^{\operatorname{dim}(\parmkm)},\;\parsm\in\mathbb{R}^{\operatorname{dim}(\parsm)};\alpha\in\mathbb{R}_{+}
  \label{eqn:ode_nn_inv}
\end{aligned}
\end{equation}
The regularization hyperparameter $\alpha$ conveys the ratio between the variances of the data (states) and model (state derivatives) under the hypothesis of homoscedasticity, and it needs to be optimized given an MKM postulate and the observed data. Furthermore, the regularization intensity plays a decisive role in the case of noisy data, since initial training steps with excessive regularization potentially lead to local convergence far from the global optimum in the model parameter space. A conservative strategy implemented in this work is to perform initial training steps at low values of $\alpha$ and then proceed with increments in $\alpha$ until an inflection point between $\log(\jd)$ and $\log(\jm)$ is observed. In \cref{sec:inv_kinns}, a sensitivity analysis is performed over $\alpha$ values for an anecdotal inverse problem. Also, the learning of model parameters must take place in a similar domain and scale as the SA-associated NN weights. As kinetic parameters have non-negative values and may substantially vary in order of magnitude, we choose to train an exponential mapping, i.e., set $\mathbf{k}=\exp(\parmkm)$. Furthermore, for the \textit{tanh} activation function, a saturation occurs at arguments of $\pm5$, and for \textit{swish} with a similar pattern as \textit{exp}, such that $\parmkm$ and $\parsm$ are of the same order of magnitude. This allows that they be grouped as a lumped variable and trained simultaneously as described in the following pseudo-algorithm.
\begin{algorithm}
\caption{Regularized Inverse \kUDE}\label{alg:kinns_inv_reg}
\begin{algorithmic}[2]
        \State \algorithmicrequire{ Observed data ($\mathbf{t}$, $\tilde{\xvec}$), maximum iterations $n_{max}$, tolerance $tol$}
        \State Build NN and initialize SA parameters $\parsm$, and kinetic model parameters $\parmkm$
        \While{iter < $n_{max}$ \textbf{or} $\jt(\parlumped)$ < $tol$}   
        \State Predict states $\xvec(t,\parsm)$ from SA over $\mathbf{t}$ 
        \State Predict states derivatives $\dxvec(t,\parsm)$ from SA over $\mathbf{t}$  
        \State Compute objective function {$\jt(\parlumped)$} over $\mathbf{t}$
        \State Compute gradients of {$\jt(\parlumped)$} with AD; $\partial_\omega\jt(\parsm,\parmkm)$
        \State Update SA parameters $\parsm$ and kinetic model parameters $\parmkm$ using \emph{Adam}
        \EndWhile
        \State \algorithmicensure{ Kinetic model parameters $\parmkm$}
\end{algorithmic}
\end{algorithm}
Different non-stiff anecdotal kinetic models are solved in the forward sense to justify the use of FFNNs as bases for the solution of the inverse problem. We increase reaction network complexity, moving through (i) an example of a simple homogeneous case (type \textit{g}), (ii) an example involving reaction between adsorbed molecular species (type \textit{da}), (iii) an example including reactions between adsorbed molecules and reaction intermediates (type \textit{dc}) and (iv) an example including a reaction between intermediates (type \textit{dcs}). Abstract species are used in the following examples, which may suit a myriad of actual chemical reactions.
\subsection{Synthetic Data Generation}\label{sec:data_generation}
\begin{figure}[ht!]
    \centering
    \includegraphics[trim=3.73in 2.57in 3.73in 2.55in,keepaspectratio=true,scale=1.,width=6.in,clip=True]{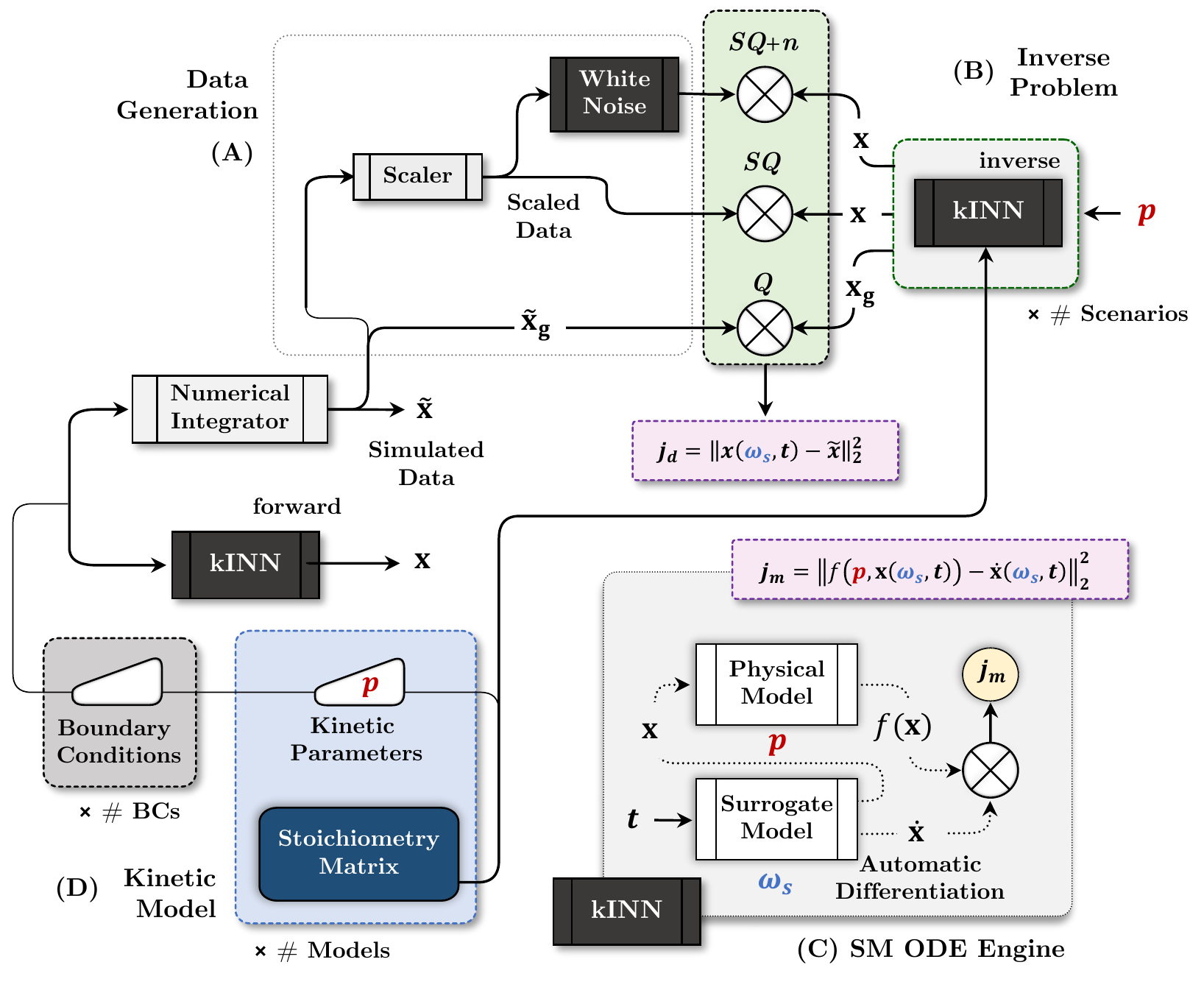}
    \caption{Data generation (A) from stiff-nonstiff algorithm LSODA for three case studies: quantitative data for gas phase only (\obsonly), quantitative gas phase and scaled semiquantitative coverage fraction without (\semiquant) and with added noise (\semiquantn). Forward (D, \kUDE~block) and inverse (B) {\kUDE}s scheme. {\kUDE} block (C) and kinetic model structure and inputs (D).}\label{fig:kinns_scheme}
\end{figure}

The scheme for solving the forward {\kUDE}s problem, and the inverse problem in three different scenarios, is shown in \cref{fig:kinns_scheme}. The scenarios, which vary in the completeness of knowledge about the chemical systems and the presence of noise representing actual measurements, as listed below.
\begin{itemize}
    \item \textbf{Quantitative Unbound Species (\obsonly):} no information about chemical intermediates and adsorbed molecules; the only measured states are related to unbound chemical species; e.g. gas, liquid species that can be measured downstream (flow-type reactors) or sampled from the reactor headspace (batch) with standard analytical equipment: gas/liquid chromatography, mass spectroscopy. For the \textit{\obsonly} scenarios, only data generated for the observable variables ie. unbound chemical species are fed to the training loop, such that $\jd$, or $\erresp_\xvec$, has the length of the observable variables. Meanwhile, $\jm$, or $\erresp_\dxvec$, in \eqref{eqn:ode_nn_inv}, encompasses derivatives for all present (or assumed present) chemical species and intermediates that are measured and unmeasured. 
    \item \textbf{\obsonly ~+ Semiquantitative Bound Species (\semiquant):} it is assumed that deconvolved {\semiquant} data can be obtained, e.g. with operando analytical techniques, for all adsorbed molecules and stable reaction intermediates, and that the {\semiquant}s are linearly related to the concentration of such species. For the {\semiquant} scenario, the surface coverages generated from the numerical solution of the forward problem are rescaled by dividing the time series values by their standard deviation (an effectively arbitrary number that mimics a calibration factor in spectroscopy). In the inverse {\kUDE}s problem, the scaling factors between fractional coverages and the \semiquant{}  intensities are also included as parameters to be learned in $\parsm$. 
    \item \textbf{\semiquant~+ noise (\semiquantn):} a normally distributed homoscedastic error (white noise) with an arbitrary  standard deviation of $0.025$ is added to the scaled data for the {\semiquant} scenarios. This setup intends to represent a more realistic scenario that evaluates the robustness of the physics regularized setup in the presence of noise. 
\end{itemize}

To obtain the ground truth set for the comparison of the forward {\kUDE}~results, the ensuing ODEs of all combinations of MKMs and {\bc}s were solved numerically with the stiff-nonstiff algorithm LSODA from the FORTRAN ODEPACK~\cite{Hindmarsh1980LSODESolvers,Petzold1983AutomaticEquations}. Similarly, the generation of abstract observed data and further per-scenario processing was carried out using the same numerical method, as shown in \cref{fig:kinns_scheme} (A). For the solution of forward {\kUDE}s, the MKM stoichiometry matrix, the kinetic parameters, $\parmkm$, and {\bc}s is provided, \cref{fig:kinns_scheme} (D). Conversely, for the inverse problem, $\parmkm$ is adjusted and synthetic data is generated and provided as if experimentally assessed, \cref{fig:kinns_scheme} (B). The {\kUDE} SA-based solution flow diagram is portrayed in \cref{fig:kinns_scheme} (C), exhibiting the interplay between the SA and the physical models, and the final cost function, $\jm$.
\begin{table}[hbt!]
    \centering
    \caption{Rate Constants and NN architecture}
    \label{tab:fwd_archx}
\begin{tabular}{lccccccccccclc}
\toprule
{}&\multicolumn{11}{c}{Rate Constants}&\multicolumn{2}{c}{Architecture (layers)}\\\cline{2-14}\\[-2.05ex]
Type&$\frac{k_{1}^g}{k_{\text{-}1}^g}$&$\frac{k_{1}^d}{k_{\text{-}1}^d}$&$\frac{k_{2}^d}{k_{\text{-}2}^d}$&$\frac{k_{3}^d}{k_{\text{-}3}^d}$&$\frac{k_{1}^a}{k_{\text{-}1}^a}$&$\frac{k_{1}^c}{k_{\text{-}1}^c}$&$\frac{k_{2}^c}{k_{\text{-}2}^c}$&$\frac{k_{3}^c}{k_{\text{-}3}^c}$&$\frac{k_{4}^c}{k_{\text{-}4}^c}$&$\frac{k_{5}^c}{k_{\text{-}5}^c}$&$\frac{k_{1}^s}{k_{\text{-}1}^s}$&\begin{tabular}{@{}l@{}}Hidden \\\footnotesize{[\textit{tanh},${swish}$,\textit{tanh}]}\end{tabular}& \begin{tabular}{@{}l@{}}Output \\\footnotesize{[linear,$\mathbf{\opnc[\cdot]}$]}\end{tabular}\\
\midrule
\textit{g}&$\frac{10}{1}$&-&-&-&-&-&-&-&-&-&-&[5, 5, 5]&[3,$\text{-}$]\\\\[-2.025ex]
\textit{da}&-&$\frac{10}{4}$&$\frac{40}{60}$&$\frac{200}{40}$&$\frac{100}{80}$&-&-&-&-&-&-&[12, 12, 12] & [3,4]\\\\[-2.025ex]
\textit{dc}&-&$\frac{20}{8}$&$\frac{16}{4}$&$\frac{12}{8}$&-&$\frac{1200}{400}$&$\frac{2000}{1600}$&-&-&-&-&[16, 16, 16]& [3,5]\\\\[-2.025ex]
\textit{dcs}&-&$\frac{20}{8}$&$\frac{24}{12}$&$\frac{16}{40}$&-&-&-&$\frac{640}{960}$&$\frac{160}{80}$&$\frac{560}{160}$&$\frac{640}{240}$&[20, 20, 20]& [3,7]\\
\bottomrule
\end{tabular}
\end{table}

The cost function to be minimized for learning the solution of the ODEs using {\kUDE}s was defined in terms of mean-square error, MSE. The JAX library was utilized for the forward AD, allowing the obtainment of the gradients with respect to the SAs' parameters, which were optimized with the \emph{Adam} algorithm~\cite{Kingma2015Adam:Optimization}. Since {\bc}s are structurally enforced by $\bcD$, the ODE solutions are found by minimizing $\jd$, i.e. finding $\parsm$ such that the state derivatives from the physical model given SA state estimates, $\mathbf{x}$, match those estimated by forward AD of the SAs with respect to their single value time input. All code necessary for reproducing the results of this work are provided in the \SM{} and via GitHub (\url{https://github.com/gusmaogabriels/kinn/tree/paper_reg}).
\section{Results and Discussions}\label{sec:results}

\subsection{Forward {\kUDE}s}\label{sec:fwd_results}

Three-hidden-layer NNs with \textit{tanh}, \textit{swish} and \textit{tanh} activation functions respectively were utilized as basis for the solution of forward {\kUDE}s. For the overall reversible reaction $A\rightleftharpoons B+C$ represented by each reaction network type, several anecdotal examples for the forward problem were solved given a set of diverse \textit{Dirichlet} {\bc}s in terms of $x_A$, $x_B$ and $x_C$ at $t=0$, and sets of rate constants chosen to portray different degrees of stiffness, i.e. faster rates for surface reactions, as in \cref{tab:fwd_archx}. For all reaction types, hyperbolic tangent \textit{Dirichlet}-type boundary-condition operators, $\bcD$, were applied to SAs, as in \cref{subsec:bcN}. Ancillary single-hidden-layer FFNNs with three (\textit{a}- and \textit{da}-type) or six (\textit{dc} and \textit{dcs}-type) neurons and \textit{swish} activation functions were trained in parallel for each forward problem to allow for a time-dependent characteristic time or gain for the {\bc} operator, which provides a learning gain-layer in the case of varying stiffness, as defined in \cref{sec:kappa}. The forward {\kUDE}s are defined in terms of their associated reaction network (stoichiometry matrix), {\bc}s and kinetic parameters. Normalization constraints are not included in the forward {\kUDE}s since they bring additional non-linearity that is unnecessary given that the problems are well-defined by their ODEs and {\bc}s.

\begin{figure}[ht!]
    \centering
    \includegraphics[trim=0in 0in 0in 0in, keepaspectratio=true,scale=1.,clip=True]{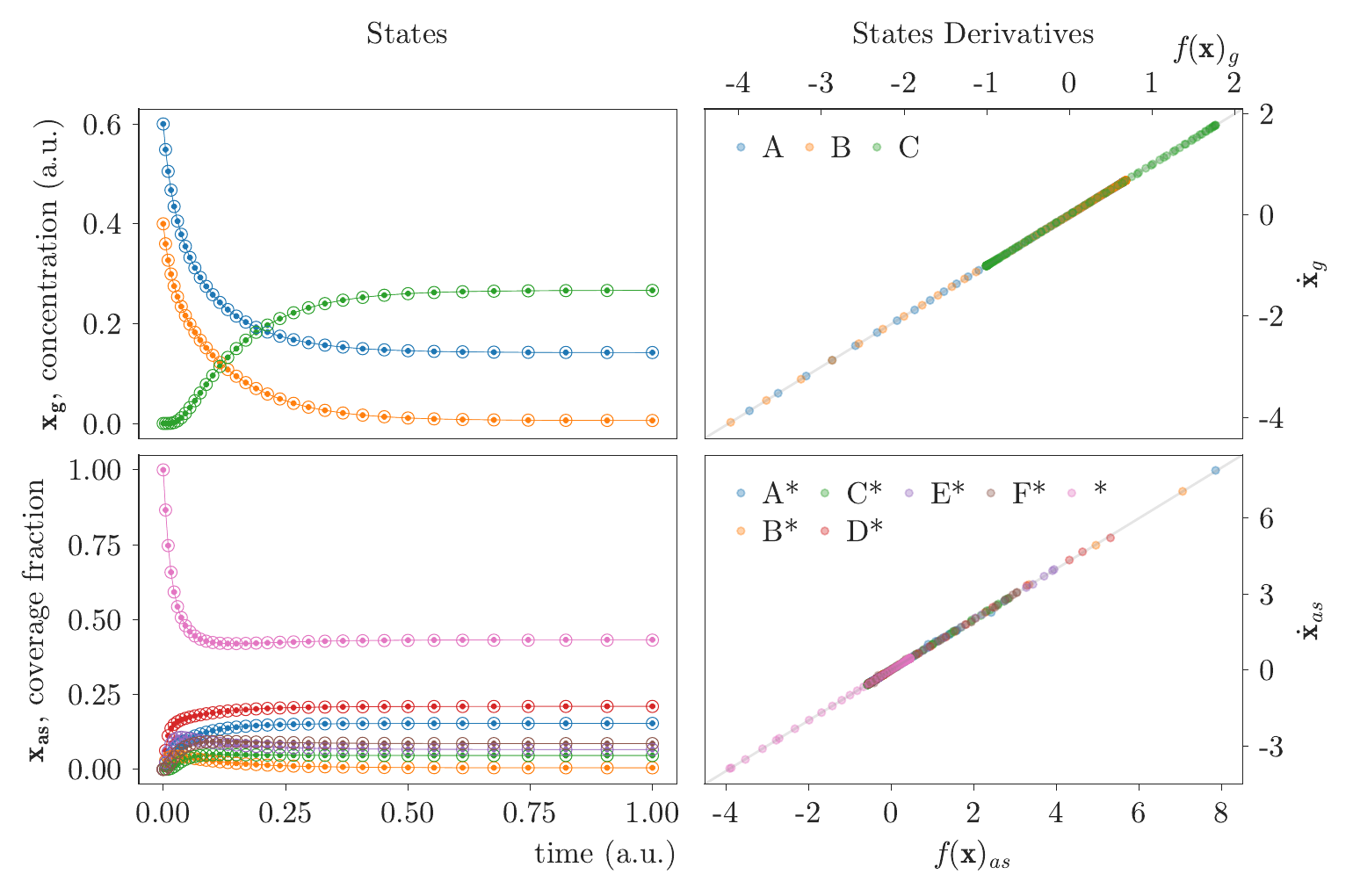}
    \caption{Reaction network type \textit{dcs} forward solution with {\bc} of $\{x_A,x_B\}_{t_0}=\{0.6,0.4\}$ and clean surface (IC 1). States solution (top) (SA, open circles; numerical solution, closed markers) under minimization of \eqref{eqn:ode_nn_opt} (bottom) for observable variables (unbound species, top) and latent variables (coverages, bottom).}\label{fig:fwd_dcs_bc0}
\end{figure}

\cref{tab:fwd_results} allows the comparison of the numerical results with the SA derived for the forward {\kUDE}s. The set of {\bc}s was defined so as to represent the two scenarios: one in which there are similar concentrations of both reactants, (IC 1, $\{x_A,x_B,x_C,x_*\}_{t_0}=\{0.6,0.4,0.0,1.0\}$), and another to represent the case where considerable amount of product is present, (IC 2, $\{x_A,x_B,x_C,x_*\}_{t_0}=\{0.2,0.3,0.5,1.0\}$). The two conditions were chosen to represent chemical transformations that predominantly run in the forward or reverse reaction, and thus inform further attempts to solve the related inverse problems. The performance metrics are defined as a function of the collocation points about which the SA is trained, which is exemplified in \cref{fig:fwd_dcs_bc0} by the \textit{dcs}-reaction network type solution for {\bc} 1. For all scenarios, about 100 collocation points were sampled in a logarithm space in the time domain, such that there is higher probability that points be placed in regions of higher derivatives near the {\bc}. To refine the solutions, SAs were trained in three stages with varying training step size from $10^{-4}$ to $10^{-6}$, over a total of $10^3$ epochs, and $10^2$ iterations per epoch. The convergence criteria was that either $\jt<10^{-12}$ or that the maximum number of epochs was reached. Evidently, absolute error metrics such as MAE and MSE cannot be compared between different MKMs and {\bc}s, since different ODEs may exhibit different stiffness and therefore larger or smaller derivative values in different scales as compared to actual state values. Irrespective of MKM and {\bc}s, the coefficient of determination ($r^2$) for all examples is unity (within round-off error). Such results support the hypothesis that NNs can be utilized as general bases for the solution of forward {\kUDE}s, which is the underpinning assumption for their utilization in the solution of inverse {\kUDE}s.

In \cref{fig:fwd_dcs_bc0}, the final solution for the type-\textit{dcs} MKM under the reactant-only {\bc} (IC 1) is shown. The predicted states (open circles) line up over the numerical solution (solid circles) for both the observable concentrations $\mathbf{x_g}$ (top) and coverage fractions $\mathbf{x_{as}}$ (bottom). The minimization of the underlying forward {\kUDE}s cost function $\jm$ can be inferred by the parity plot between physical (kinetic) model derivatives from estimated states, and their derivative estimates from forward AD. For the remaining cases, the standardized derivative parity plots for the observed and latent states can be compared with the coefficient of determination reported in \cref{tab:fwd_results}, for which coefficients of determinations are computed as the average squared correlation coefficient over different states.

\begin{table}[hbt!]
    \centering
    \caption{Forward {\kUDE}s Results Summary - Performance Metrics (vs Ground Truth)}
    \label{tab:fwd_results}

\begin{tabular}{llrrrrrr}
\toprule
    &    & \multicolumn{3}{c}{Observed States Derivatives} & \multicolumn{3}{c}{Latent States Derivatives } \\ 
 Type & {\bc}     &r$^2$ & MAE & MSE & r$^2$ & MAE & MSE \\
\midrule
\textit{g} & 1 &     1.00 &  5.50 × 10$^{-4}$ &   3.92 × 10$^{-7}$ &           - &                 - &                 - \\
& 2 &     1.00 &  1.79 × 10$^{-5}$ &  4.27 × 10$^{-10}$ &           - &                 - &                 - \\\hline\\[-1.95ex]
\textit{da} & 1 &     1.00 &  2.51 × 10$^{-3}$ &   1.12 × 10$^{-5}$ &        1.00 &  2.97 × 10$^{-3}$ &  1.84 × 10$^{-5}$ \\
& 2 &     1.00 &  8.93 × 10$^{-3}$ &   2.49 × 10$^{-4}$ &        1.00 &  1.73 × 10$^{-2}$ &  1.32 × 10$^{-3}$ \\\hline\\[-1.95ex]
\textit{dc} & 1 &     1.00 &  3.30 × 10$^{-3}$ &   2.09 × 10$^{-5}$ &       0.999 &  1.26 × 10$^{-2}$ &  4.70 × 10$^{-4}$ \\
& 2 &     1.00 &  2.78 × 10$^{-3}$ &   1.68 × 10$^{-5}$ &        1.00 &  7.38 × 10$^{-3}$ &  1.42 × 10$^{-4}$ \\\hline\\[-1.95ex]
\textit{dcs} & 1 &     1.00 &  2.18 × 10$^{-3}$ &   9.87 × 10$^{-6}$ &        1.00 &  7.13 × 10$^{-3}$ &  4.00 × 10$^{-4}$ \\
& 2 &     1.00 &  6.89 × 10$^{-4}$ &   8.45 × 10$^{-7}$ &        1.00 &  3.33 × 10$^{-3}$ &  3.58 × 10$^{-5}$ \\
\bottomrule
\end{tabular}
\end{table}

\subsection{Inverse {\kUDE}s}\label{sec:inv_kinns}

~{The forward numerical solution data for the different reaction network types were utilized as the ground truth data for the inverse {\kUDE}~studies. We define three scenarios or categories for the inverse {\kUDE}s involving heterogeneous models, i.e. reaction networks other than type-\textit{g} only, as portrayed by scheme (A) in \cref{fig:kinns_scheme}. Unlike the forward {\kUDE}s, whose solutions for the two {\bc}s are independent of each other, the inverse {\kUDE}s are interdependent since they share, and thus act on, the same kinetic model and its kinetic parameters encoded in $\parmkm$. The solution of inverse {\kUDE}s can involve multiple experimental (observed) data to which independent SAs are trained under a single underlying physical model. In this section, all inverse {\kUDE} estimated model parameters were obtained by the simultaneous interpolation of IC 1 and 2 data from the numerical solution of the MKM ODEs.

We start by analyzing the {\semiquant} and {\semiquantn} scenarios, since these scenarios contain the most information about the system. The normalization constraint operator, $\opnc$, plays a major role in the learning of scaling factors for the {\semiquant} scenarios, since it constrains the scaling factors to describe normalized (fractional) coverages, and indirectly imposes a DAE constraint to ODE solution. Importantly, the regularization parameter, $\alpha$, \eqref{eqn:ode_nn_inv} must be determined for each specific reaction network type and associated data (as in any method where hyperparameters need be optimized). The optimal $\alpha$ range can be inferred by simultaneously minimizing the residual terms with respect to data, $\jd$, and to the physical model, $\jm$. Since this involves a trade-off between the two parts of the objective function, there will be a Pareto frontier of optimal models as a function of $\alpha$. We utilized sensitivity analyses on the residual terms for the anecdotal \textit{dcs}-type reaction network scenarios to estimate their Pareto frontiers. We successively increased $\alpha$ from $10^{-6}$ to $10^{6}$ and further decreased it back to $10^{-6}$ to assess the non-dominated points (i.e. Pareto-optimal set of points with optimal trade-off between the two objectives) in the estimated Pareto sets for the {\semiquant} and \textit{\semiquantn} scenarios. For each value of $\alpha$, the same SA learning strategy in terms of convergence criteria, number of epochs, iteration per epoch and solution refinement as defined in \cref{sec:fwd_results} was adopted.
}~
~{\begin{figure}[ht!]
    \centering
    \includegraphics[trim=3.96in 3.43in 3.96in 3.43in,keepaspectratio=true,scale=1.,width=6.13in,clip=True]{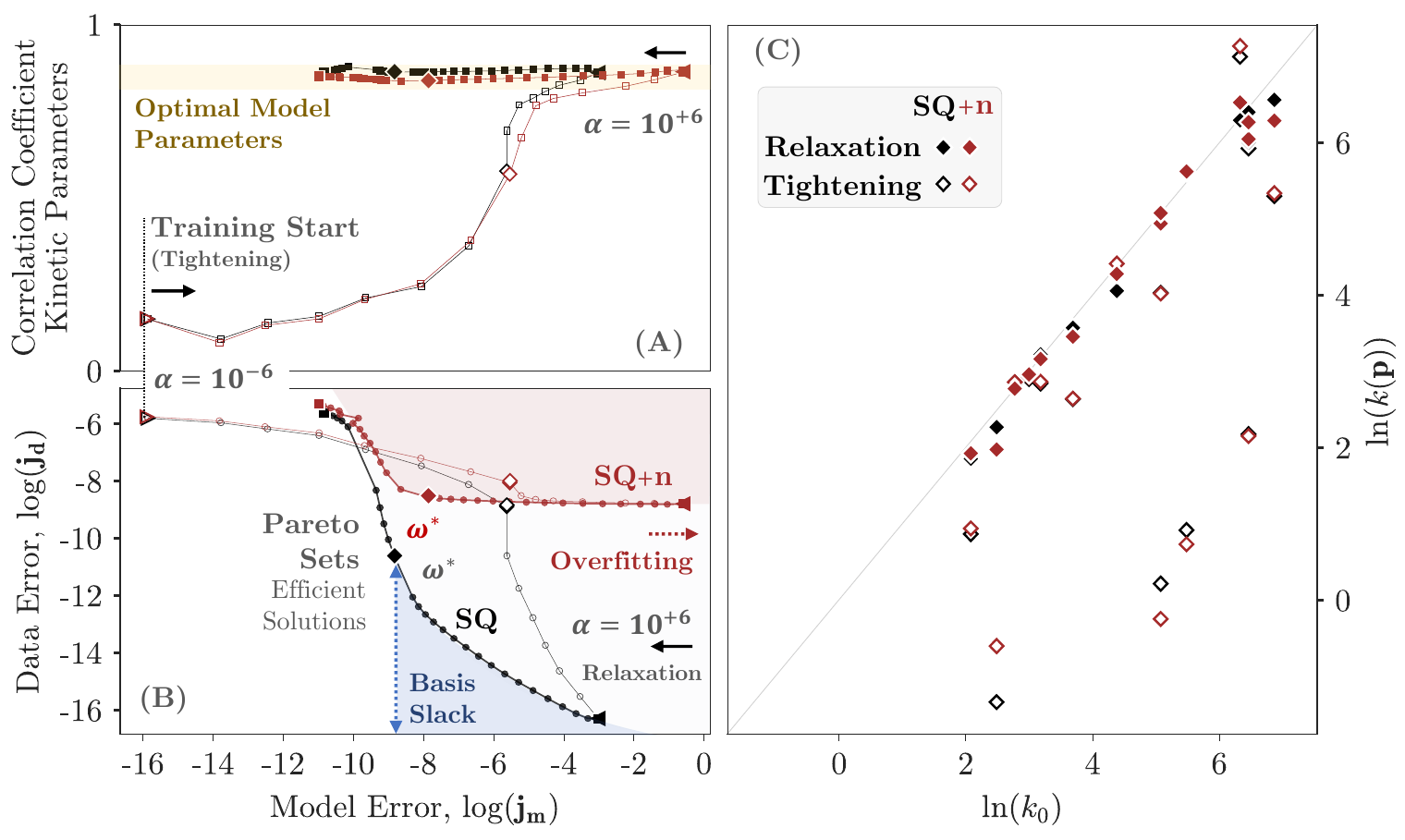}
    \caption{Inverse {\kUDE}~- Regularization Sensitivity Analysis: Pareto Set Estimate. Regression correlation coefficient between fitted and actual kinetic parameters (A), where training (tightening, open markers) starts at the lower left corner at low data regularization ($\alpha$), increasing from left to right as a function of model error $\mathbf{j_m}$. The relaxation series succeeds from right to left (closed markers), where model parameters exhibit minor improvement; however data ($\mathbf{j_d}$) and model- ($\mathbf{j_m}$) error evolution as a function of $\alpha$ define the Pareto frontier in (B) from which non-dominated (efficient) solutions can be obtained. Optimal parameters for scenarios~\semiquant~(red) and~\semiquantn~(black) are shown in (C) at the open and closed inflection points in (B). The basis slack region in (B) conveys the inherent constraint imposed by the choice of simple NN architecture to the derivatives (model) space, which translates into a compromise between interpolation and satisfying the physical model as $\alpha$ is relaxed.}\label{fig:inv_pareto}
\end{figure}}~

\cref{fig:inv_pareto} (B) shows the learned Pareto sets as $\alpha$ is increased (tightened) with open markers from left to right. The SA begins by learning the underlying ODE structure (low $\alpha$, low $\jm$, high $\jd$). As $\alpha$ increases, $\jd$ monotonically decreases with a corresponding increase in $\jm$, resulting in a forward concave set with an inflection point for $\jm$ around $\log(\jm)\approx-4$ (open markers) for both the {\semiquant} (black) and \textit{\semiquantn} (red) scenarios. In \cref{fig:inv_pareto} (A), the steepest improvement in the correlation between the ground-truth and learned model parameter is located in the same region of the concave tightening-Pareto inflection point $\alpha$. When relaxing $\alpha$ with the same SAs trained in the tightening direction, the Pareto set becomes convex, \cref{fig:inv_pareto} (A), as expected for a multiobjective minimization problem between two cost functions. 
This tightening-relaxation strategy and the observed Pareto-set behavior allow the proper estimation of $\alpha$ values in the Pareto set that provide an equilibrated trade-off between satisfying the MKM ODE and interpolating the observed data. Specifically, in the case of \semiquantn, further tightening $\alpha$ beyond the inflection point leads to little improvement (plateau in the red curve, \cref{fig:inv_pareto} B) in the inverse problem solution, instead further fitting the SA to data noise (overfitting). This can be inferred from the trend in the correlation between learned and ground truth model parameters in \cref{fig:inv_pareto} (A), which reaches a stable value for the same range of $\alpha$ where the plateau is observed in \cref{fig:inv_pareto} (B) over successive tightening learning steps. Additionally, if $\alpha$ is further increased, a second inflection point is expected, where there is a detachment of the SA from the underlying MKM ODE and a steep reduction in $\jd$ due to overfitting. In the case of \semiquant, the basis slack region in (B) portrays the inability of the specific choice of SA architecture (i.e. number of layers, activation function) to satisfy both the MKM ODE (physical model) and interpolate the ground-truth data. Such an argument is supported by the universal approximation theorem~\cite{Hornik1990}, which states that, with a sufficient number of neurons, states and their associate derivatives can be approximated to arbitrary accuracy. Nevertheless, it ensues from the universal approximation theorem that there should exist pathological scenarios for sufficiently large or deep NNs where exact interpolation of datapoints is obtained, while excess degrees of freedom allows steep kinks to be created about the interpolated points. In such a scenario, derivatives can span a wider range of possible kinetic parameters that would still satisfy the underlying physics constraints. We opted for constructing NNs that are sufficiently large to adhere to the observed data, but not too large to enable multiple degenerate physical solutions. \cref{fig:inv_pareto} also illustrates, in (C), that the learned physical model parameters at the relaxation inflection point are highly correlated with the ground-truth parameters irrespective of noise. 

In all models in this work, the tightening steps started with $\alpha=10^{-6}$, with $10$-fold $\alpha$ increment between each training cycle, and the relaxation steps reduce $\alpha$ by the finer increments, with the model at the inflection point in relaxation taken as the optimal model. \cref{supp:tab:inv_results} in the \SM\ compiles the performance metrics for the observable variables states and their derivatives for the four different reaction network types, two {\bc}s and three inverse scenarios. Accurate matches between observed states within noise limits and highly correlated derivatives are observed for all cases. In terms of latent variables, the performance metrics in \cref{tab:inv_latent_results} show the agreement between simulated and fitted latent states and their derivatives, irrespective of reaction network type, for scenarios \semiquant~and \semiquantn~. As for the $\obsonly$ scenarios, even with lack of direct knowledge on surface coverages, the surface dynamics information can be indirectly retrieved for model \textit{da}. Hence, the latent (bound) state profile can be inferred from unbound data only. With the increase in depth, the correlation between predicted states starts to fade. This occurs because adsorbed species (type \textit{c} and \textit{s}) coverages are not directly measured in the \obsonly{} scenario, thus there exist multiple solutions for the combination of their evolution over time and the kinetic parameters associated with their reactions. The temporal profiles for the final solutions for observed and latent states as well as their derivative (SA vs. physics model) are reported in the \SM, \cref{supp:fig:inv_da_bc0,supp:fig:inv_da_bc1,supp:fig:invsc_da_bc0,supp:fig:invsc_da_bc1,supp:fig:invvwn_da_bc0,supp:fig:invvwn_da_bc1,supp:fig:inv_dc_bc0,supp:fig:inv_dc_bc1,supp:fig:invsc_dc_bc0,supp:fig:invsc_dc_bc1,supp:fig:invvwn_dc_bc0,supp:fig:invvwn_dc_bc1,supp:fig:inv_dcs_bc0,supp:fig:inv_dcs_bc1,supp:fig:invsc_dcs_bc0,supp:fig:invsc_dcs_bc1,supp:fig:invvwn_dcs_bc0,supp:fig:invvwn_dcs_bc1}.
\begin{table}[ht!]
    \centering
    \caption{Inverse {\kUDE}s Results Summary for reaction types \textit{da}, \textit{dc} and \textit{dcs} - latent states (vs. ground truth, right) and derivatives (physical model, left) error metrics.}
    \label{tab:inv_latent_results}
\begin{tabular}{lllllllll}
\toprule
&&& \multicolumn{3}{c}{Latent States Derivatives} & \multicolumn{3}{c}{Latent States (ground truth)} \\
Type &    {\bc}  & Mode   &r$^2$ &MAE& MSE &  r$^2$ &MAE &MSE \\
\midrule
da & 1 & \obsonly &        1.00 &  9.59 × 10$^{-3}$ &  2.99 × 10$^{-4}$ &        0.972 &  3.17 × 10$^{-2}$ &  1.40 × 10$^{-3}$ \\
    &   & \semiquant &       0.944 &             0.131 &             0.325 &        0.999 &  1.09 × 10$^{-3}$ &  2.30 × 10$^{-6}$ \\
    &   & \semiquantn &       0.941 &  7.62 × 10$^{-2}$ &  1.90 × 10$^{-2}$ &        0.863 &  1.56 × 10$^{-2}$ &  4.79 × 10$^{-4}$ \\\cline{2-9}\\[-1.95ex]
    & 2 & \obsonly &        1.00 &  7.59 × 10$^{-3}$ &  1.69 × 10$^{-4}$ &        0.998 &  3.48 × 10$^{-2}$ &  1.69 × 10$^{-3}$ \\
    &   & \semiquant &       0.998 &  7.82 × 10$^{-2}$ &  4.47 × 10$^{-2}$ &         1.00 &  1.26 × 10$^{-3}$ &  2.23 × 10$^{-6}$ \\
    &   & \semiquantn &        1.00 &  4.47 × 10$^{-2}$ &  3.37 × 10$^{-3}$ &        0.948 &  1.79 × 10$^{-2}$ &  4.15 × 10$^{-4}$ \\\hline\\[-1.95ex]
dc & 1 & \obsonly &        1.00 &  1.91 × 10$^{-2}$ &  5.92 × 10$^{-4}$ &        0.965 &             0.118 &  2.32 × 10$^{-2}$ \\
    &   & \semiquant &       0.999 &  2.63 × 10$^{-2}$ &  3.50 × 10$^{-3}$ &         1.00 &  6.39 × 10$^{-4}$ &  6.21 × 10$^{-7}$ \\
    &   & \semiquantn &       0.983 &  5.96 × 10$^{-2}$ &  5.93 × 10$^{-3}$ &        0.895 &  2.67 × 10$^{-2}$ &  1.29 × 10$^{-3}$ \\\cline{2-9}\\[-1.95ex]
    & 2 & \obsonly &       0.996 &  1.04 × 10$^{-2}$ &  1.54 × 10$^{-4}$ &        0.940 &             0.151 &  3.68 × 10$^{-2}$ \\
    &   & \semiquant &        1.00 &  1.13 × 10$^{-2}$ &  4.60 × 10$^{-4}$ &         1.00 &  6.64 × 10$^{-4}$ &  6.96 × 10$^{-7}$ \\
    &   & \semiquantn &       0.988 &  4.08 × 10$^{-2}$ &  3.70 × 10$^{-3}$ &        0.988 &  3.24 × 10$^{-2}$ &  2.05 × 10$^{-3}$ \\\hline\\[-1.95ex]
dcs & 1 & \obsonly &       0.975 &  1.87 × 10$^{-2}$ &  8.11 × 10$^{-4}$ &        0.374 &             0.146 &  3.84 × 10$^{-2}$ \\
    &   & \semiquant &        1.00 &  5.53 × 10$^{-3}$ &  8.53 × 10$^{-5}$ &         1.00 &  2.66 × 10$^{-3}$ &  1.28 × 10$^{-5}$ \\
    &   & \semiquantn &        1.00 &  1.28 × 10$^{-2}$ &  3.24 × 10$^{-4}$ &        0.996 &  1.47 × 10$^{-2}$ &  3.89 × 10$^{-4}$ \\\cline{2-9}\\[-1.95ex]
    & 2 & \obsonly &       0.736 &  1.81 × 10$^{-2}$ &  9.85 × 10$^{-4}$ &        0.236 &             0.181 &  5.76 × 10$^{-2}$ \\
    &   & \semiquant &        1.00 &  9.30 × 10$^{-3}$ &  2.51 × 10$^{-4}$ &         1.00 &  2.68 × 10$^{-3}$ &  1.30 × 10$^{-5}$ \\
    &   & \semiquantn &       0.998 &  2.11 × 10$^{-2}$ &  1.44 × 10$^{-3}$ &        0.989 &  1.06 × 10$^{-2}$ &  1.45 × 10$^{-4}$ \\
\bottomrule
\end{tabular}
\end{table}
~{

To showcase the robustness of SA-based inverse {\kUDE}s, we consider the final refined solution profiles for the \semiquantn~scenario for the \textit{dcs}-type reaction network with IC 1 in \cref{fig:inv_dcs_bc0}. In the \semiquantn~scenarios, not only are kinetic parameters learned, but also the scaling factor that connects intermediate species abundance as a signal or intensity to fractional coverages. For the optimized regularization, which is dependent on the $\alpha$ value, the underlying SA interpolation suppresses the effect of the added homoscedastic noise. Larger $\alpha$ values would lead to predicted points (interpolation, open markers) moving further towards the measured data (closed circle markers), resulting in SA overfitting with respect to data (i.e. the resulting curve would not satisfy the differential equations of the kinetic model). In contrast to the forward {\kUDE}s, {\bc}s are not explicitly provided, so all observed data act as a composite set of {\bc}s whose weight is conveyed by $\jd$. As depicted by \cref{eqn:ode_nn_inv}, the efficient solutions are given by Pareto-optimal (i.e. non-dominated) points that minimize the multiobjective optimization function $\jt$, involving $\jd$ and $\jm$. The latter, which represents the error in satisfying the MKM ODE, has similar structure as the one for forward {\kUDE}s except that $\parmkm$ becomes a variable for which to solve.
\begin{figure}[ht!]
    \centering
    \includegraphics[trim=0in 0in 0in 0in, keepaspectratio=true,scale=1.,clip=True]{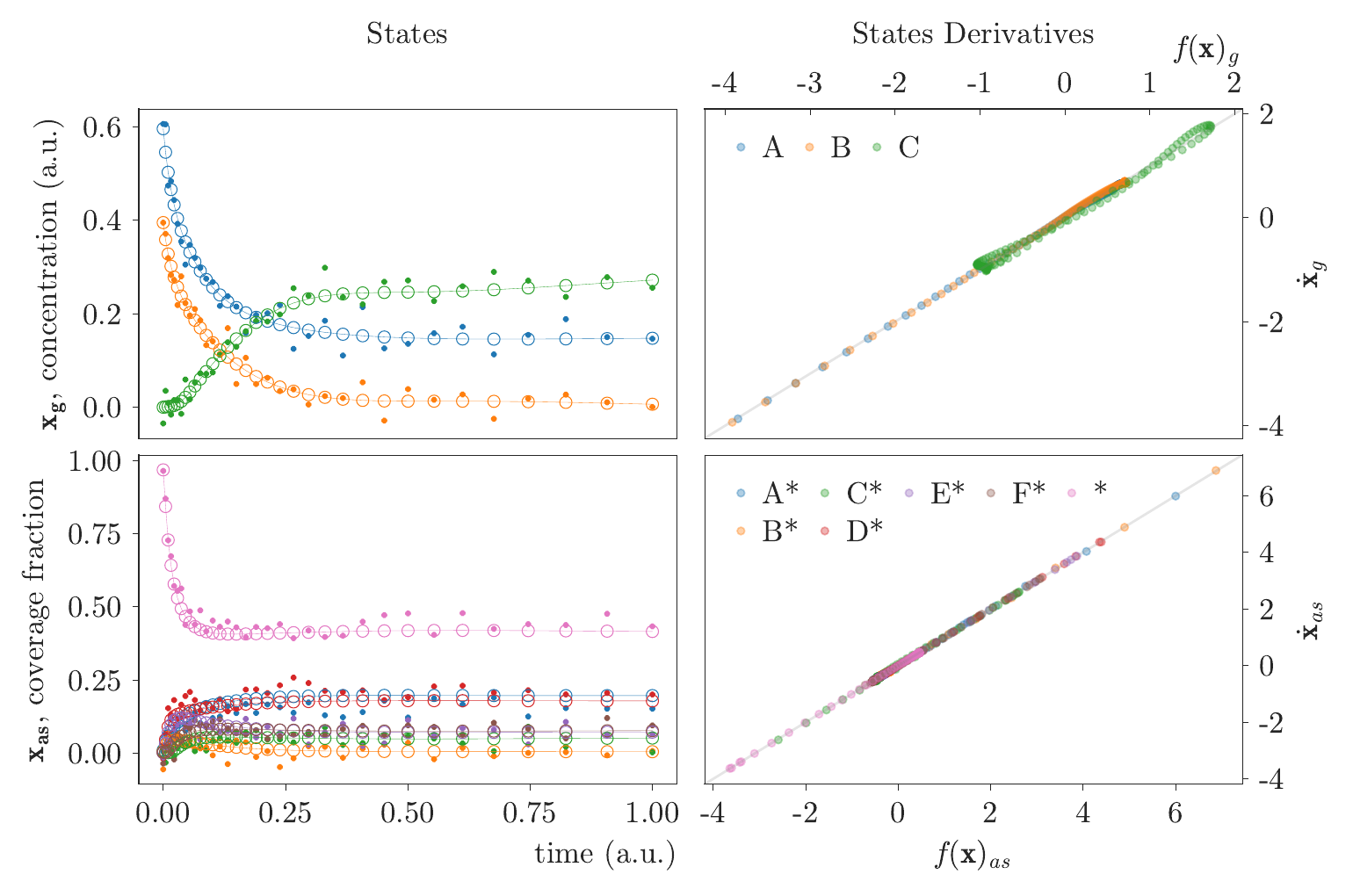}
    \caption{Reaction network type \textit{dcs} inverse solution. States solution (left) (KINN approximation, open circles; \semiquantn data, closed markers) under minimization of \eqref{eqn:ode_nn_inv} (right) for observable variables (unbound species, top) and latent variables (coverages, bottom).}\label{fig:inv_dcs_bc0}
\end{figure}
In inverse {\kUDE}s, as the SA is trained to minimize \cref{eqn:ode_nn_inv}, the kinetic parameters are learned from the underlying physical model (MKM). For all studied cases, kinetic parameters, $\parmkm$, were randomly uniformly initialized at scale of $10^{-2}$, whereas ground-truth values would range approximately from $0\le\ln(k_0)\le8$, as can be inferred from the natural logarithm of the values presented in \cref{tab:fwd_archx}. The final refined solution for the \semiquantn~scenario of \textit{dcs}-type reaction network, with {\bc} 1 and 2 solved (interpolated) in parallel is shown in \cref{fig:inv_dcs_bc0_reg}. These results are illustrative of cases in which knowledge on surface coverages over time, despite the presence of noise, can provide a description of the system dynamics from which kinetic constants can be retrieved. This highlights the importance of operando spectroscopic techniques, especially for complex reaction networks involving reactions between adsorbed intermediates.
}~{
\begin{figure}[ht!]
    \centering
    \includegraphics[trim=0in 0in 0in 0in, keepaspectratio=true,scale=1.,clip=True]{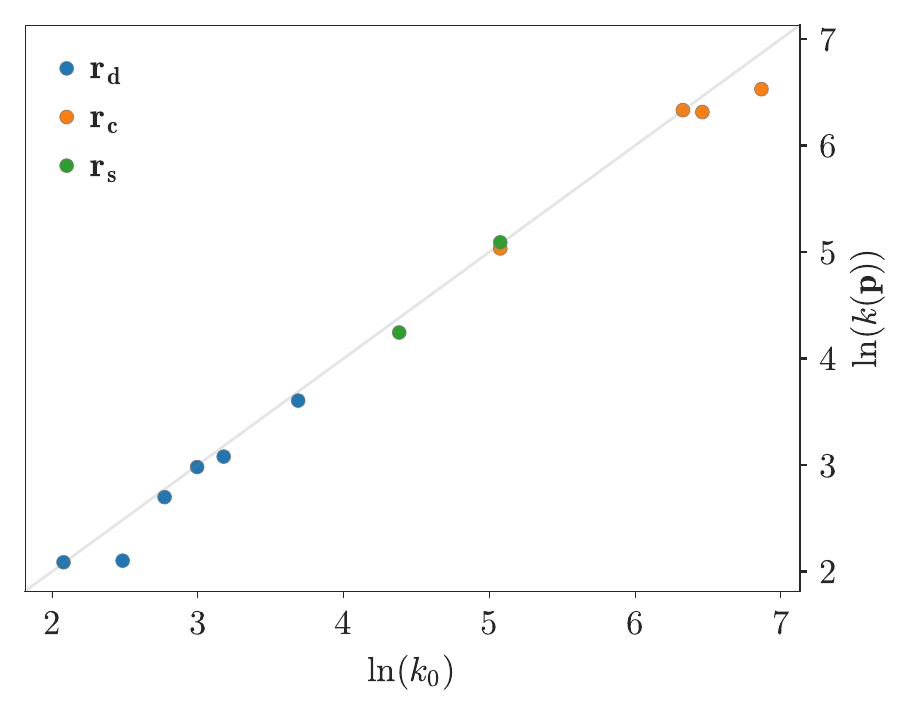}
    \caption{Reaction network type \textit{dcs}, \semiquantn, inverse solution. Natural logarithms of ground truth parameter $\ln{(k_0)}$ vs. regressed parameters, $\ln(k(\parmkm))=\parmkm$.}\label{fig:inv_dcs_bc0_reg}
\end{figure}

A summary of final refined regression results for all inverse {\kUDE}s scenarios is given by \cref{tab:inv_pars}. All scenarios comprise the simultaneous training of SA weights and the natural logarithm of the true kinetic model parameters for {\bc}s 1 and 2. For the purely homogeneous \textit{g} reaction networks, kinetic parameters are easily retrieved with MAE below $10^{-3}$ for the $\obsonly$ scenario (in the absence of noise). The inverse {\kUDE}s for reaction networks of types \textit{g}, \textit{d} and \textit{da} may also be solved, since they involve reactions between unbound species and corresponding bound molecules, but not reactions between intermediate species, as can be seen from the results for reaction network types \textit{g} and \textit{da}. As for reaction networks that include \textit{c} and \textit{g}-type reactions, the results support the need for surface composition information over time, as depicted by the detached correlation coefficients and high MAE for type \textit{c} reactions. In the availability of surface composition information (i.e. signal, intensity), forward and reverse kinetic parameters can be regressed even for the highest-depth reactions, those between surface intermediates (\textit{s}-type). Notably, the framework presented allows for spectroscopic information about any subset of species to be included, making it possible to quantitatively combine transient operando spectroscopic data with transient kinetic data.
\begin{table}[hbt!]
    \centering
    \caption{Inverse {\kUDE}s regression performance metrics summary for type \textit{g}, \textit{da}, \textit{dc} and \textit{dcs} reactions types and scenarios \obsonly, \semiquant and \semiquantn.}
    \label{tab:inv_pars}
\begin{tabular}{llcccccc}
\toprule
    &    & \multicolumn{5}{c}{MAE - $\log(k_0)$} &   total \\
     Type & Mode  & g &  d & a &  c & s &  $\rho$ \\
\midrule
    
\textit{g} & \obsonly &  9.90 × 10$^{-4}$ &                 - &                 - &                 - &                 - &    1.00 \\
    & \obsonly+n~ &  2.29 × 10$^{-4}$ &                 - &                 - &                 - &                 - &    1.00 \\\hline\\[-1.95ex]
\textit{da} & \obsonly &                 - &             0.161 &  9.76 × 10$^{-2}$ &                 - &                 - &   0.991 \\
    & \semiquant &                 - &  5.95 × 10$^{-2}$ &  1.16 × 10$^{-2}$ &                 - &                 - &   0.999 \\
    & \semiquantn &                 - &             0.150 &  8.11 × 10$^{-2}$ &                 - &                 - &   0.999 \\\hline\\[-1.95ex]
\textit{dc} & \obsonly &                 - &             0.520 &                 - &   1.80 × 10$^{1}$ &                 - &  -0.523 \\
    & \semiquant &                 - &  1.70 × 10$^{-2}$ &                 - &  7.32 × 10$^{-2}$ &                 - &    1.00 \\
    & \semiquantn &                 - &             0.236 &                 - &             0.420 &                 - &   0.996 \\\hline\\[-1.95ex]
\textit{dcs} & \obsonly &                 - &              5.06 &                 - &   1.02 × 10$^{1}$ &              1.90 &   0.205 \\
    & \semiquant &                 - &  4.10 × 10$^{-2}$ &                 - &             0.116 &  6.61 × 10$^{-2}$ &   0.999 \\
    & \semiquantn &                 - &             0.110 &                 - &             0.114 &  9.62 × 10$^{-2}$ &   0.997 \\
\bottomrule
\end{tabular}
\end{table}
}~
%
\subsection{Discussion of Numerical Methods Aspects}

As with any numerical technique, there are numerous factors to consider when setting up and solving problems with KINNs. This section contains a brief discussion of each of these factors, the choices used in this work, and the possibility of other options.

\textbf{Architecture }Three-layer NNs have proven satisfactorily flexible to accommodate the different dynamics between the latent and observable variable domains in the inverse and forward problems in this study. Deeper NNs appear to be particularly important when trigonometric {\bc}s, which encompass DAE constraints, are enforced onto the output of the latent variables in inverse {\kUDE}s. Larger or smaller optimized NN architectures and associated activation functions could have produced similar or better results; however, NN architecture optimization is beyond the scope of this work. For non-normalized non-stiff problems, a single \textit{swish} hidden layer suffices to solve the MKM ODEs. For solutions that exhibit stiff behavior in certain regions and flat response in others, the characteristic time can be modelled as a function of the independent variable (e.g., time), which was implemented for all forward {\kUDE}s in this work in accordance with the mathematical structure proposed \cref{sec:kappa}.  

\textbf{Training }Given the high number of weights to be optimized from the SA and underlying NNs, and the possibility of non-convexity especially in cases with large number of reaction intermediates (species of type \textit{s}), we opt for the current standard algorithm for deep NN training, \emph{Adam}, which combines a stochastic gradient descent optimizer with adaptive momentum. Although reliant on a noisy average estimation, SGD has been demonstrated to provide fast convergence and associated parameters, even compared to more elaborate optimizers~\cite{Lecun2015a}. The RMSProp built onto SGD in \emph{Adam} is known to result in improved performance in on-line settings. Considering these properties and the literature-reported success of \emph{Adam} in NN settings~\cite{Kingma2015Adam:Optimization}, it was chosen as an optimizer in both forward and inverse settings.

\textbf{Sampling \& Regularization }As regards the sampling procedure, other strategies could have been utilized instead of sampling evenly spaced, fixed points in the logarithm space. Cross-validation routines in inverse {\kUDE} schemes could prevent overfitting for $\alpha$ values beyond the tightening inflection point in \cref{fig:inv_pareto}. Furthermore, pre-training the SA with low $\alpha$ is paramount, especially in the case of noisy data, since it allows the SA to first learn the MKM ODE, which is further presented the observed data to which to interpolate. Such a procedure is also important when there are too few observed data points and large NNs, to prevent the physical model parameters from rapidly squashing at the beginning of training cycle and to avoid the issue of vanishing gradients. Proper scaling of the physical model parameters is also beneficial to accelerate the learning procedure in our current inverse {\kUDE}s setup, since it provides the learning (optimization) algorithm \emph{Adam} with parameters of similar order of magnitude. 

\textbf{Inverse Problem }In this work, we combine the optimization of the SA and kinetic model parameters into the NN training algorithm. We suggest that the solution for this combined approach may serve as initial guess to accelerate the solution of methods with mathematically rigorous formulations from the standpoint of optimization, such as the adjoint state method~\cite{ayed2019learning} that enforces Karush-Kuhn-Tucker, KKT~\cite{karush1939minima,kuhn1951} optimality conditions. Such methods may vastly benefit from pre-optimized initial guesses, since their deterministic behavior, especially for the adjoint formulation, often leads to local convergence when deterministic gradient-based optimization algorithms are utilized. 

\textbf{MKMs \& {\kUDE}s} Analysis of the inverse solutions for the different reaction network types examples of varying complexity or depth (\cref{subsec:mf_mkm_fmwrk}) highlights the need for operando techniques to allow for information related to species of type \textit{c} and \textit{s} to be retrieved. The \semiquant{} and \semiquantn{} pose similar scenarios with respect to operando techniques. Coverage fraction time series obtained from the forward numerical solution of the initial value problems are rescaled with their multiplication by a random positive scalar, emulating deconvolved spectroscopic signals. Inverse {\kUDE}s therefore entail learning the kinetic parameters associated with the MKM as well as the scaling factor that maps signal intensities to coverages of intermediate species, which is only possible due to the normalization operator that enforces DAE constraints to the ODE solutions. Accurate  parameter estimation for the complex \textit{dcs} mechanism is supporting evidence that inverse {\kUDE}s, seen as either physics-regularized interpolation of data or data-regularized SA-based solution of ODEs, are a robust method for the estimation of kinetic model parameters.
\section{Conclusion}


In this work, we use artificial neural networks (NNs) as basis functions for microkinetic models (MKMs), proposing strategies to create surrogate approximators (SAs) for the solution of the kinetic ODEs. We present a general classification and notation for catalytic reaction networks based on reaction and species types and the phases of species involved. This notation provides an immediate understanding of the extent or depth of dissociation between the observable or measurable states and the underlying intermediates, and hence the complexity of attempting mechanism elucidation. The suitability of NNs as basis functions for the solution of ordinary differential equations is demonstrated by their ability to solve kinetic forward problems. Structural {\bc}s are imposed on the underlying NNs to enforce ODE {\bc}s as in initial value problems. Furthermore, we also propose the inclusion of ancillary NNs to learn the characteristic timescale of the ODEs, which is important in the case of stiffness about the {\bc}. Normalization of the coverage fractions of intermediates is enforced by separating bulk and surface species into two independent neural networks, which structurally enforces a DAE-type of constraint to the ODE solution. 

Using a lumped approach for the simultaneous training of NNs weights and kinetic model parameters in a multiobjective optimization framework, we demonstrate the ability to utilize NNs as basis functions and NN training algorithms to retrieve kinetic parameters from data generated for anecdotal MKM ODEs. The optimization problem is framed in terms of a regularization parameter that can be used to guide the training cycles by (1) low regularization, allowing the SAs to learn or adapt to the physical model ODE, slowly transitioning to (2) high regularization to force the SA to interpolate observed data while parameters of the physical model are simultaneously learned. The regularization parameters convey the relative variance between errors in the physical model (derivatives) space and those associated with states (observed data). This framework can be utilized to estimate Pareto sets and establish associated stopping criteria for training cycles. 

Our SA-based approach for the solution of inverse problem was tested against different types of reaction networks. We show that for homogeneous reactions, when the concentration of species over time are known, or for reactions involving non-dissociated adsorbed chemical species, the approach can be utilized to readily retrieve parameters associated with the underlying kinetic model. For more complex scenarios, where there is incomplete knowledge about concentration of intermediate species on the surface of a catalyst, we assessed the possibility of using deconvolved operando spectroscopic signals for the adsorbed species to extend regression capabilities. Scaling or calibration factors are included as variables in the training process, and we show that our approach can retrieve such factors as well as the kinetic model parameters for these complex systems. Although our analyses are limited to cases where data can be represented as derived from closed, concentrated-parameter models (i.e. models where parameters are spatially uniform), we suggest that SA-based formulation of inverse kinetic ODEs can be applied to transient techniques for the acquisition of kinetic parameters from experimental data, such as temporal analysis of products (TAP) or step-response experiments.

%
%
\section*{Acknowledgements}
The authors acknowledge the U.S. Department of Energy for financial support through contract DE-FE0031719. We are grateful to Dr. John Kitchin and acknowledge his early seminal work on the utilization of neural networks for the solution of simple coupled forward kinetics ODEs available via his blog (\url{kitchingroup.cheme.cmu.edu/blog}). We are also grateful to Dr. Ashi Savara for comments on the pre-print of this manuscript. 
\section*{Conflict of Interest}
The authors declare the nonexistence of conflicts of interest.
\bibliography{main}
\bibliographystyle{unsrt} 
\end{document}


\maketitle

\section*{MKM ODEs Solutions}

This section summarizes the performance metrics for inverse problem solutions of each type of reaction network (\textit{g}, \textit{da}, \textit{dc}, \textit{dcs}) discussed in Results. For each example, results are then shown for quantitative bound species (\textit{\obsonly~}), semiquantitative bound species (\textit{\semiquant~}), and semiquantitative bound species with white noise (\textit{\semiquantn~}) under two boundary conditions (\bc1, \bc2). For these problems, we list the parameters used for data generation, stoichiometry matrices (M), ground truth versus NN predictions for concentration and coverage profiles, and regressed versus true kinetic parameters. The shape of the stoichiometry matrices reflect the elementary reactions of the system such that the rows are associated with the mass balance of each species, and the columns are associated with reaction rates, therefore having the same size as kinetic parameters k. 




\begin{table}[hbt!]
    \centering
    \caption{Inverse {\kUDE}s Results Summary - Observable Variables - Performance Metrics}
    \label{tab:inv_results}
    \begin{tabular}{lllllllll}
\toprule
&&&\multicolumn{3}{c}{Observed States Derivatives} &\multicolumn{3}{c}{Observed States}\\
Type  {\bc}  &Mode   &r$^2$ &$10^{1}$ MAE &$10^{2}$ MSE&r$^2$ &$10^{4}$ MAE &$10^{8}$ MSE \\
\midrule
g &1 &\obsonly~&1.00 &1.24$\times$10$^{-2}$ &4.36$\times$10$^{-4}$&1.00 &0.235 &7.13$\times$10$^{-2}$ \\
    &&\semiquant~&1.00 &1.24$\times$10$^{-2}$ &4.36$\times$10$^{-4}$&1.00 &0.235 &7.13$\times$10$^{-2}$ \\
    &&\semiquantn~&0.998 &0.262 &0.112 &0.999 &4.01$\times$10$^{1}$ &2.77$\times$10$^{3}$ \\\cline{2-9}\\[-1.95ex]
    &2 &\obsonly~&1.00 &1.35$\times$10$^{-3}$ &4.13$\times$10$^{-6}$&1.00 &2.95$\times$10$^{-2}$ &1.14$\times$10$^{-3}$ \\
    &&\semiquant~&1.00 &1.35$\times$10$^{-3}$ &4.13$\times$10$^{-6}$&1.00 &2.95$\times$10$^{-2}$ &1.14$\times$10$^{-3}$ \\
    &&\semiquantn~&0.133 &0.411 &0.330 &0.398 &4.36$\times$10$^{1}$ &3.24$\times$10$^{3}$ \\\hline\\[-1.95ex]
da &1 &\obsonly~&0.960 &0.151 &3.76$\times$10$^{-2}$&1.00 &7.72$\times$10$^{-2}$ &1.42$\times$10$^{-2}$ \\
    &&\semiquant~&0.889 &0.618 &3.74&1.00 &1.91 &6.81 \\
    &&\semiquantn~&0.820 &0.736 &1.09 &0.927 &5.35$\times$10$^{1}$ &7.89$\times$10$^{3}$ \\\cline{2-9}\\[-1.95ex]
    &2 &\obsonly~&1.00 &0.198 &0.151&1.00 &7.78$\times$10$^{-2}$ &1.62$\times$10$^{-2}$ \\
    &&\semiquant~&0.998 &0.529 &1.42&1.00 &0.537 &0.587 \\
    &&\semiquantn~&0.988 &0.551 &0.524 &0.983 &4.33$\times$10$^{1}$ &3.86$\times$10$^{3}$ \\\hline\\[-1.95ex]
dc &1 &\obsonly~&0.987 &0.492 &0.365&1.00 &4.66$\times$10$^{-2}$ &6.69$\times$10$^{-3}$ \\
    &&\semiquant~&1.00 &0.104 &7.12$\times$10$^{-2}$&1.00 &0.235 &9.80$\times$10$^{-2}$ \\
    &&\semiquantn~&0.998 &0.534 &0.530 &0.998 &4.25$\times$10$^{1}$ &3.89$\times$10$^{3}$ \\\cline{2-9}\\[-1.95ex]
    &2 &\obsonly~&0.999 &0.275 &0.139&1.00 &2.70$\times$10$^{-2}$ &1.93$\times$10$^{-3}$ \\
    &&\semiquant~&1.00 &7.08$\times$10$^{-2}$ &9.40$\times$10$^{-3}$&1.00 &4.45$\times$10$^{-2}$ &3.32$\times$10$^{-3}$ \\
    &&\semiquantn~&0.996 &0.652 &0.605 &0.947 &4.47$\times$10$^{1}$ &3.10$\times$10$^{3}$ \\\hline\\[-1.95ex]
dcs &1 &\obsonly~&0.997 &0.730 &1.08&1.00 &0.688 &1.40 \\
    &&\semiquant~&1.00 &6.78$\times$10$^{-2}$ &1.12$\times$10$^{-2}$&1.00 &0.304 &0.412 \\
    &&\semiquantn~&0.997 &0.396 &0.229 &0.999 &3.16$\times$10$^{1}$ &2.01$\times$10$^{3}$ \\\cline{2-9}\\[-1.95ex]
    &2 &\obsonly~&0.995 &0.768 &1.03&1.00 &0.877 &3.16 \\
    &&\semiquant~&1.00 &4.81$\times$10$^{-2}$ &4.08$\times$10$^{-3}$&1.00 &0.475 &0.651 \\
    &&\semiquantn~&0.999 &0.364 &0.208 &0.996 &3.47$\times$10$^{1}$ &2.10$\times$10$^{3}$ \\
\bottomrule
\end{tabular}
\end{table}

\clearpage


\section*{Reaction Network Examples}


\section{Reaction Network Type \textit{g}}\label{sec:typeg}
Reaction Network Type \textit{g} conveys the simplest homogeneous case for the reversible reaction of species $A$ and $B$ into $C$ in gas phase or in solution. To simplify the problem solution, the rate constants $\mathbf{k}$ are suppressed to its bulk states and can be represented in terms of \ref{reac:hom1}, such as: dissociation and dimerization reaction, and lumped-kinetics surrogates derived from complex kinetics data.

\begin{align}
    A+B&\underset{k_2}{\stackrel{k_1}{\rightleftharpoons}} C\label{reac:hom1}\tag{\textbf{g}.1}
\end{align}

\begin{equation}
\begin{aligned}
\ln(\mathbf{k}_0)=\left[2.30\;\;0.00\right]\\\\
M = \left[\begin{array}{rr}
-1&1 \\
-1&1 \\
 1 &-1 \\
\end{array}\right]\\\\
\mathbf{x}^T=\left[x_{A}\;\;x_{B}\;\;x_{C}\right]
\end{aligned}
\end{equation}

\begin{figure}[ht!] 
    \centering

    \includegraphics[trim=0in 0in 0in 0in, keepaspectratio=true,scale=0.95,clip=True]{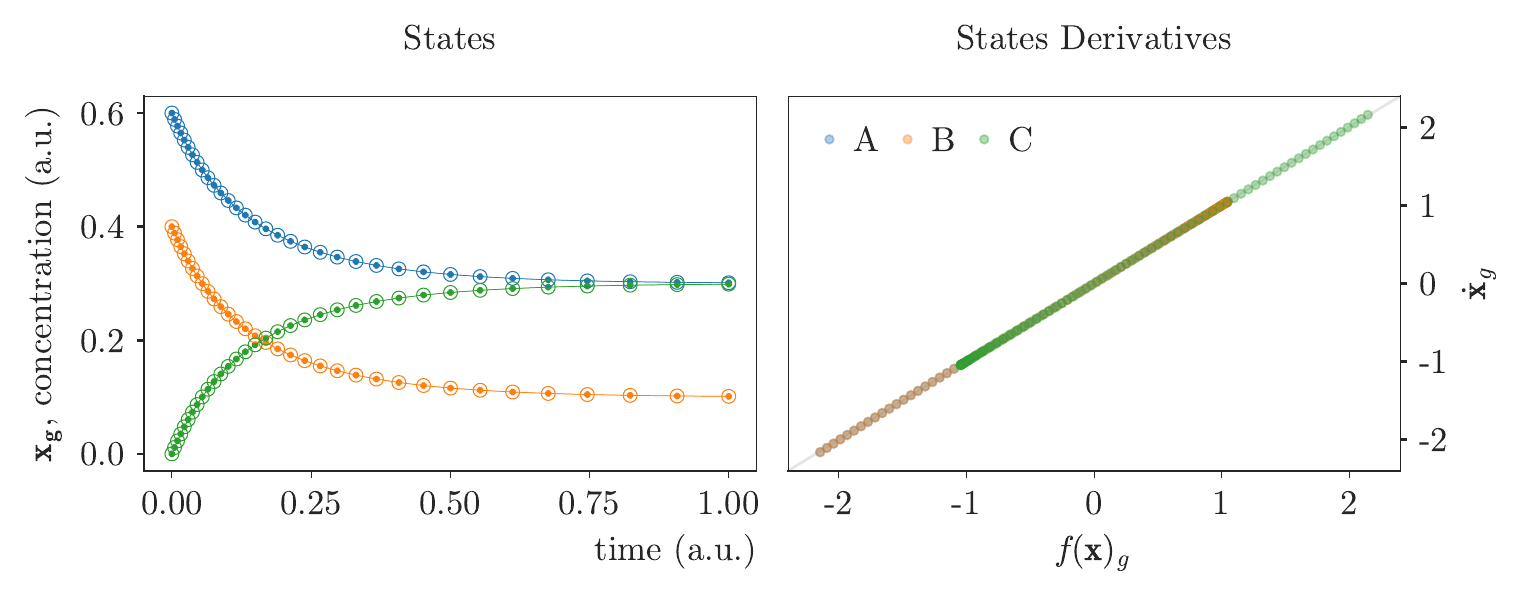}
    \caption{Reaction network type \textit{g} inverse solution for \obsonly~data using \bc1. States solution (left) (\kUDE, open circles; observed data, closed markers) under minimization of \eqref{main:eqn:ode_nn_inv} (right) for observable variables (unbound species, top).}\label{fig:inv_g_bc0}
\end{figure}

\begin{figure}[ht!] 
    \centering

    \includegraphics[trim=0in 0in 0in 0in, keepaspectratio=true,scale=0.95,clip=True]{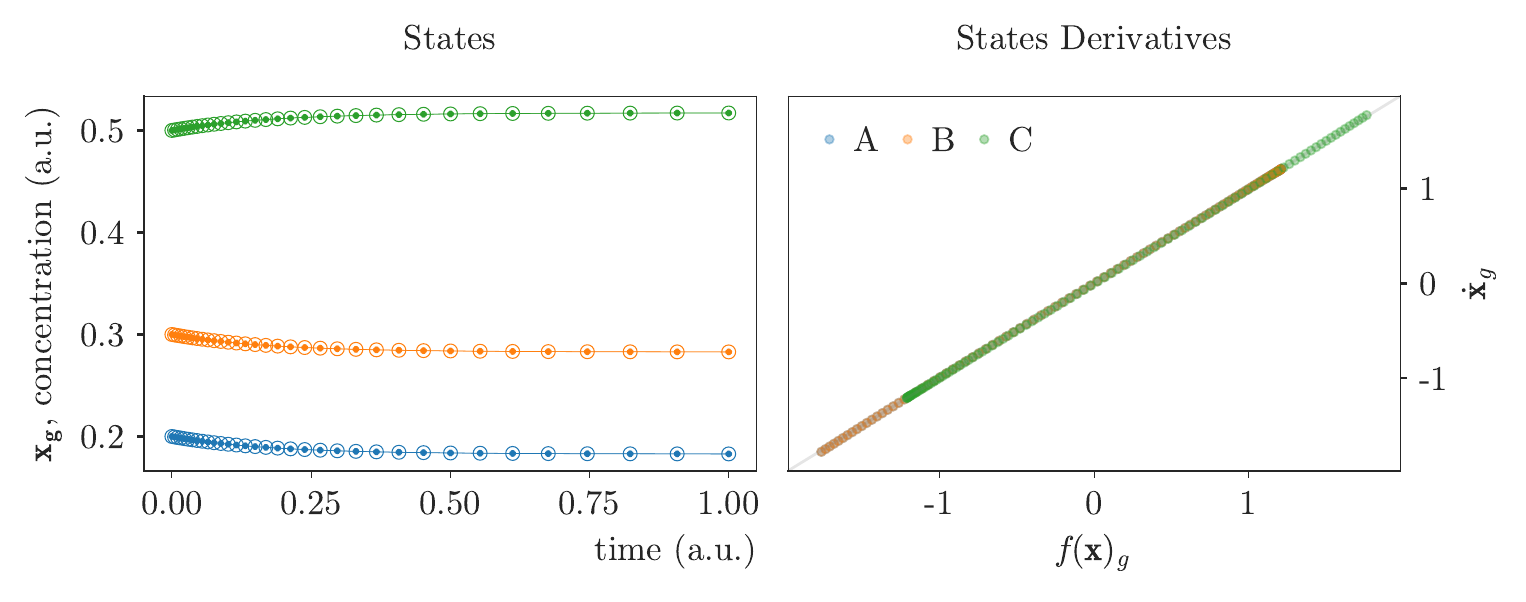}
    \caption{Reaction network type \textit{g} inverse solution for \obsonly~data using \bc2. States solution (left) (\kUDE, open circles; observed data, closed markers) under minimization of \eqref{main:eqn:ode_nn_inv} (right) for observable variables.}\label{fig:inv_g_bc1}
\end{figure}

\begin{figure}[ht!] 
    \centering

    \includegraphics[trim=0in 0in 0in 0in, keepaspectratio=true,scale=0.95,clip=True]{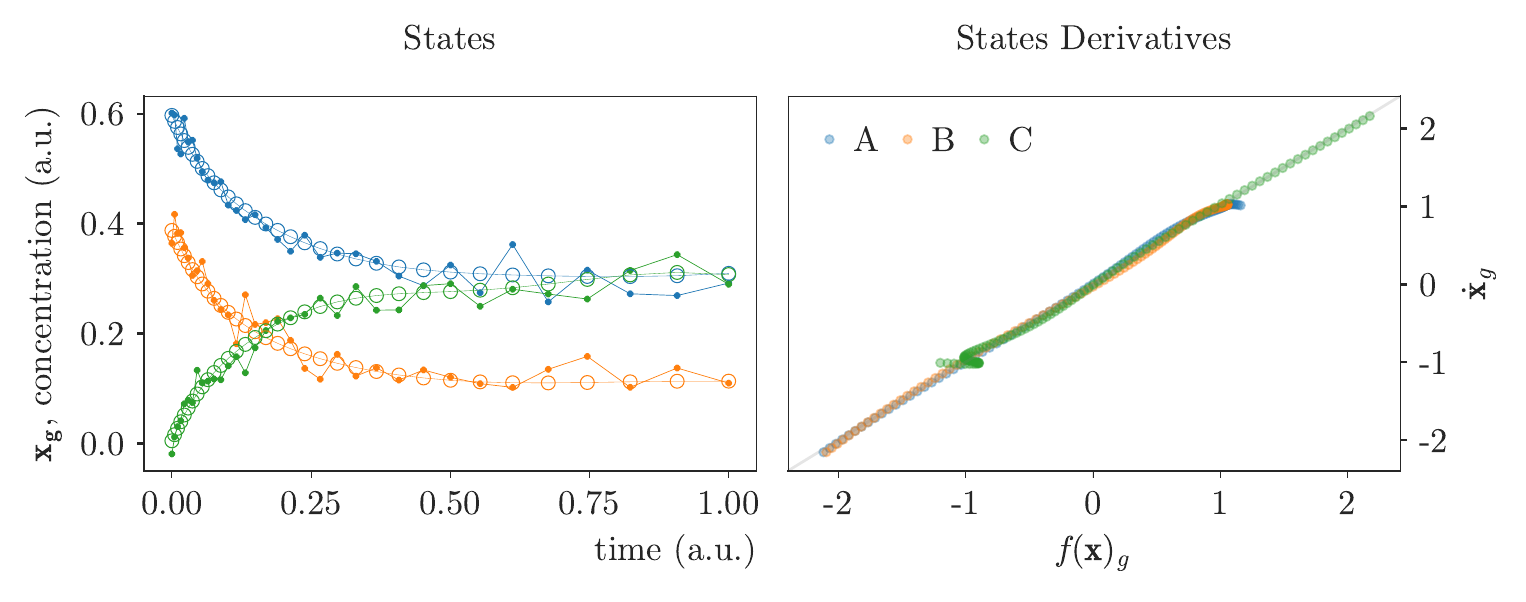}
    \caption{Reaction network type \textit{g} inverse solution for data with noise using \bc1. States solution (left) (\kUDE, open circles; \obsonly~+n data, closed markers) under minimization of \eqref{main:eqn:ode_nn_inv} (right) for observable variables (unbound species, top).}\label{fig:invvwn_g_bc0}
\end{figure}

\begin{figure}[ht!] 
    \centering

    \includegraphics[trim=0in 0in 0in 0in, keepaspectratio=true,scale=0.95,clip=True]{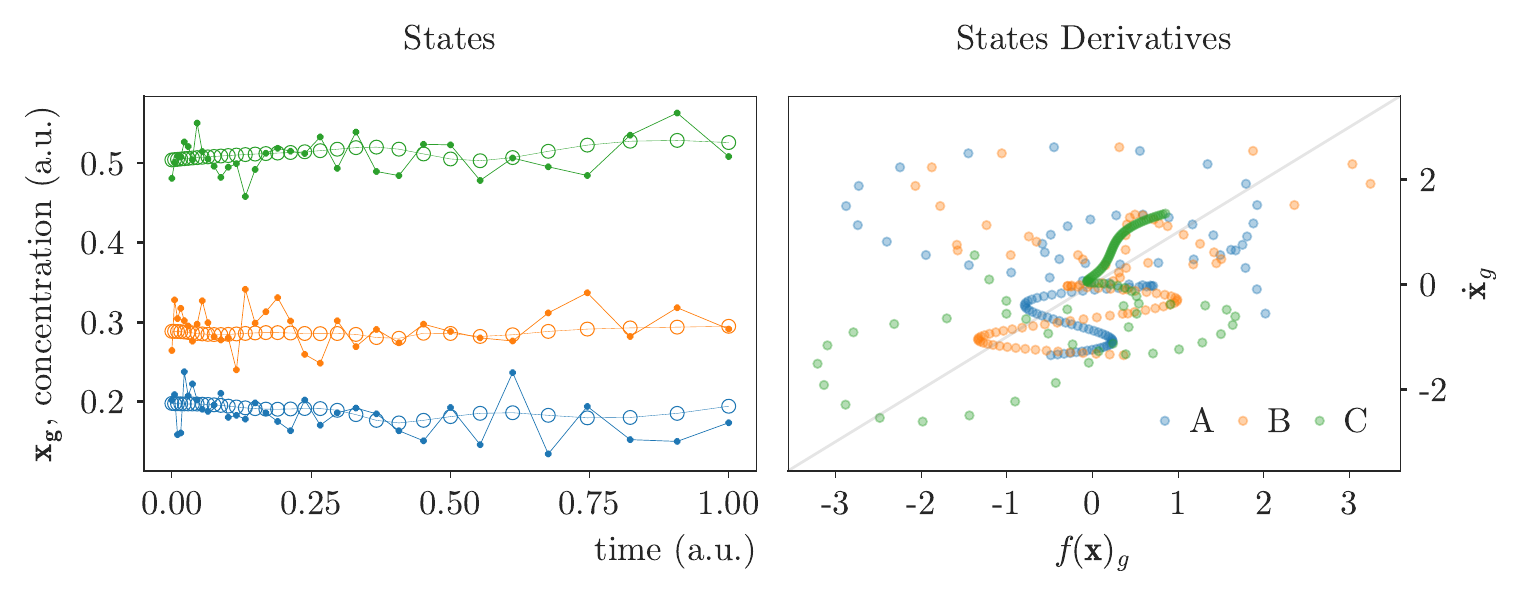}
    \caption{Reaction network type \textit{g} inverse solution for data with noise using \bc2. States solution (left) (\kUDE, open circles; \obsonly~+n data, closed markers) under minimization of \eqref{main:eqn:ode_nn_inv} (right) for observable variables.}\label{fig:invvwn_g_bc1}
\end{figure}

\begin{figure}[ht!] 
    \centering

    \includegraphics[trim=0in 0in 0in 0in, keepaspectratio=true,scale=1.,clip=True]{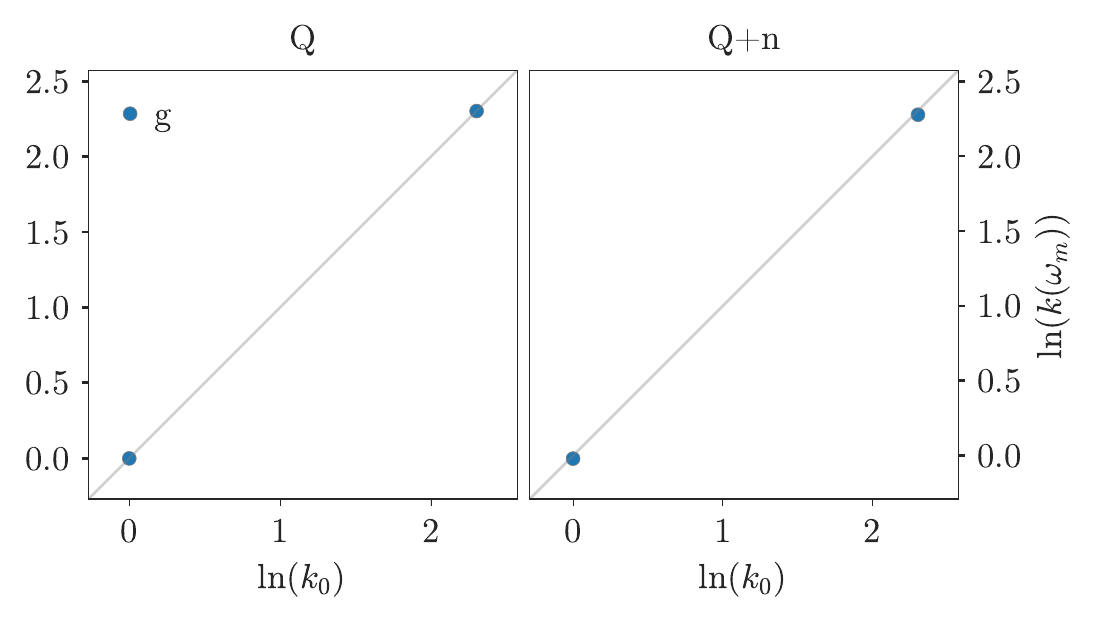}
    \caption{Reaction network type \textit{g} inverse solution. Natural logarithms of ground truth parameter $\ln{(k_0)}$ vs. regressed parameters, $\ln(k(\parmkm))=\parmkm$, with and without noise.}\label{fig:inv_g_reg}
\end{figure}

\clearpage


\section{Reaction Network Type \textit{da}}\label{sec:typeda}

Here we transpose the reactive step from the homogeneous phase in \ref{reac:hom1} to the heterogeneous phase by including the adsorption and desorption elementary steps, \textbf{d}.(1-3). and a surface reaction between adsorbed molecules, \ref{reac:a1}.

\begin{align}
    A+*&\underset{k_4}{\stackrel{k_3}{\rightleftharpoons}} A*\label{reac:adsA}\tag{\textbf{d}.1}\\
    B+*&\underset{k_6}{\stackrel{k_5}{\rightleftharpoons}} B*\label{reac:adsB}\tag{\textbf{d}.2}\\
    C+*&\underset{k_8}{\stackrel{k_7}{\rightleftharpoons}} C*\label{reac:adsC}\tag{\textbf{d}.3}\\[-1.ex]
    \mathclap{\vspace{-10pt}\rule{5cm}{0.2pt}}\notag \\[-1.ex]
    A*+B*&\underset{k_{10}}{\stackrel{k_9}{\rightleftharpoons}} C*\label{reac:a1}\tag{\textbf{a}.1}
\end{align}

\begin{equation}
\begin{aligned}
\ln(\mathbf{k}_0)=\left[2.30\:\;1.39\:\;3.69\:\;4.09\:\;5.30\:\;3.69\:\;4.61\:\;4.38\right]\\\\M=
\left[\begin{array}{rrrrrrrrrr}
-1&1&0&0&0&0&0&0 \\
 0&0 &-1&1&0&0&0&0 \\
 0&0&0&0 &-1&1&0&0 \\
 1 &-1&0&0&0&0 &-1&1 \\
 0&0&1 &-1&0&0 &-1&1 \\
 0&0&0&0&1 &-1&1 &-1 \\
-1&1 &-1&1 &-1&1&1 &-1 \\
\end{array}\right]\\\\
\mathbf{x}^T=\left[x_{A}\;\;x_{B}\;\;x_{C}\;\;x_{A*}\;\;x_{B*}\;\;x_{C*}\;\;x_{*}\right]
\end{aligned}
\end{equation}

\begin{figure}[ht!] 
    \centering
    
    \includegraphics[trim=0in 0in 0in 0in, keepaspectratio=true,scale=0.95,clip=True]{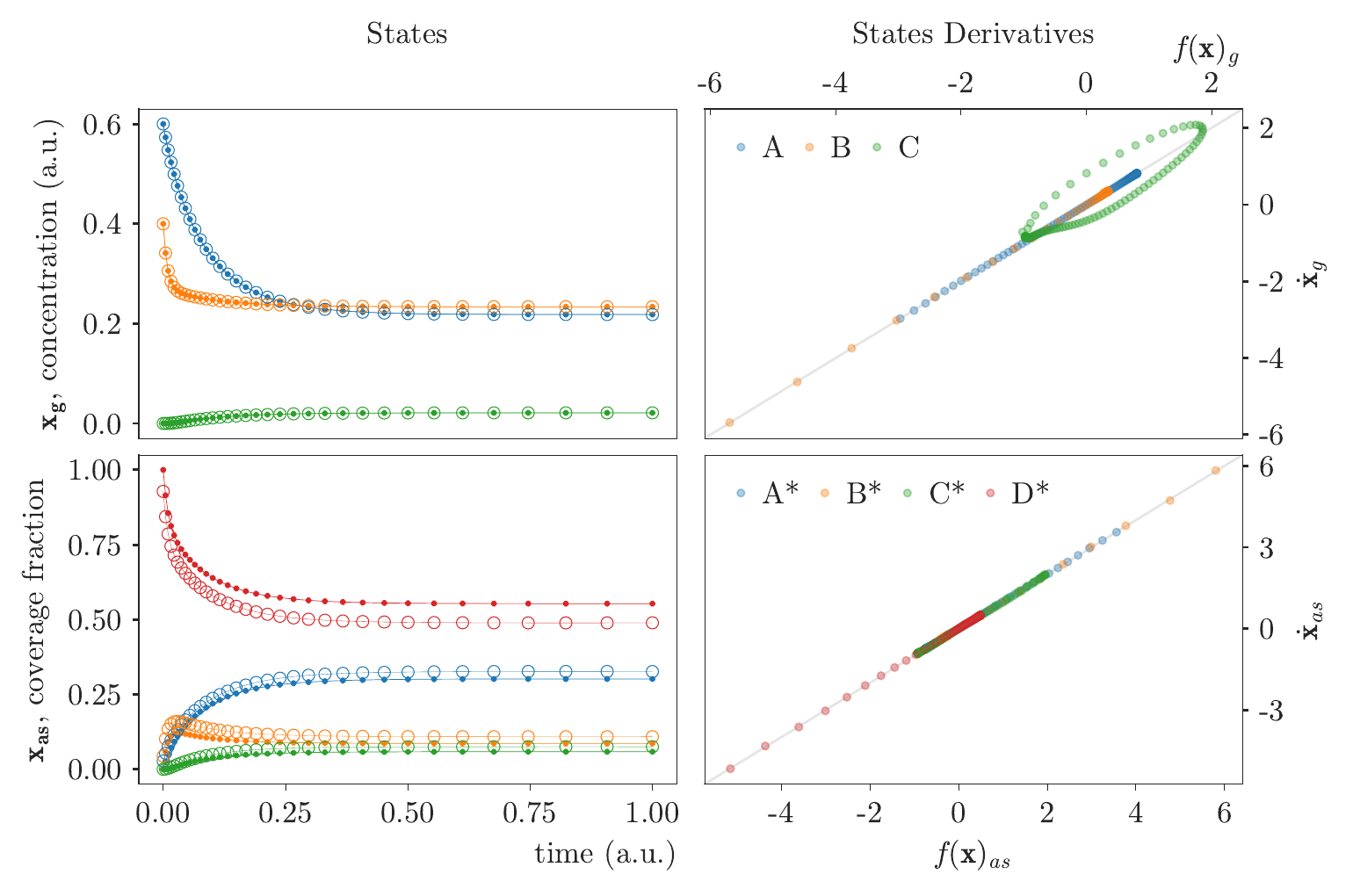}
    \caption{Reaction network type \textit{da} inverse solution for \obsonly~data using \bc1. States solution (left) (\kUDE, open circles; observed data, closed markers) under minimization of \eqref{main:eqn:ode_nn_inv} (right) for observable variables (unbound species, top).}\label{fig:inv_da_bc0}
\end{figure}

\begin{figure}[ht!] 
    \centering
    
    \includegraphics[trim=0in 0in 0in 0in, keepaspectratio=true,scale=0.95,clip=True]{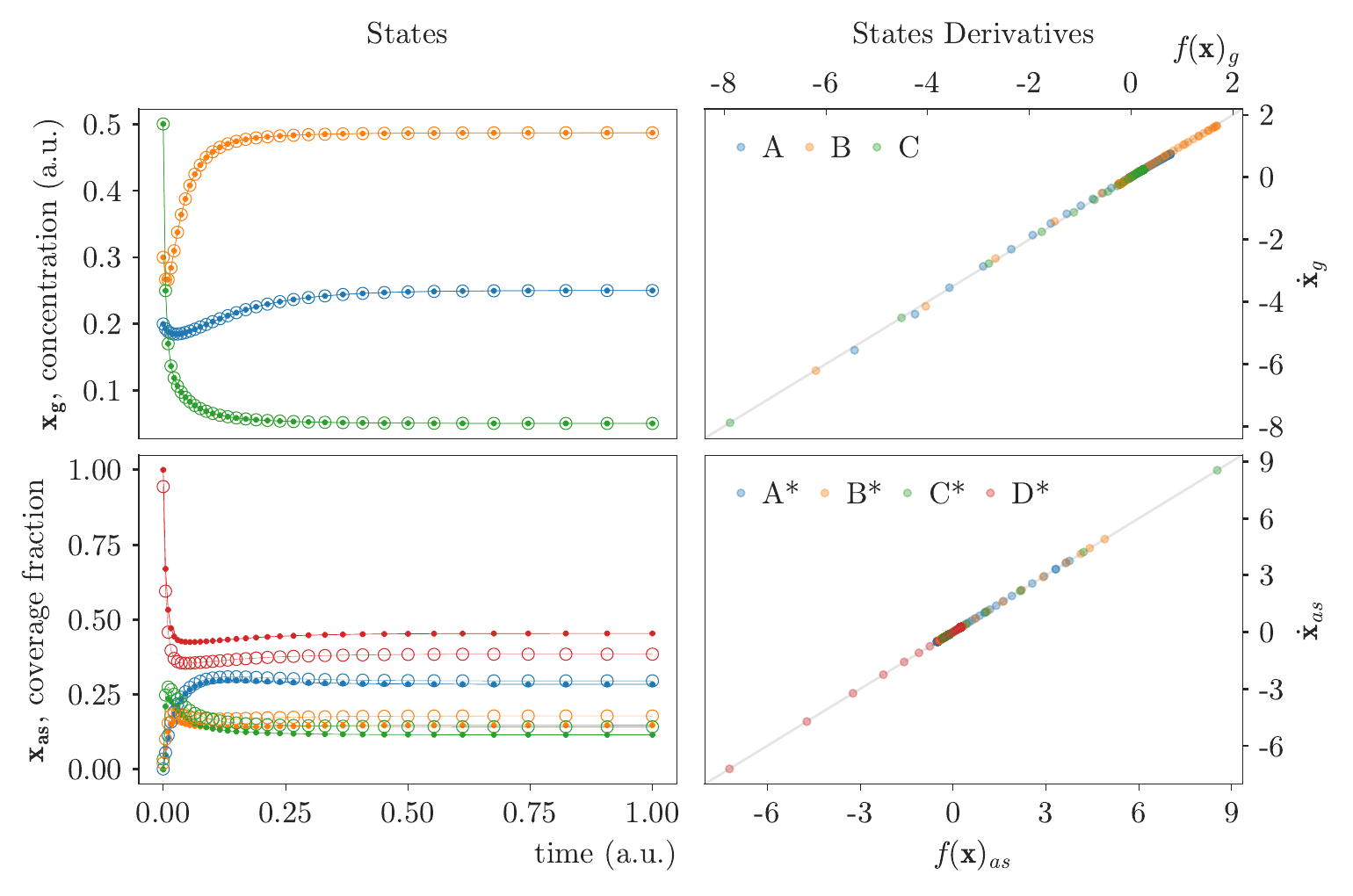}
    \caption{Reaction network type \textit{da} inverse solution for \obsonly~data using \bc2. States solution (left) (\kUDE, open circles; observed data, closed markers) under minimization of \eqref{main:eqn:ode_nn_inv} (right) for observable variables (unbound species, top).}\label{fig:inv_da_bc1}
\end{figure}

\begin{figure}[ht!]
    \centering
    
    \includegraphics[trim=0in 0in 0in 0in, keepaspectratio=true,scale=0.95,clip=True]{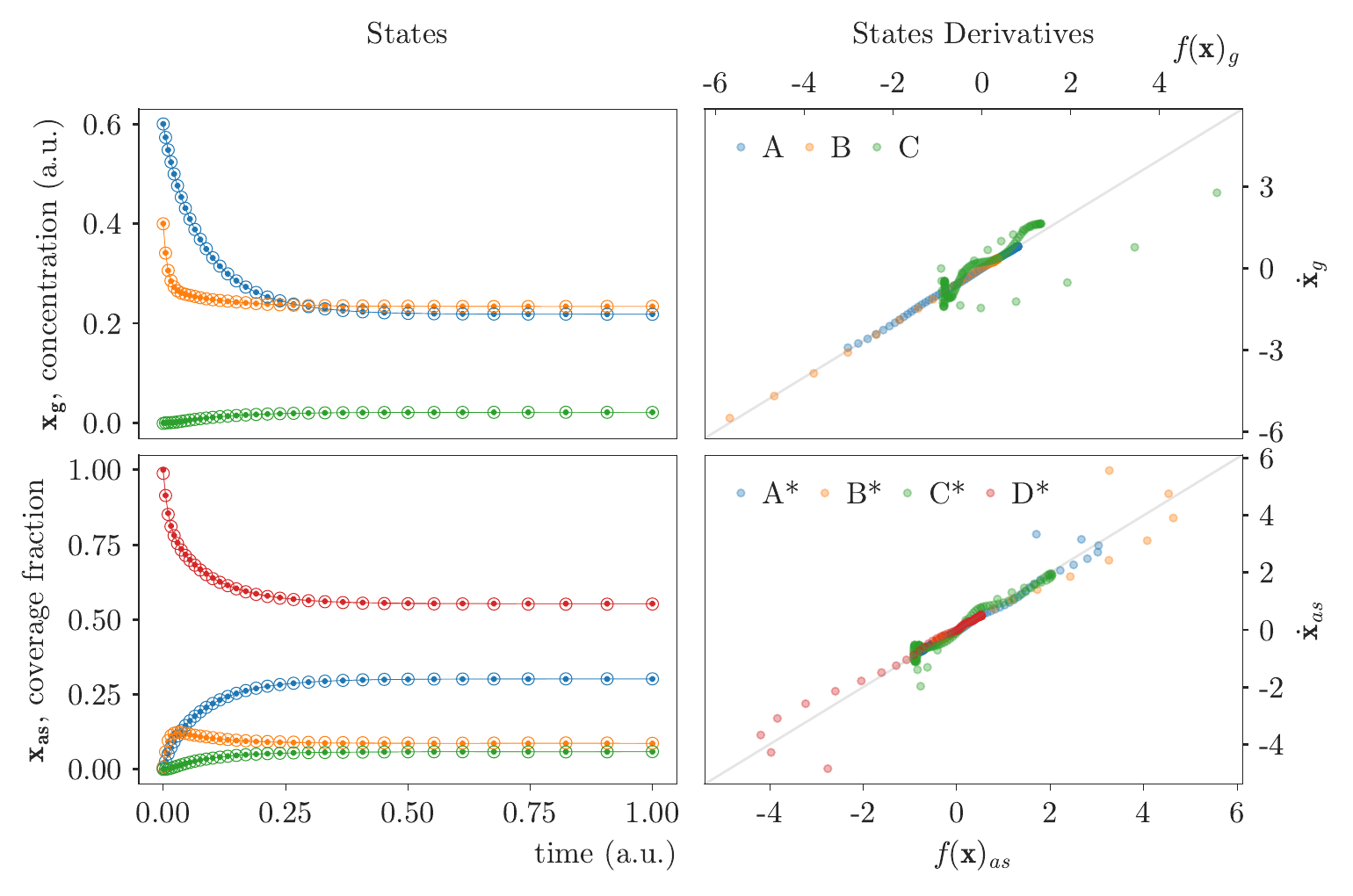}
    \caption{Reaction network type \textit{da} inverse solution for data with noise using \bc1. States solution (left) (\kUDE, open circles; \semiquant~data, closed markers) under minimization of \eqref{main:eqn:ode_nn_inv} (right) for observable variables.}\label{fig:invsc_da_bc0}
\end{figure}

\begin{figure}[ht!] 
    \centering
    
    \includegraphics[trim=0in 0in 0in 0in, keepaspectratio=true,scale=0.95,clip=True]{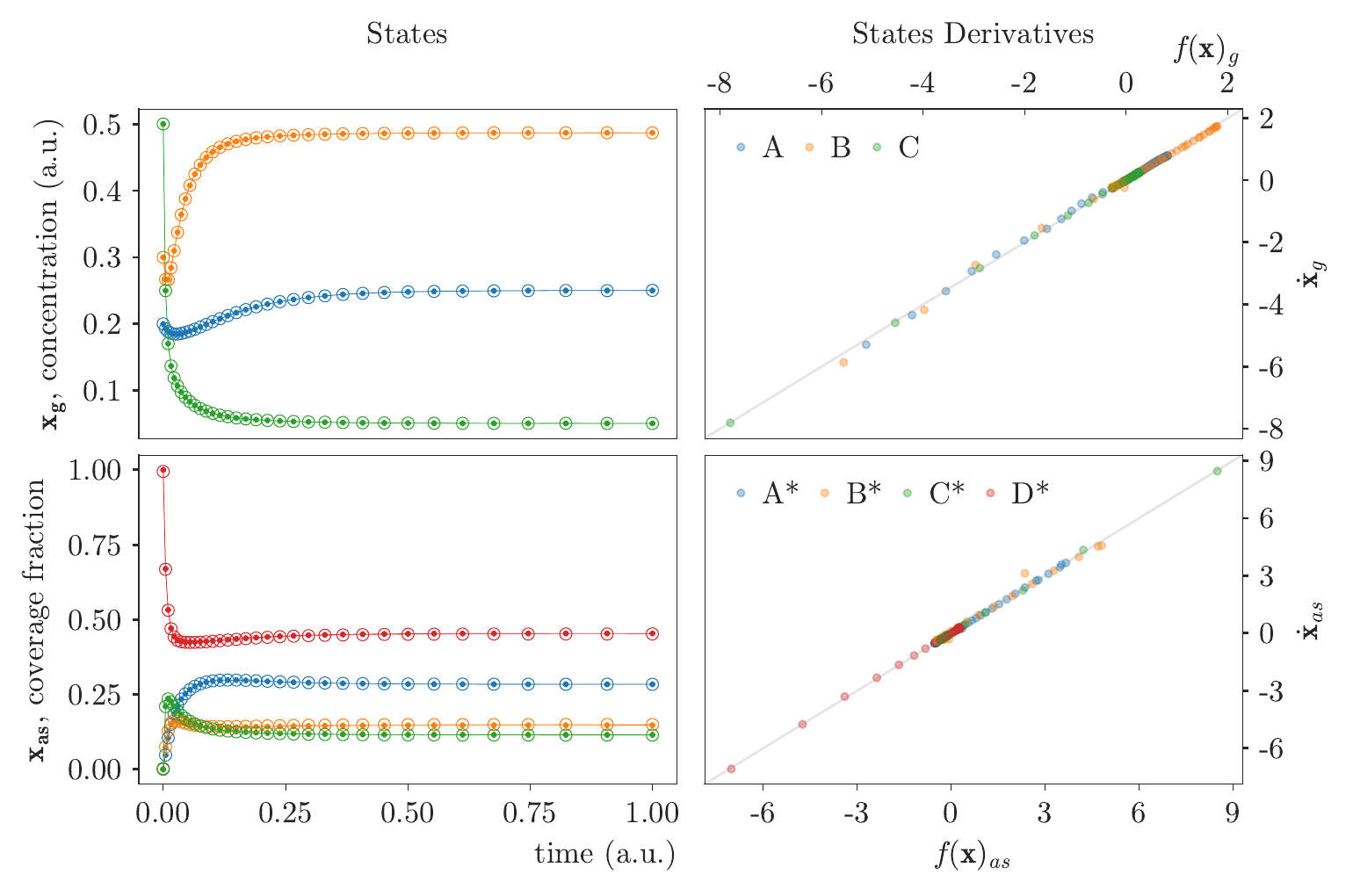}
    \caption{Reaction network type \textit{da} inverse solution for data with noise using \bc2. States solution (left) (\kUDE, open circles; \semiquant~data, closed markers) under minimization of \eqref{main:eqn:ode_nn_inv} (right) for observable variables.}\label{fig:invsc_da_bc1}
\end{figure}

\begin{figure}[ht!] 
    \centering
    
    \includegraphics[trim=0in 0in 0in 0in, keepaspectratio=true,scale=0.95,clip=True]{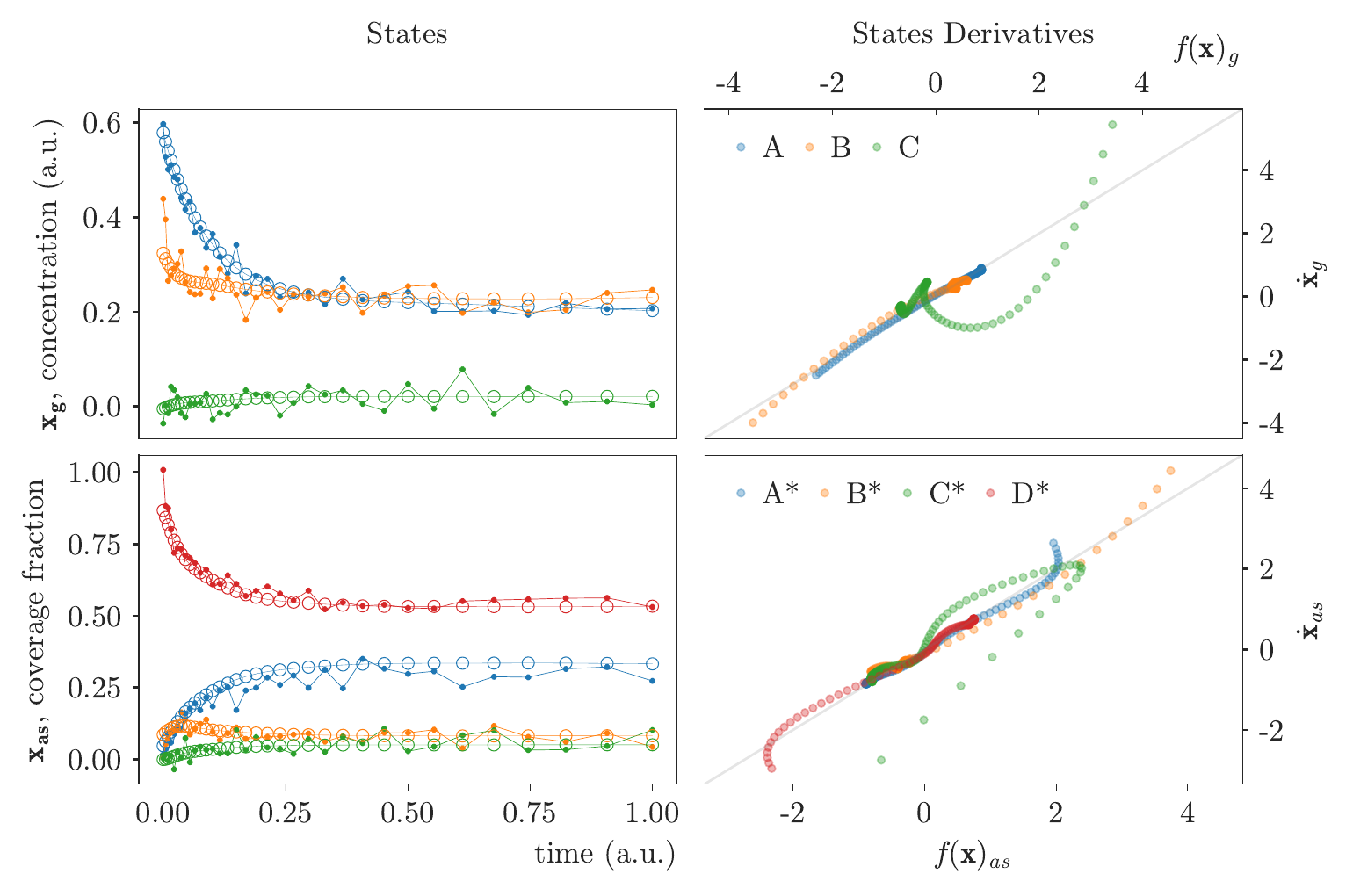}
    \caption{Reaction network type \textit{da} inverse solution for data with noise using \bc1. States solution (left) (\kUDE, open circles; \semiquantn~data, closed markers) under minimization of \eqref{main:eqn:ode_nn_inv} (right) for observable variables.}\label{fig:invvwn_da_bc0}
\end{figure}

\begin{figure}[ht!]
    \centering
    
    \includegraphics[trim=0in 0in 0in 0in, keepaspectratio=true,scale=0.95,clip=True]{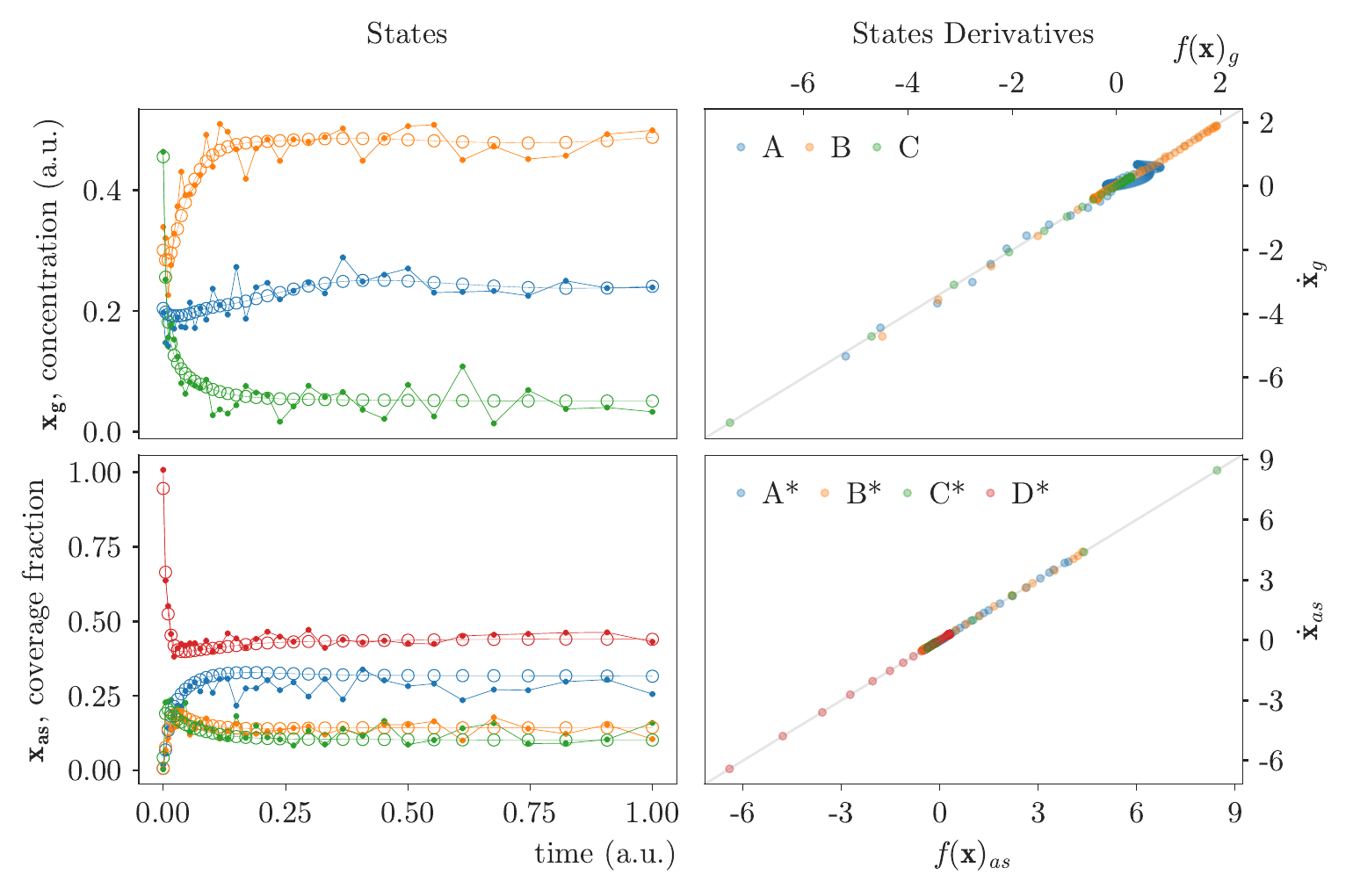}
    \caption{Reaction network type \textit{da} inverse solution for data with noise using \bc2. States solution (left) (\kUDE, open circles; \semiquantn~data, closed markers) under minimization of \eqref{main:eqn:ode_nn_inv} (right) for observable variables.}\label{fig:invvwn_da_bc1}
\end{figure}

\begin{figure}[ht!] 
    \centering

    \includegraphics[trim=0in 0in 0in 0in, keepaspectratio=true,scale=1.,clip=True]{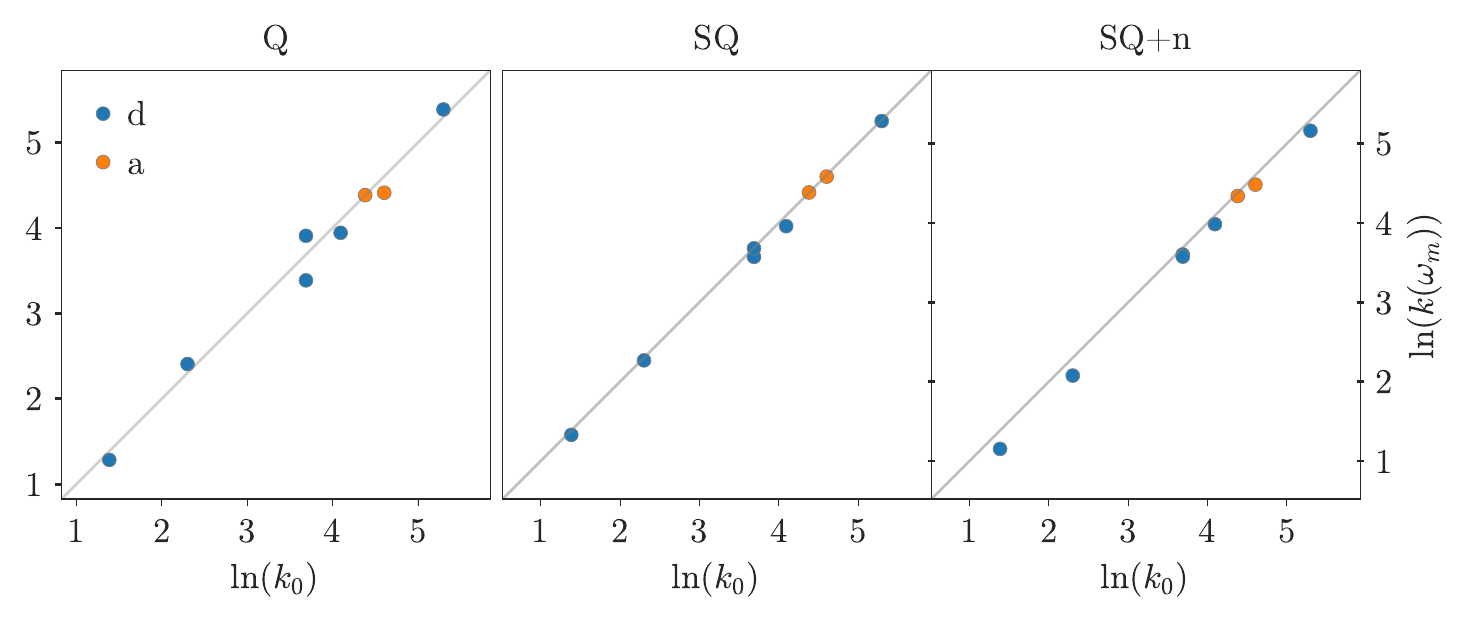}
    \caption{Reaction network type \textit{da} inverse solution. Natural logarithms of ground truth parameter $\ln{(k_0)}$ vs. regressed parameters, $\ln(k(\parmkm))=\parmkm$, for \obsonly, \semiquant~and \semiquantn~data.}\label{fig:inv_da_reg}
\end{figure}

\clearpage


\section{Reaction Network Type \textit{dc}}

The \textit{a}-type reaction is replaced by two \textit{d}-type reactions, i.e. reaction involving one intermediate species ($D*$), which does not have a corresponding gas phase species. The \textit{d}-type are suppressed in the following set of elementary reactions. 

\begin{align}
    B*+*&\underset{k_{12}}{\stackrel{k_{11}}{\rightleftharpoons}} 2D*\tag{\textbf{c}.1}\\
    A*\:+\:D*&\underset{k_{14}}{\stackrel{k_{13}}{\rightleftharpoons}} C*\:+\:*\tag{\textbf{c}.2}
\end{align}

\begin{equation}
\begin{aligned}
\ln(\mathbf{k}_0)=\left[3.00\:\;2.08\:\;2.77\:\;1.39\:\;2.48\:\;2.08\:\;7.09\:\;5.99\:\;7.60\:\;7.38\right]
\\\\
M= 
\left[\begin{array}{rrrrrrrrrrrr}
 -1&1&0&0&0&0&0&0&0&0 \\
 0&0 &-1&1&0&0&0&0&0&0 \\
 0&0&0&0 &-1&1&0&0&0&0 \\
 1 &-1&0&0&0&0&0&0 &-1&1 \\
 0&0&1 &-1&0&0 &-1&1&0&0 \\
 0&0&0&0&1 &-1&0&0&1 &-1 \\
 0&0&0&0&0&0&2 &-2 &-1&1 \\
-1&1 &-1&1 &-1&1 &-1&1&1 &-1 \\
\end{array}\right]\\\\
\mathbf{x}^T=\left[x_{A}\;\;x_{B}\;\;x_{C}\;\;x_{A*}\;\;x_{B*}\;\;x_{C*}\;\;x_{D*}\;\;x_{*}\right]
\end{aligned}
\end{equation}

\begin{figure}[ht!]
    \centering
    
    \includegraphics[trim=0in 0in 0in 0in, keepaspectratio=true,scale=0.95,clip=True]{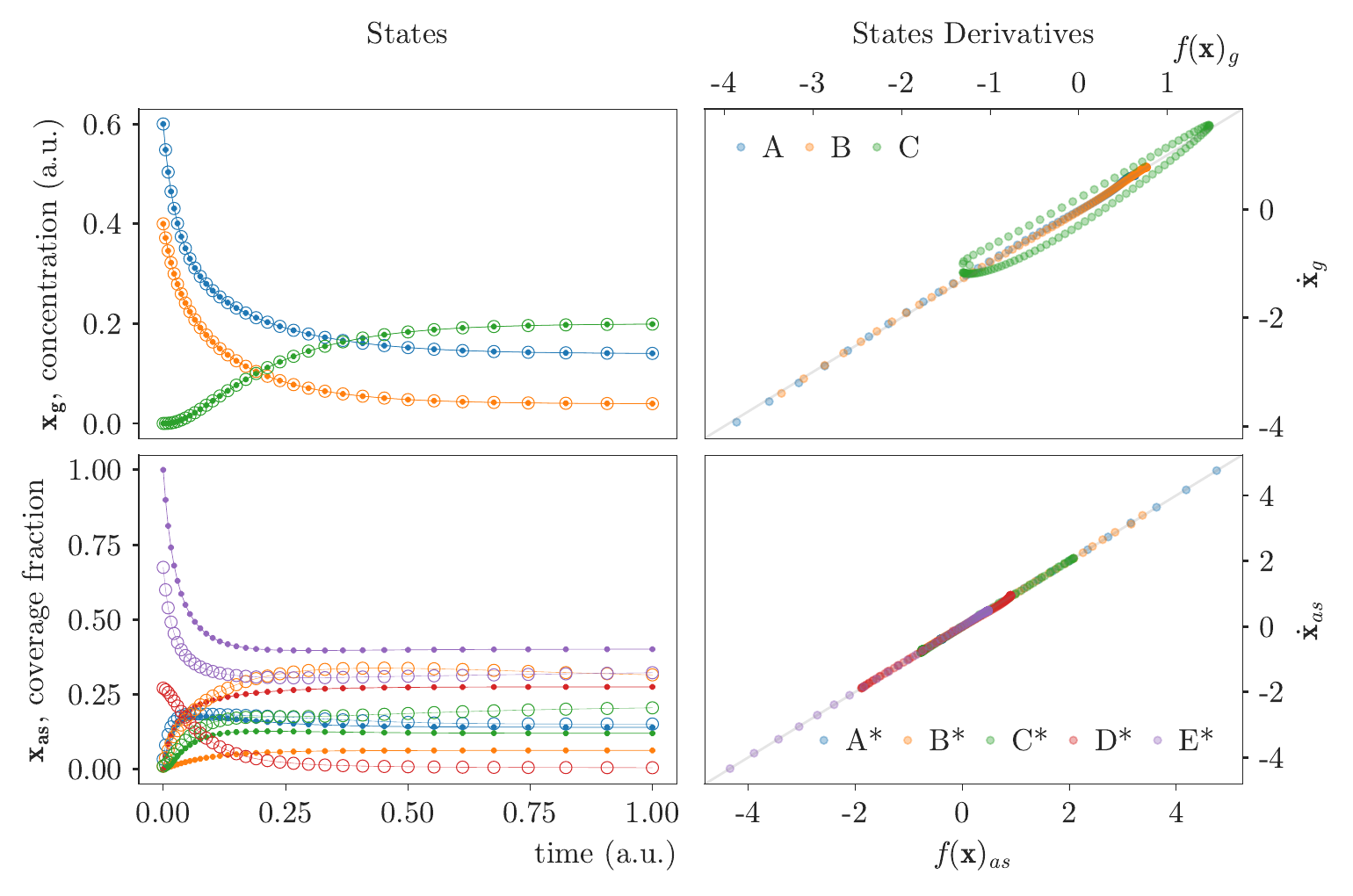}
    \caption{Reaction network type \textit{dc} inverse solution for \obsonly~data using \bc1. States solution (left) (\kUDE, open circles; observed data, closed markers) under minimization of \eqref{main:eqn:ode_nn_inv} (right) for observable variables (unbound species, top).}\label{fig:inv_dc_bc0}
\end{figure}

\begin{figure}[ht!]
    \centering
    
    \includegraphics[trim=0in 0in 0in 0in, keepaspectratio=true,scale=0.95,clip=True]{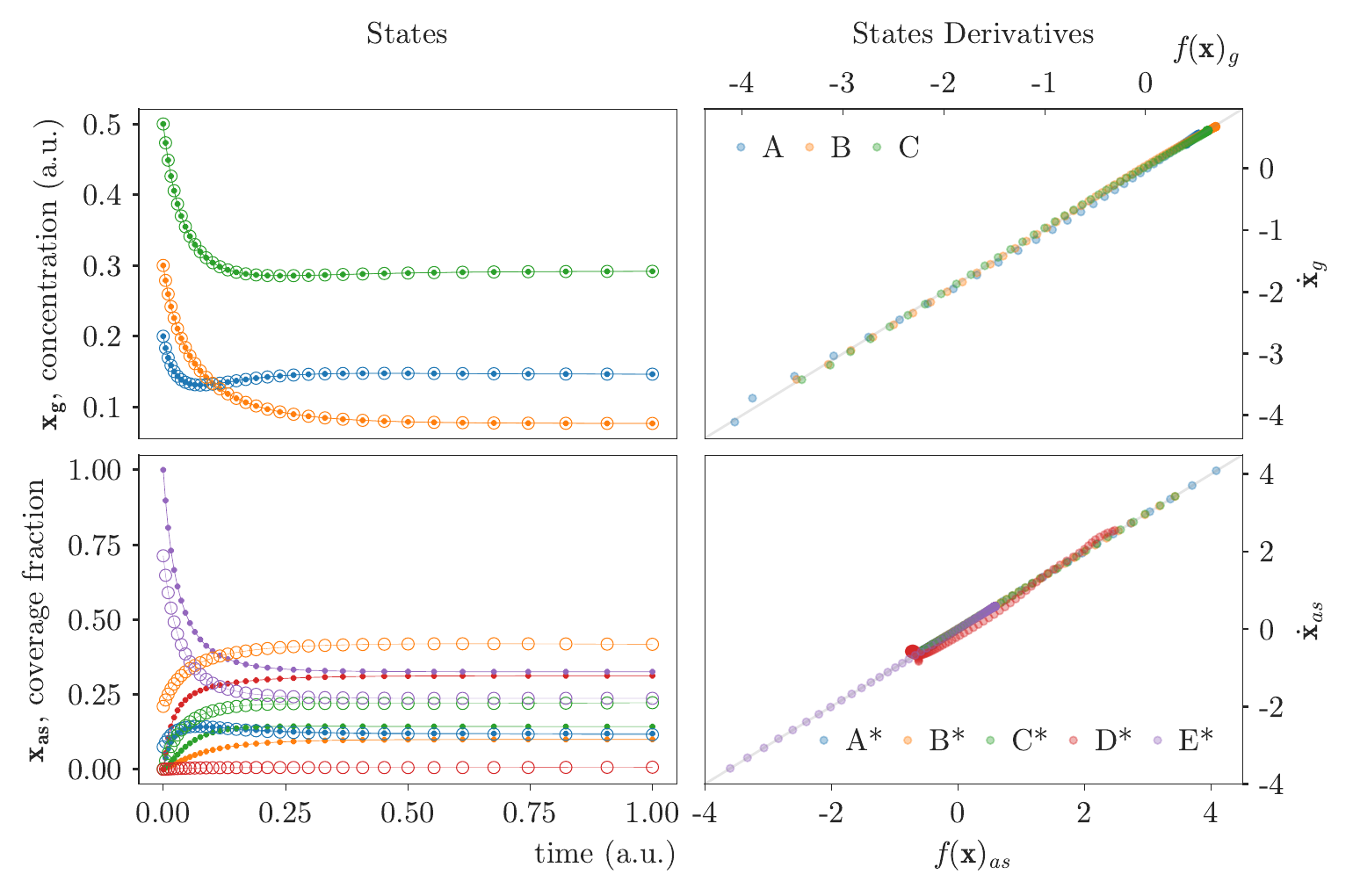}
    \caption{Reaction network type \textit{dc} inverse solution for \obsonly~data using \bc2. States solution (left) (\kUDE, open circles; observed data, closed markers) under minimization of \eqref{main:eqn:ode_nn_inv} (right) for observable variables (unbound species, top).}\label{fig:inv_dc_bc1}
\end{figure}

\begin{figure}[ht!]
    \centering
    
    \includegraphics[trim=0in 0in 0in 0in, keepaspectratio=true,scale=0.95,clip=True]{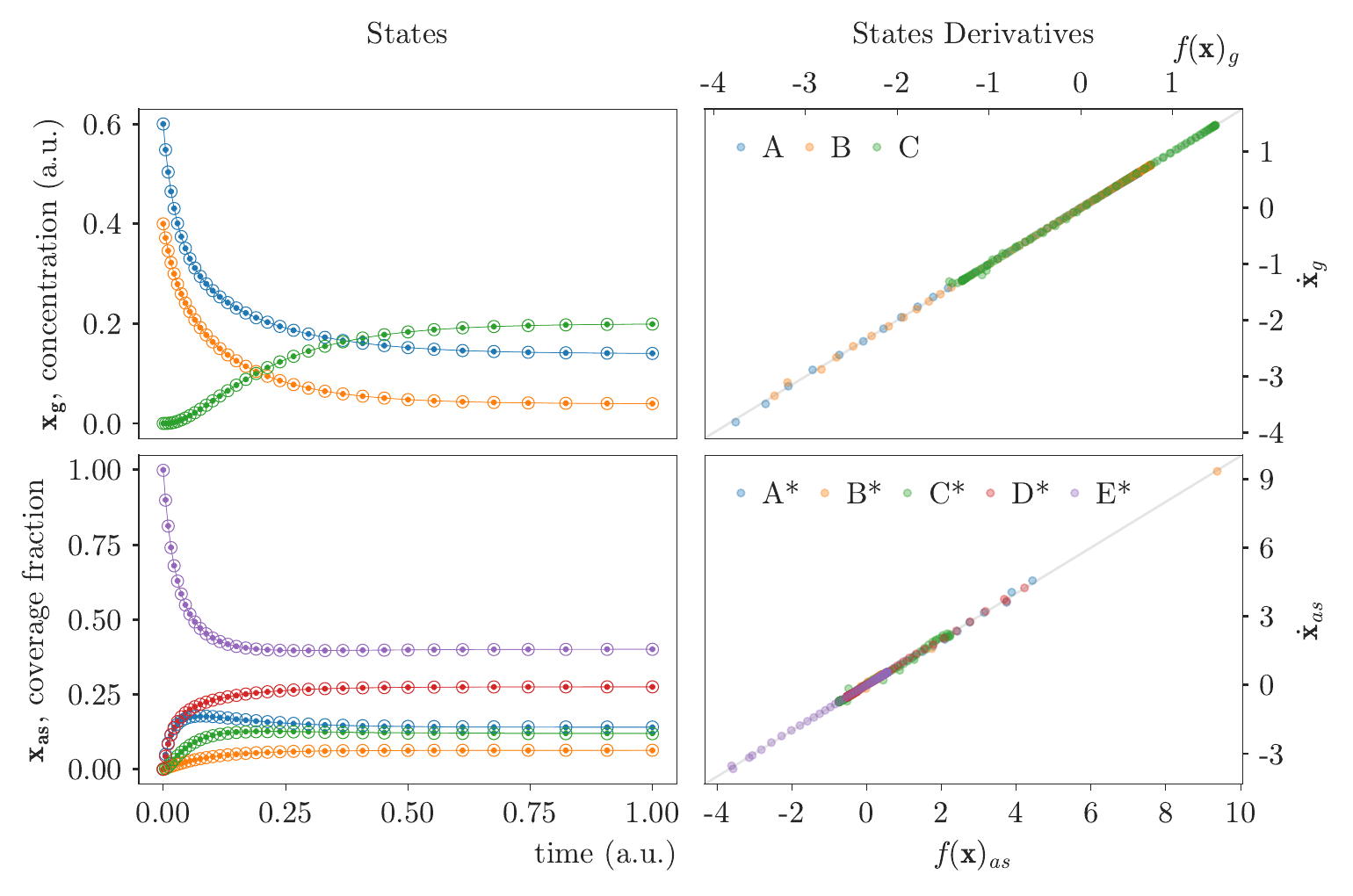}
    \caption{Reaction network type \textit{dc} inverse solution for \semiquant~data using \bc1. States solution (left) (\kUDE, open circles; observed data, closed markers) under minimization of \eqref{main:eqn:ode_nn_inv} (right) for observable variables (unbound species, top).}\label{fig:invsc_dc_bc0}
\end{figure}

\begin{figure}[ht!]
    \centering
    
    \includegraphics[trim=0in 0in 0in 0in, keepaspectratio=true,scale=0.95,clip=True]{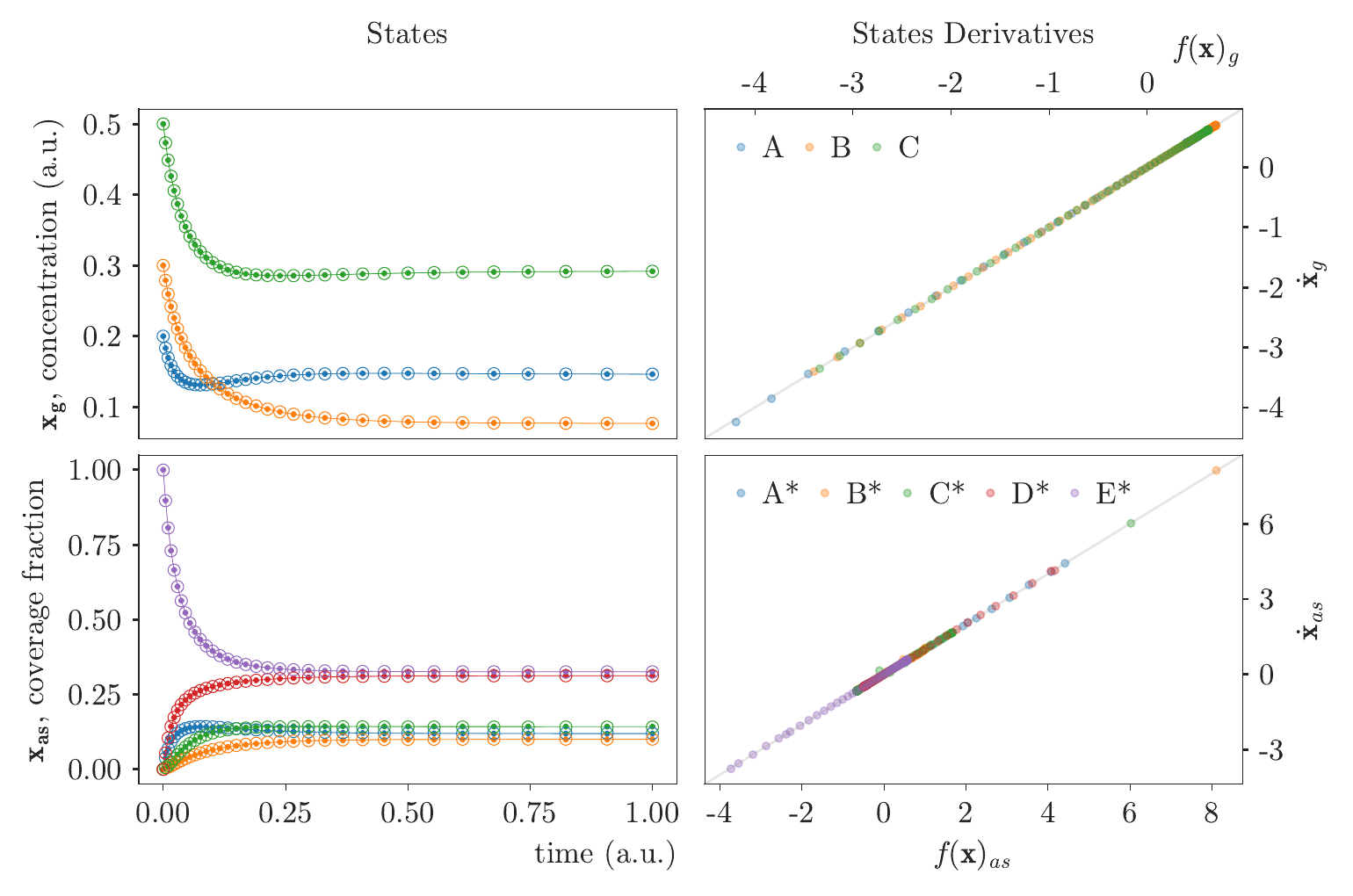}
    \caption{Reaction network type \textit{dc} inverse solution for \semiquant~data using \bc2. States solution (left) (\kUDE, open circles; observed data, closed markers) under minimization of \eqref{main:eqn:ode_nn_inv} (right) for observable variables (unbound species, top).}\label{fig:invsc_dc_bc1}
\end{figure}

\begin{figure}[ht!]
    \centering
    \includegraphics[trim=0in 0in 0in 0in, keepaspectratio=true,scale=0.95,clip=True]{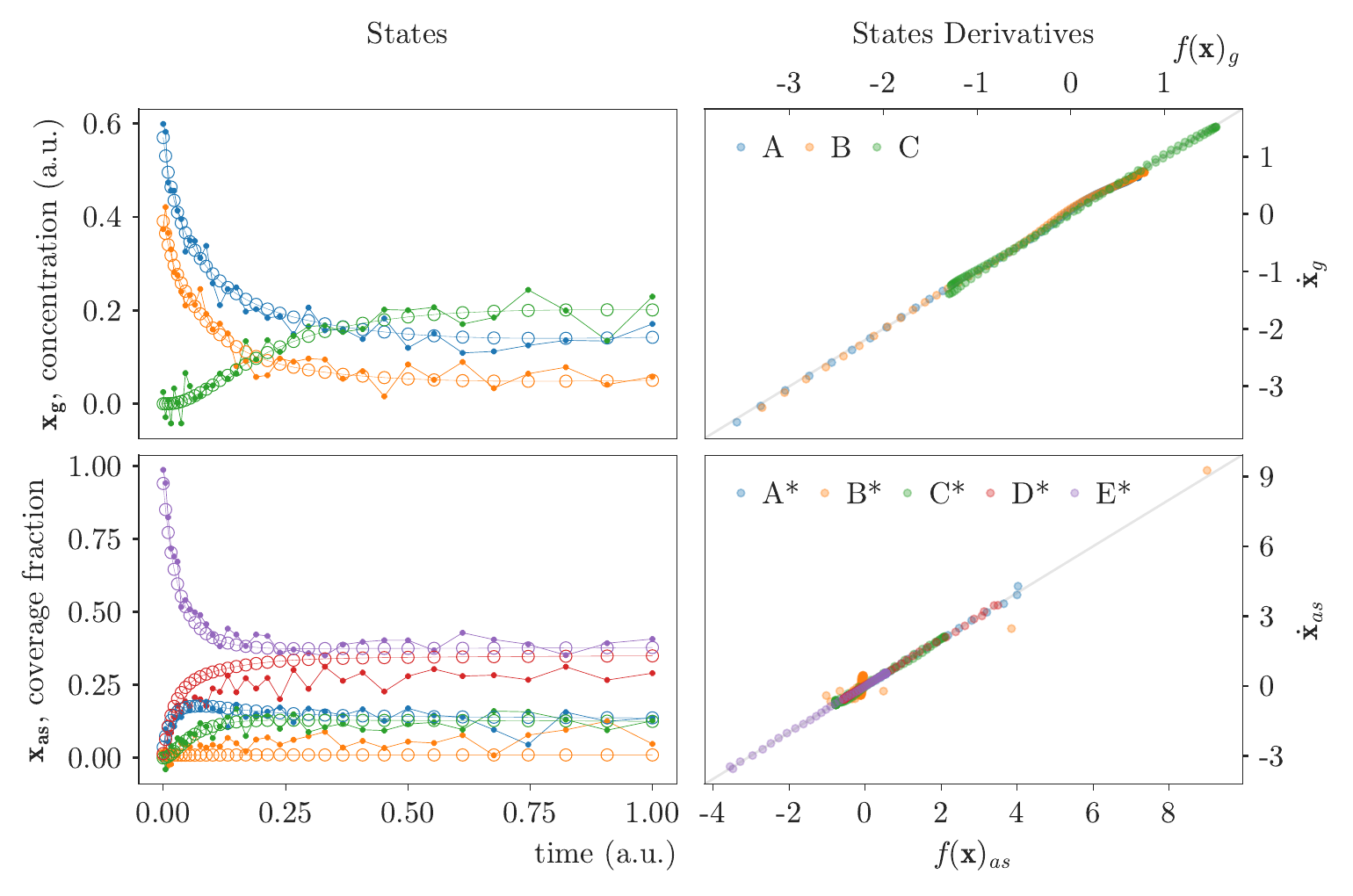}
    \caption{Reaction network type \textit{dc} inverse solution for \semiquantn~data using \bc1. States solution (left) (\kUDE, open circles; observed data, closed markers) under minimization of \eqref{main:eqn:ode_nn_inv} (right) for observable variables (unbound species, top).}\label{fig:invvwn_dc_bc0}
\end{figure}

\begin{figure}[ht!]
    \centering
    
    \includegraphics[trim=0in 0in 0in 0in, keepaspectratio=true,scale=0.95,clip=True]{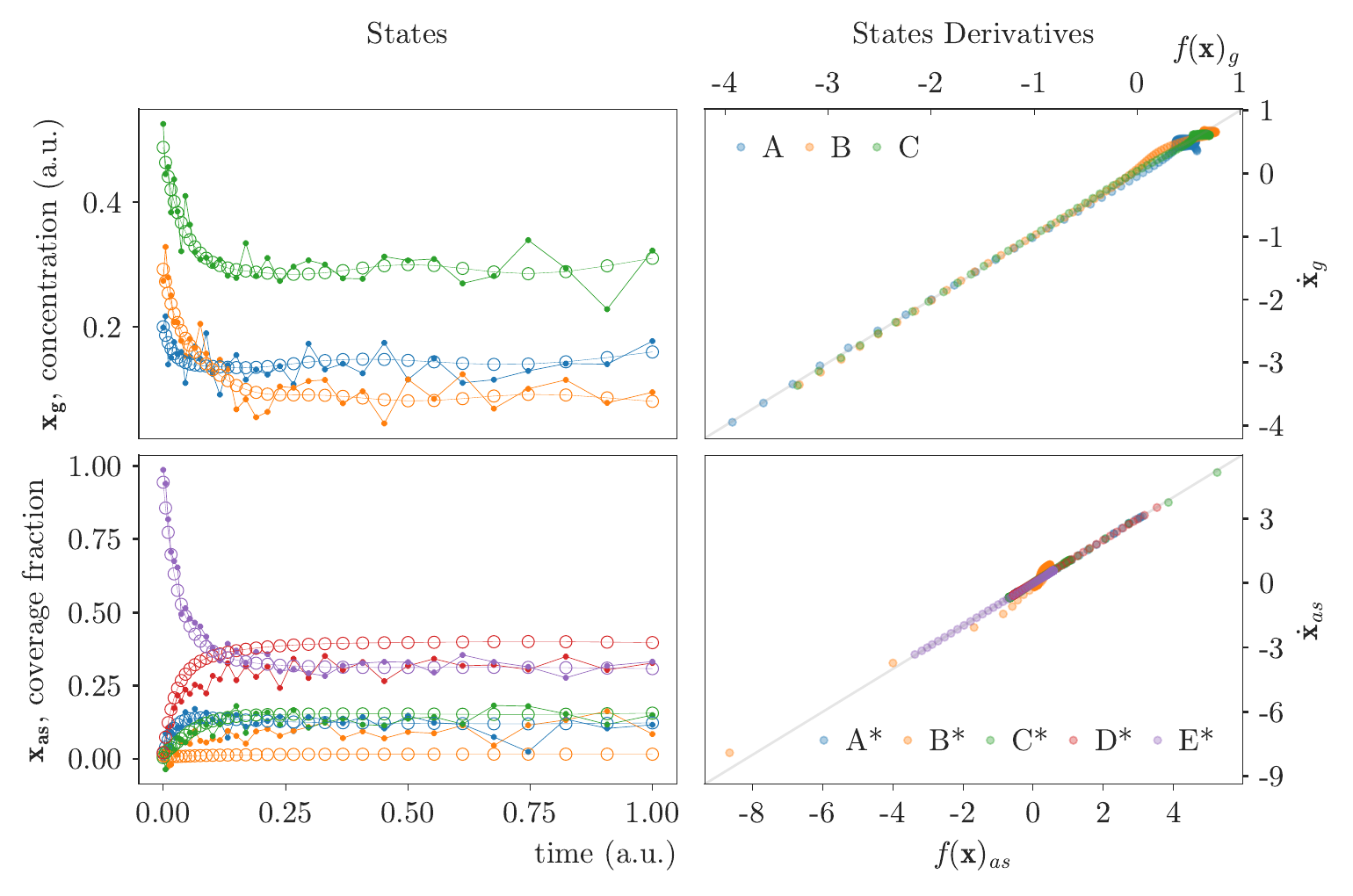}
    \caption{Reaction network type \textit{dc} inverse solution for \semiquantn~data using \bc2. States solution (left) (\kUDE, open circles; observed data, closed markers) under minimization of \eqref{main:eqn:ode_nn_inv} (right) for observable variables (unbound species, top).}\label{fig:invvwn_dc_bc1}
\end{figure}

\begin{figure}[ht!] 
    \centering
    \includegraphics[trim=0in 0in 0in 0in, keepaspectratio=true,scale=0.95,clip=True]{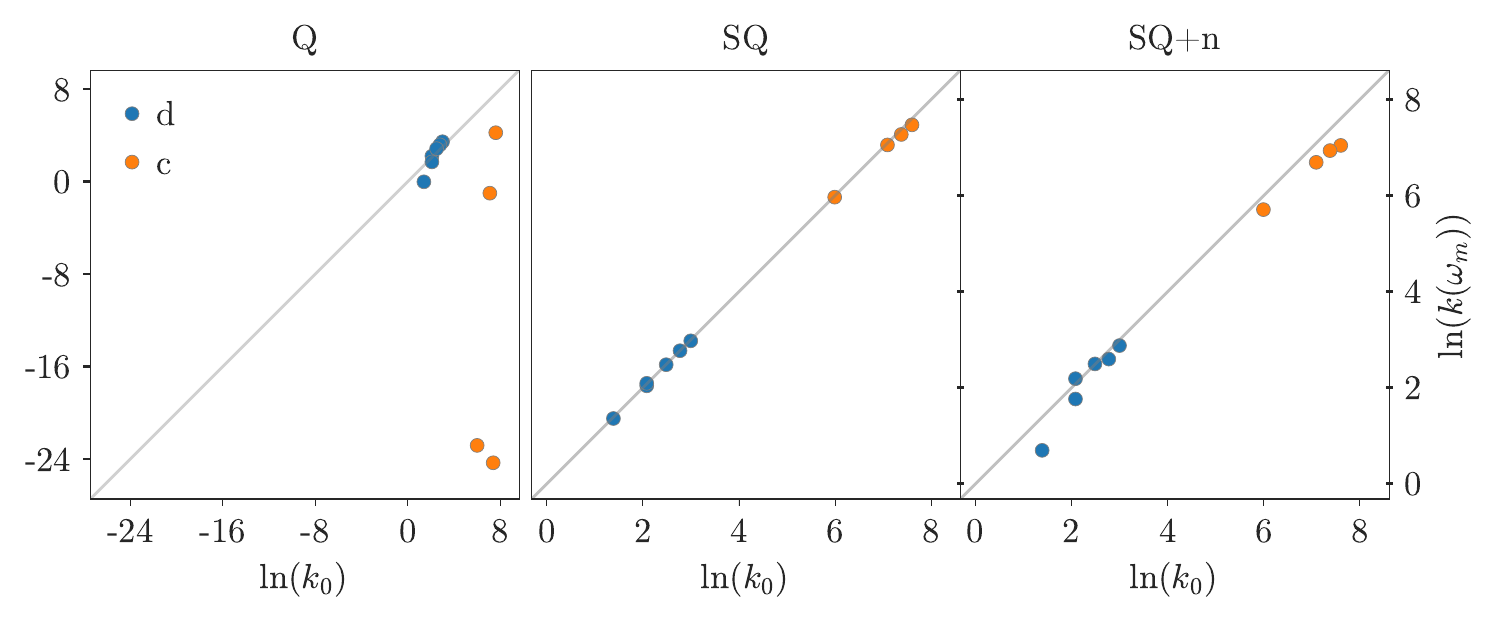}
    \caption{Reaction network type \textit{dc} inverse solution. Natural logarithms of ground truth parameter $\ln{(k_0)}$ vs. regressed parameters, $\ln(k(\parmkm))=\parmkm$, for \obsonly, \semiquant~and \semiquantn~data.}\label{fig:inv_dc_reg}
\end{figure}

\clearpage


\section{Reaction Network Type \textit{dcs}}

The \textit{s}-type elementary steps embody elementary reactions between surface intermediates exclusively. In the following example, $D*$, $E*$ and $F*$ are reaction intermediates for which there are no stable desorbed counterparts, which cannot be directly measured or inferred from ordinary analytical chemistry techniques.

\begin{align}
A*+*&\underset{k_{16}}{\stackrel{k_{15}}{\rightleftharpoons}} 2D*\tag{\textbf{c}.2}\\
B*+*&\underset{k_{18}}{\stackrel{k_{17}}{\rightleftharpoons}} 2E*\tag{\textbf{c}.3}\\
D*\:+\:E*&\underset{k_{20}}{\stackrel{k_{19}}{\rightleftharpoons}} F*\:+\:*\tag{\textbf{s}.1}\\
F*\:+\:E*&\underset{k_{22}}{\stackrel{k_{21}}{\rightleftharpoons}} C*\:+\:*\tag{\textbf{c}.4}
\end{align}

\begin{equation}
\begin{aligned}
    \ln(\mathbf{k}_0)^T=\left[3.00\:\;2.08\:\;3.18\:\;2.48\:\;2.77\:\;3.69\:\;6.46\:\;6.87\:\;5.08\:\;4.38\:\;6.46\:\;5.48\:\;6.33\:\;5.08\right]
    \\\\
    M=
    \left[
    \begin{array}{rrrrrrrrrrrrrr}
     -1&1&0&0&0&0&0&0&0&0&0&0&0&0 \\
      0&0&-1&1&0&0&0&0&0&0&0&0&0&0 \\
      0&0&0&0&-1&1&0&0&0&0&0&0&0&0 \\
      1&-1&0&0&0&0&-1&1&0&0&0&0&0&0 \\
      0&0&1&-1&0&0&0&0&-1&1&0&0&0&0 \\
      0&0&0&0&1&-1&0&0&0&0&0&0&1&-1 \\
      0&0&0&0&0&0&2&-2&0&0&-1&1&0&0 \\
      0&0&0&0&0&0&0&0&2&-2&-1&1&-1&1 \\
      0&0&0&0&0&0&0&0&0&0&1&-1&-1&1 \\
     -1&1&-1&1&-1&1&-1&1&-1&1&1&-1&1&-1
    \end{array}
    \right]\\\\
    \mathbf{x}^T=\left[x_{A}\;\;x_{B}\;\;x_{C}\;\;x_{A*}\;\;x_{B*}\;\;x_{C*}\;\;x_{D*}\;\;x_{E*}\;\;x_{F*}\;\;x_{*}\right]
\end{aligned}
\end{equation}

\begin{figure}[ht!]
    \centering
    \includegraphics[trim=0in 0in 0in 0in, keepaspectratio=true,scale=0.95,clip=True]{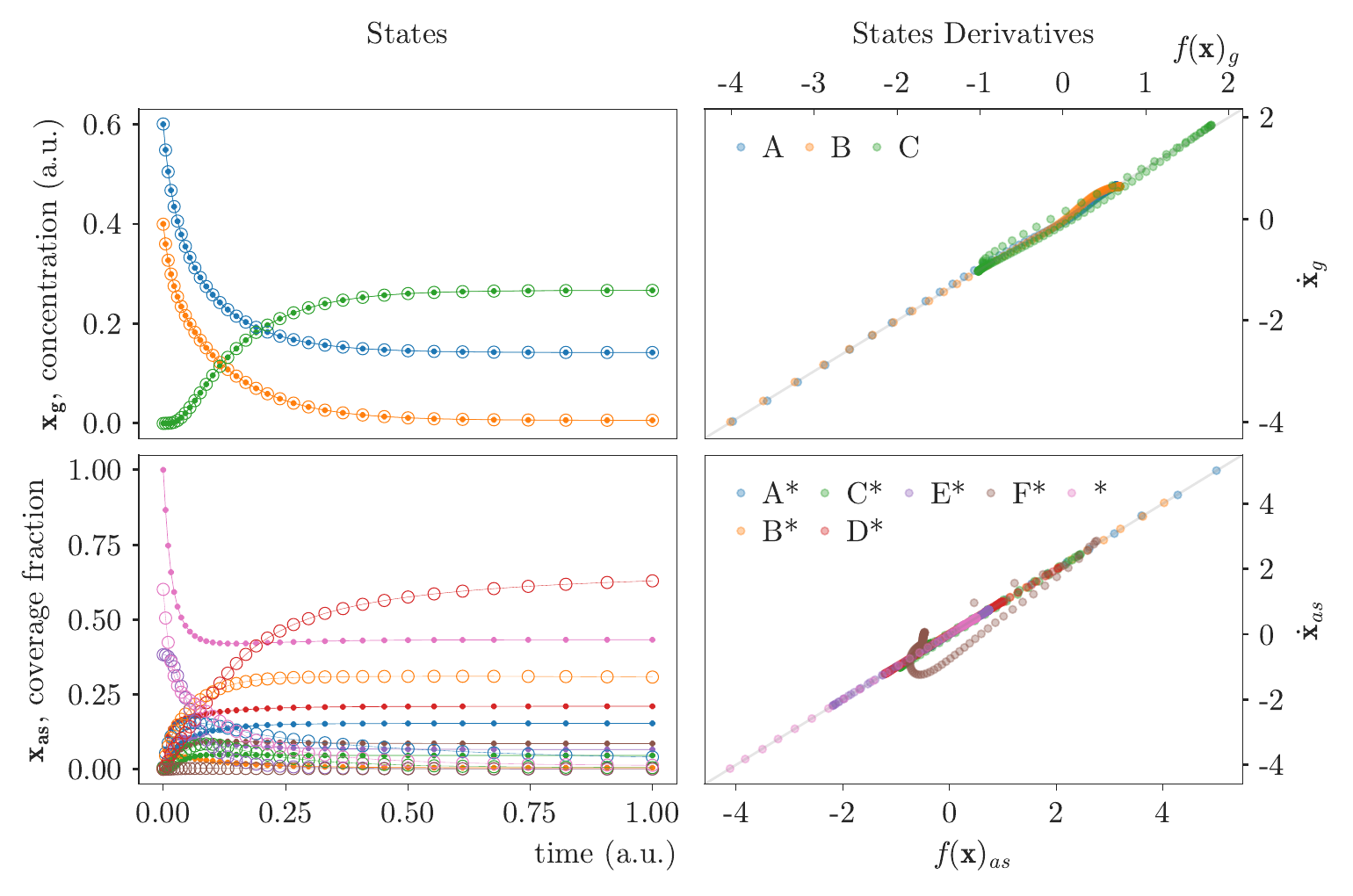}
    \caption{Reaction network type \textit{dcs} inverse solution for \obsonly~data using \bc1. States solution (left) (\kUDE, open circles; observed data, closed markers) under minimization of \eqref{main:eqn:ode_nn_inv} (right) for observable variables (unbound species, top).}\label{fig:inv_dcs_bc0}
\end{figure}

\begin{figure}[ht!]
    \centering
    \includegraphics[trim=0in 0in 0in 0in, keepaspectratio=true,scale=0.95,clip=True]{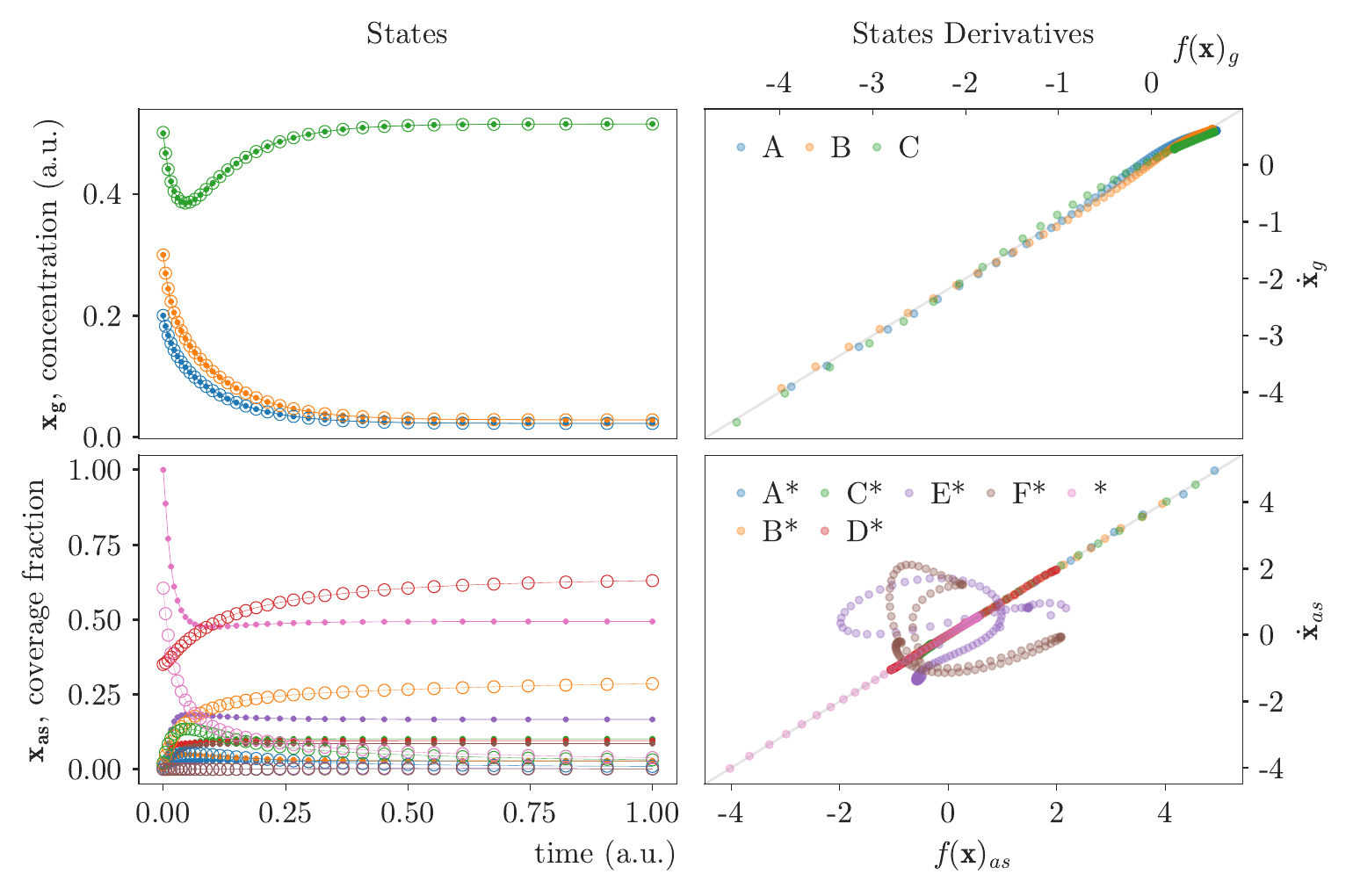}
    \caption{Reaction network type \textit{dcs} inverse solution for \obsonly~data using \bc2. States solution (left) (\kUDE, open circles; observed data, closed markers) under minimization of \eqref{main:eqn:ode_nn_inv} (right) for observable variables (unbound species, top).}\label{fig:inv_dcs_bc1}
\end{figure}

\begin{figure}[ht!]
    \centering
    \includegraphics[trim=0in 0in 0in 0in, keepaspectratio=true,scale=0.95,clip=True]{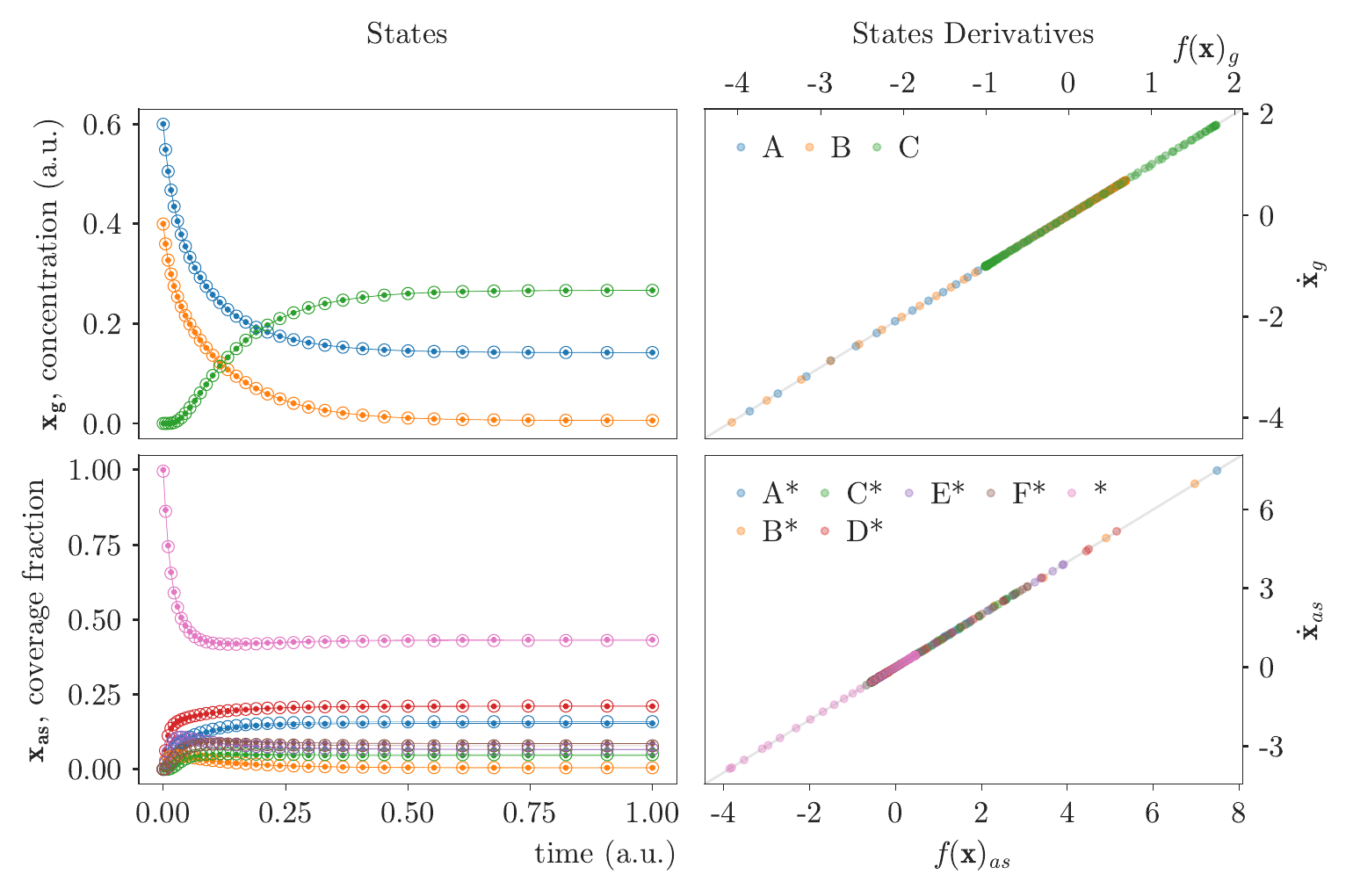}
    \caption{Reaction network type \textit{dcs} inverse solution for \semiquant~data using \bc1. States solution (left) (\kUDE, open circles; observed data, closed markers) under minimization of \eqref{main:eqn:ode_nn_inv} (right) for observable variables (unbound species, top).}\label{fig:invsc_dcs_bc0}
\end{figure}

\begin{figure}[ht!]
    \centering
    \includegraphics[trim=0in 0in 0in 0in, keepaspectratio=true,scale=0.95,clip=True]{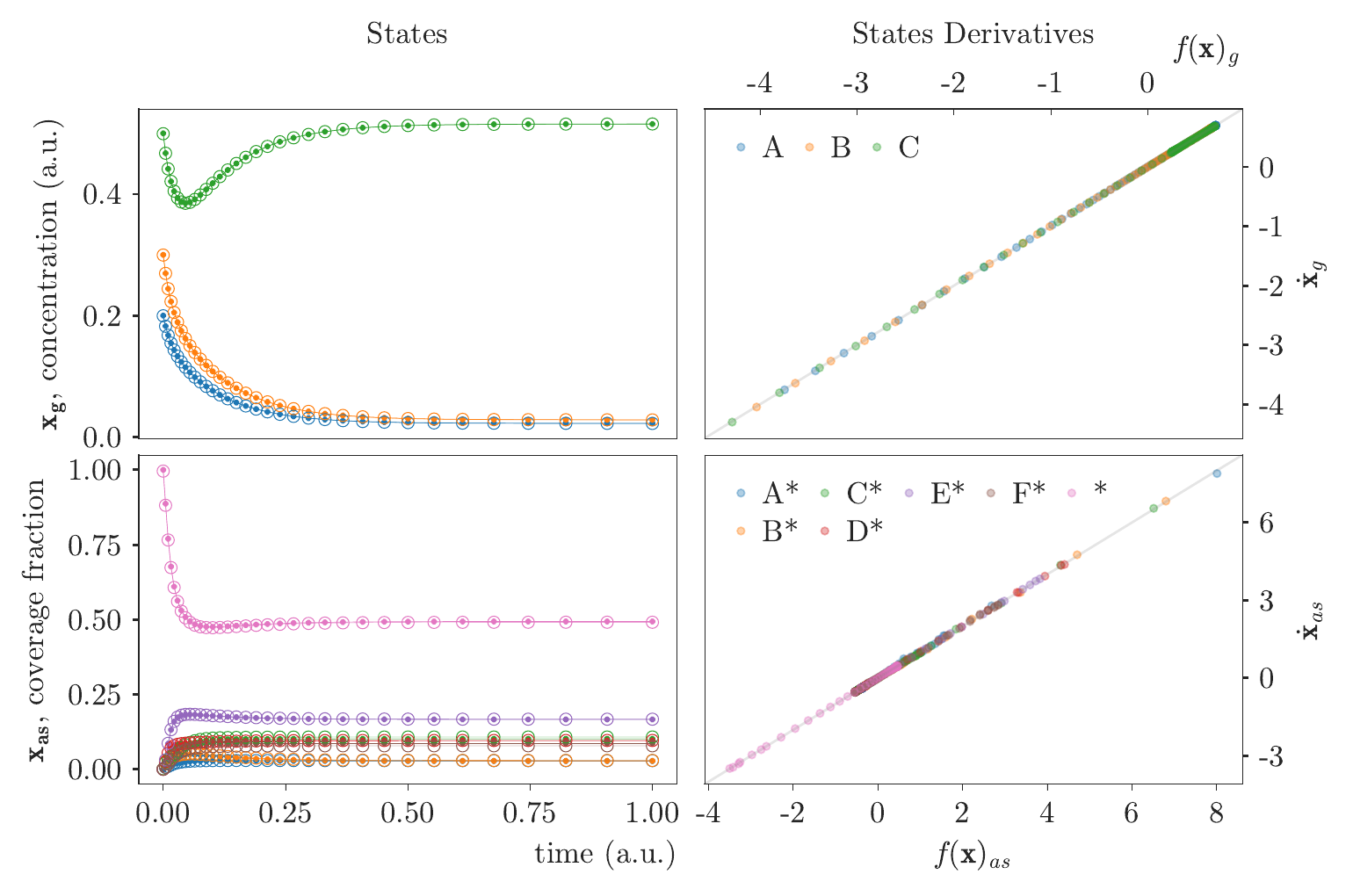}
    \caption{Reaction network type \textit{dcs} inverse solution for \semiquant~data using \bc2. States solution (left) (\kUDE, open circles; observed data, closed markers) under minimization of \eqref{main:eqn:ode_nn_inv} (right) for observable variables (unbound species, top).}\label{fig:invsc_dcs_bc1}
\end{figure}

\begin{figure}[ht!]
    \centering
    \includegraphics[trim=0in 0in 0in 0in, keepaspectratio=true,scale=0.95,clip=True]{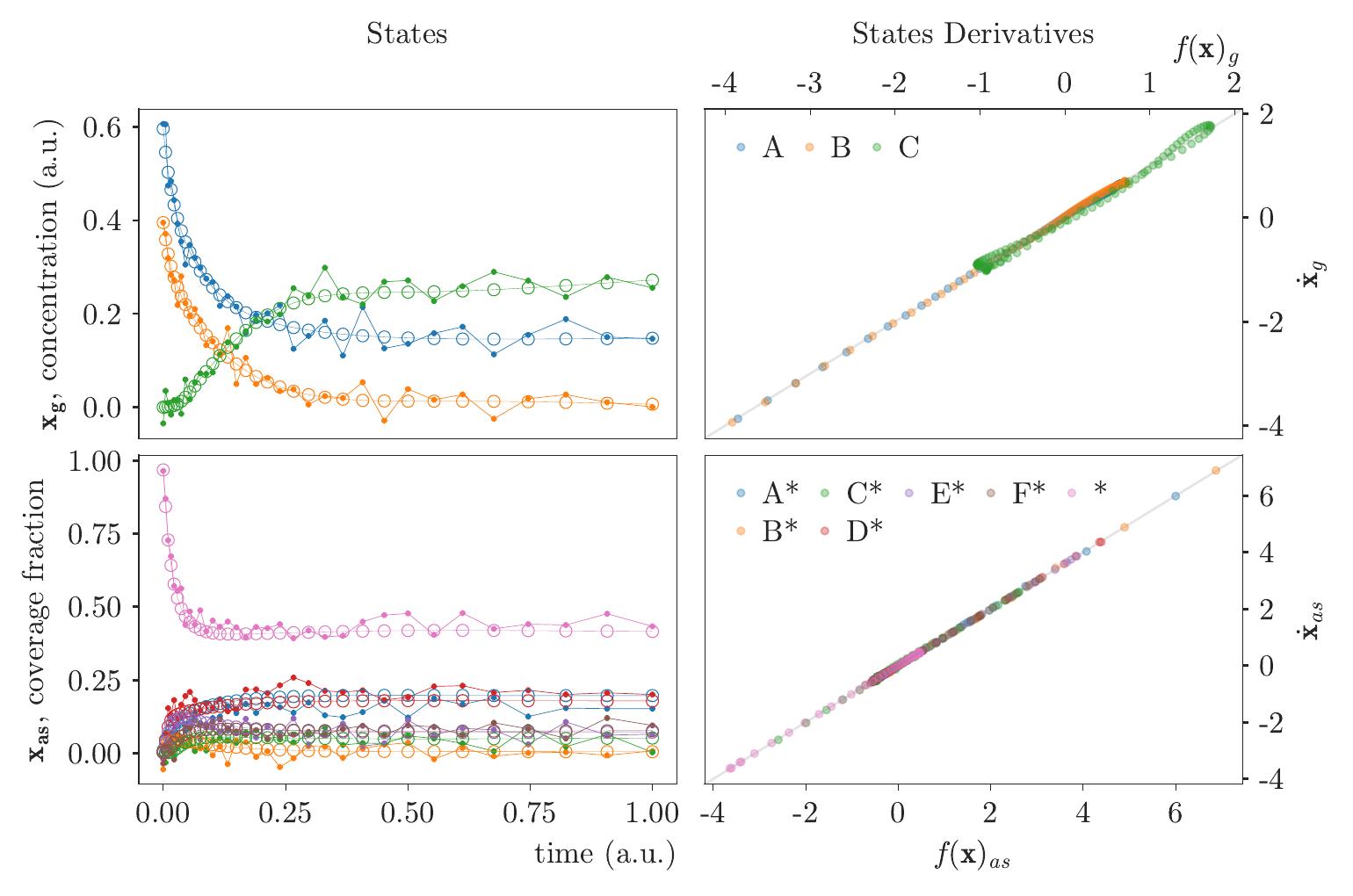}
    \caption{Reaction network type \textit{dcs} inverse solution for \semiquantn~data using \bc1. States solution (left) (\kUDE, open circles; observed data, closed markers) under minimization of \eqref{main:eqn:ode_nn_inv} (right) for observable variables (unbound species, top).}\label{fig:invvwn_dcs_bc0}
\end{figure}

\begin{figure}[ht!]
    \centering
    \includegraphics[trim=0in 0in 0in 0in, keepaspectratio=true,scale=0.95,clip=True]{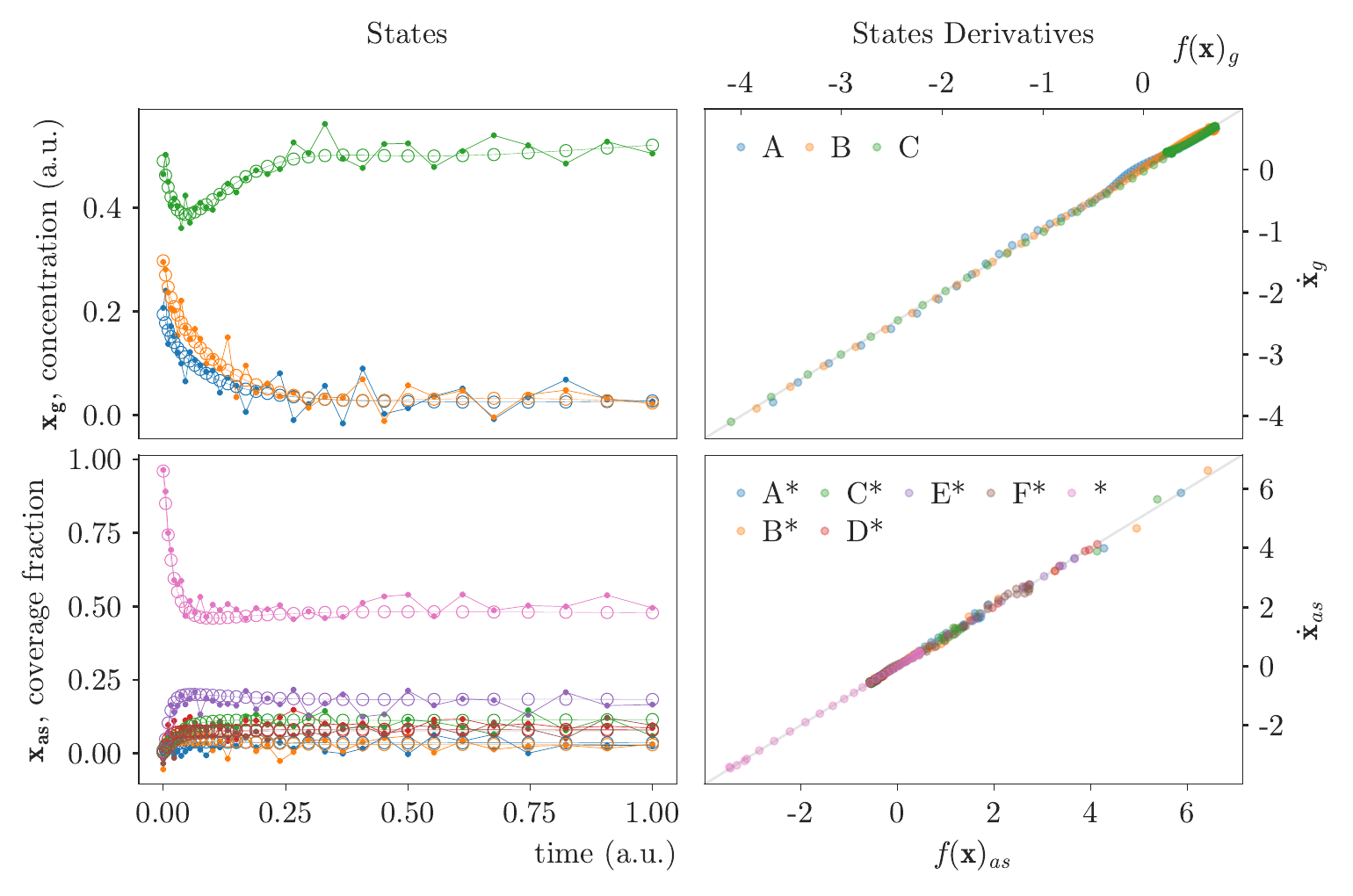}
    \caption{Reaction network type \textit{dcs} inverse solution for \semiquantn~data using \bc2. States solution (left) (\kUDE, open circles; observed data, closed markers) under minimization of \eqref{main:eqn:ode_nn_inv} (right) for observable variables (unbound species, top).}\label{fig:invvwn_dcs_bc1}
\end{figure}

\begin{figure}[ht!] 
    \centering
    \includegraphics[trim=0in 0in 0in 0in, keepaspectratio=true,scale=1.,clip=True]{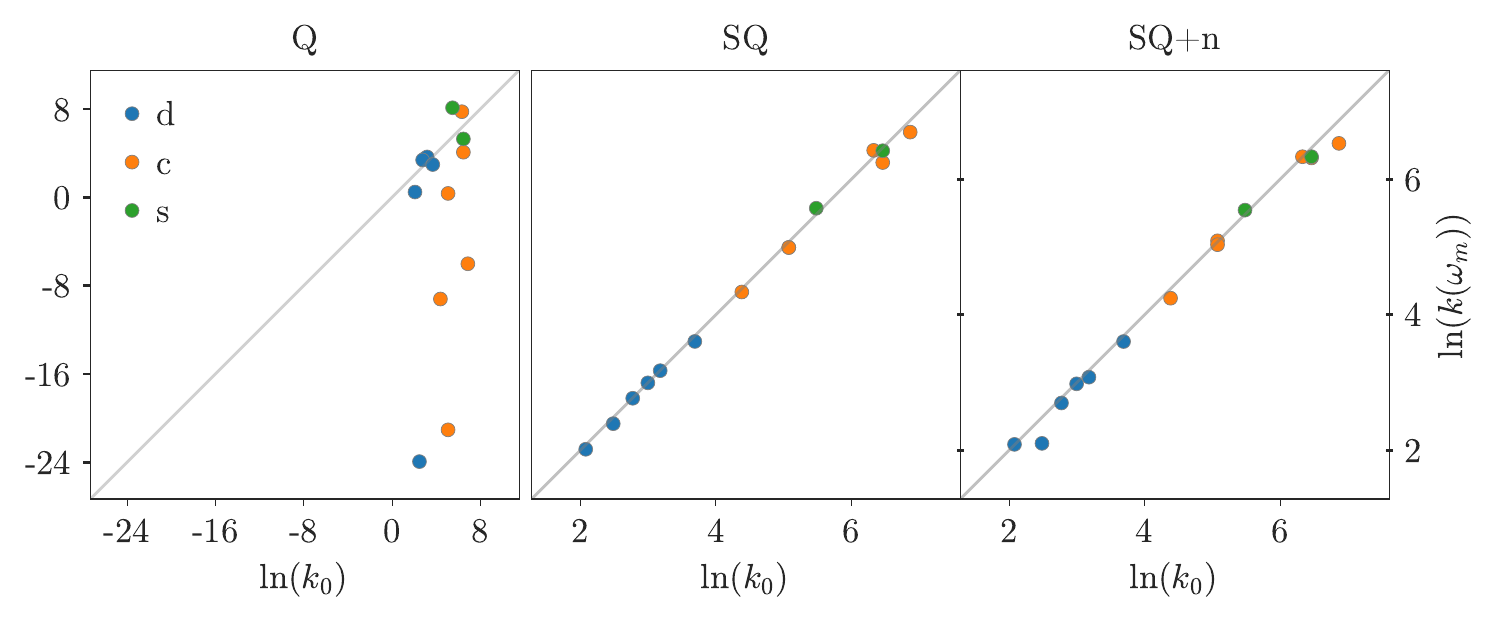}
    \caption{Reaction network type \textit{dcs} inverse solution. Natural logarithms of ground truth parameter $\ln{(k_0)}$ vs. regressed parameters, $\ln(k(\parmkm))=\parmkm$, for \obsonly, \semiquant~and \semiquantn~data.}\label{fig:inv_dcs_reg}
\end{figure}

\clearpage